\documentclass[10pt,twocolumn,letterpaper]{article}

\usepackage{cvpr}              

\usepackage{graphicx}
\usepackage{amsmath}
\usepackage{amssymb}
\usepackage{booktabs}
\usepackage{color,soul}
\usepackage{arydshln}

\newcommand{\FData}{\textit{Fusion 360 Gallery} assembly dataset}
\newcommand{\volume}{{\ooalign{\hfil$V$\hfil\cr\kern0.08em--\hfil\cr}}}

\newcommand\Tstrut{\rule{0pt}{2.75ex}}       
\newcommand\Bstrut{\rule[-1.5ex]{0pt}{0pt}} 

\def\code#1{\texttt{#1}}

\usepackage{paralist}

  \pltopsep=\medskipamount
\usepackage[pagebackref,breaklinks,colorlinks]{hyperref}

\usepackage[capitalize]{cleveref}
\crefname{section}{Sec.}{Secs.}
\Crefname{section}{Section}{Sections}
\Crefname{table}{Table}{Tables}
\crefname{table}{Tab.}{Tabs.}

\def\cvprPaperID{1667} 
\def\confName{CVPR}
\def\confYear{2022}

\begin{document}

\title{JoinABLe: Learning Bottom-up Assembly of Parametric CAD Joints}

\author{
Karl D.D. Willis$^{1}$\quad 
Pradeep Kumar Jayaraman$^{1}$ \quad
Hang Chu$^{1}$ \quad
Yunsheng Tian$^{2}$ \quad
Yifei Li$^{2}$ \and 
Daniele Grandi $^{1}$\quad
Aditya Sanghi$^{1}$\quad
Linh Tran$^{1}$\quad
Joseph G. Lambourne$^{1}$\and
Armando Solar-Lezama$^{2}$\quad
Wojciech Matusik$^{2}$ \\
$^{1}$Autodesk Research \quad
$^{2}$MIT CSAIL \\
}

\maketitle

\begin{abstract}
Physical products are often complex assemblies combining a multitude of 3D parts modeled in computer-aided design (CAD) software. CAD designers build up these assemblies by aligning individual parts to one another using constraints called joints. In this paper we introduce JoinABLe, a learning-based method that assembles parts together to form joints. JoinABLe uses the weak supervision available in standard parametric CAD files without the help of object class labels or human guidance. 
Our results show that by making network predictions over a graph representation of solid models
we can outperform multiple baseline methods with an accuracy (79.53\%) that approaches human performance (80\%). Finally, to support future research we release the \FData{}, containing assemblies with rich information on joints, contact surfaces, holes, and the underlying assembly graph structure.
\end{abstract}

\section{Introduction}

The physical products that surround us every day are often complex assemblies combining a multitude of parts modeled using computer-aided design (CAD) software. Well-designed assemblies are critical to ensure that products are cost-efficient, reliable, and easy to physically assemble.
CAD designers build up assemblies by aligning pairs of parts together using constraints called \textit{joints}. These joints determine the relative pose and allowed degrees of freedom (DOF) of parts in an assembly~\cite{lupinetti2019content}. For example, a bolt can be constrained to a hole, then a nut constrained to the bolt, and so on until an entire assembly is designed. 
Assemblies may contain thousands of parts, represented as solid models in the boundary representation (B-Rep) format~\cite{lee2001partialentity, weiler1986}, and are used for everything from furniture, to vehicles, to electronic devices. Defining individual global positions for each part without using joints quickly becomes cumbersome and prone to error.
Joints enable designers to make quick parametric changes to a design while preserving existing part relationships and maintaining design intent.

\begin{figure}
    \begin{center}
        \includegraphics[width=\columnwidth]{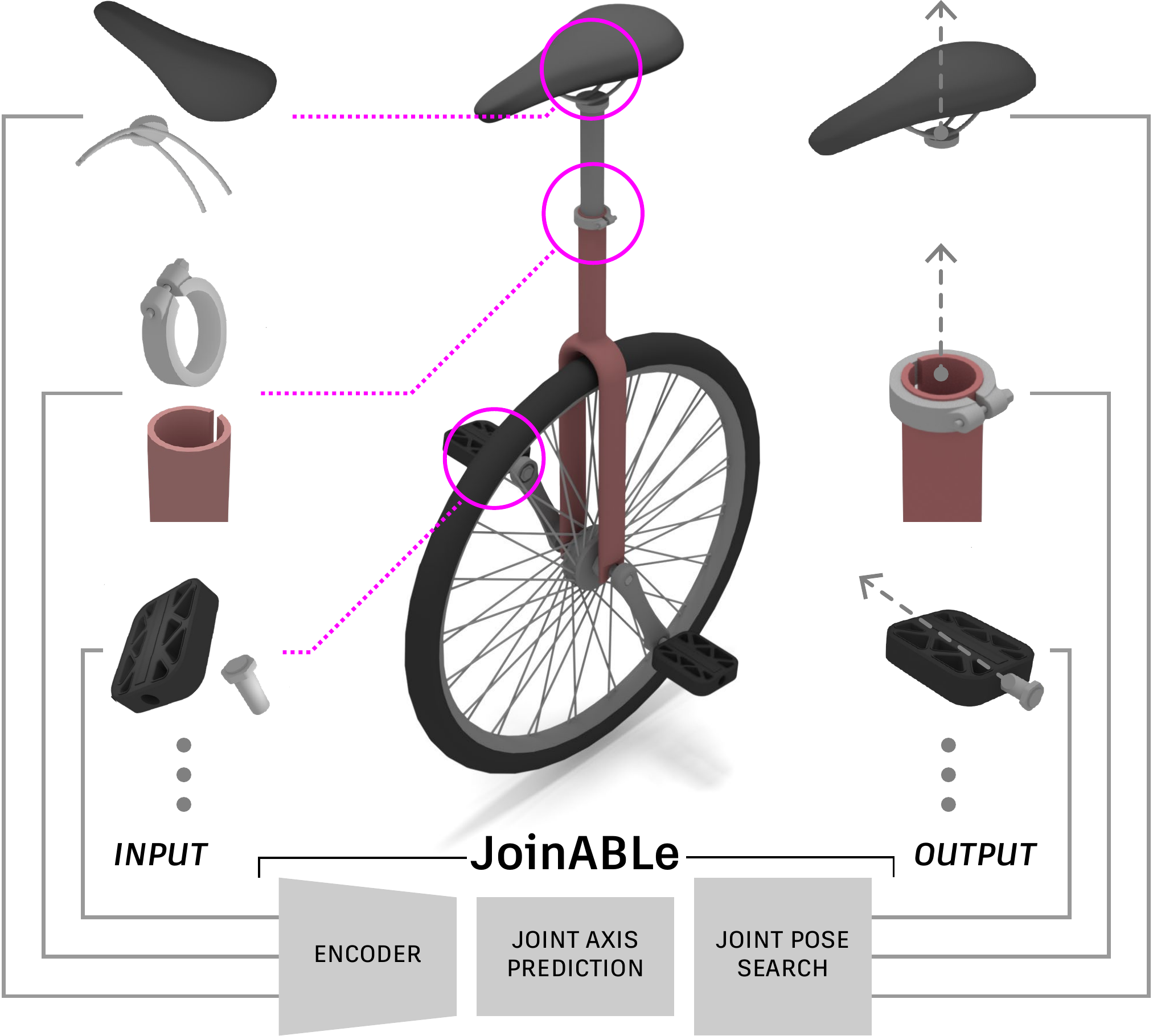}
        \caption{CAD assemblies contain valuable \textit{joint} information describing how parts are locally constrained and positioned together. We use this weak supervision to learn a bottom-up approach to assembly. \textit{JoinABLe} combines an encoder and joint axis prediction network together with a neurally guided joint pose search to assemble pairs of parts without class labels or human guidance.}
        \label{figure:teaser}
    \end{center}
\end{figure}

However, fully defining joints in assemblies is time-consuming -- roughly one third of time in CAD is spent doing assembly work~\cite{jones2021sb}. As a result many assemblies have missing or partly defined joints. 
A learning-based approach capable of predicting joints could ease the burden of joint definition and enable other
applications such as CAD assembly synthesis~\cite{sung2017complementme}, robotic assembly~\cite{lee2019ikea}, optimization of dynamic assemblies~\cite{zhao2020robogrammar}, part motion prediction~\cite{wang2019shape2motion}, assembly-aware similarity search~\cite{boussuge2019application} and many more. Although joints for real world assemblies are configured in a bottom-up fashion, recent work largely takes a top-down approach to assembly related tasks\cite{li2020learning3D, huang2020generative, harish2021rgl}. Top-down approaches learn a global arrangement of parts from set object and part classes in carefully annotated data. An open challenge remains to learn to assemble parts without relying on the strong object and part class priors provided in heavily annotated datasets. In this work we ask the following question, illustrated in Figure~\ref{figure:teaser}: Given a pair of parts, can we automatically assemble them without prior knowledge of the global design, class labels, or additional human input?
Solving this problem is a fundamental building block for leveraging learning-based methods with assemblies. Our long-term motivation is to enable the next generation of assembly aware tools that can increase the reuse of existing components and streamline robotic assembly and disassembly -- important steps in reducing the negative impact of physical products~\cite{kerr2001eco, marconi2019feasibility, li2019multi, liu2019human, chang2017approaches}.

To begin to address this challenge we introduce \textit{\mbox{JoinABLe}} (Joint Assembly Bottom-up Learning), our \textit{bottom-up} approach to assembly that learns how parts connect locally to form parametric CAD joints. \textit{JoinABLe} uses the weak supervision available in parametric CAD files, containing only partial joint labels, to automatically assemble pairs of parts. We make the following contributions:
\begin{itemize}
	\item We propose a novel learning-based method to automatically assemble pairs of parts using the weak supervision available in parametric CAD files. We do this without the help of object or part class labels, human annotation, or user guidance for the first time.
    \item We create and release the \FData{}, containing CAD assemblies with rich information on joints, contact surfaces, holes, and the underlying assembly graph structure.
    \item We provide experimental results on both joint axis and joint pose prediction tasks, a human baseline study, and comparisons with multiple other methods.
\end{itemize}

Our results show that by making network predictions over a graph representation of solid models, we can outperform multiple baseline methods while using fewer network parameters. We demonstrate that our approach performs well with difficult cases where heuristic algorithms can struggle and achieves an accuracy (79.53\%) that approaches human performance (80\%) on the joint axis prediction task.

\section{Related Work}
Assemblies have been a critical part of design and engineering for centuries. Since the digitization of CAD in the 1980s, a number of research areas have been explored.

\textbf{Shape Combination}~~
As early as 2004 the power of designing assemblies by combining and reusing existing parts was demonstrated in \textit{Modeling by Example}~\cite{funkhouser2004modeling}. 
Since then a body of work has focused on finding compatible parts to combine together into assemblies \cite{zheng2013smart,huang2015analysis,xu2012fit,jain2012exploring, chaudhuri2011probabilistic}. The ability to parametrically assemble parts into novel designs has numerous applications in the media and entertainment industry, where digital worlds can be populated with novel content. Other lines of work have focused on assemblies that can be physically fabricated~\cite{lau2011converting,schulz2014design,song2017reconfigurable,wang2018desia,shao2016dynamic,fu2015computational,lin2017recovering,wang2021sota} or conversion to/from assembly instructions~\cite{agrawala2003designing, shao2016dynamic}. Our work differs in that we automate the pair-wise assembly of real-world CAD parts using a learning based method without class labels or human guidance.

\textbf{Structure Aware Deep Generative Models}~~
3D shape synthesis has rapidly advanced with the use of structure aware deep generative models~\cite{chaudhuri2020learning, li2020learning, mo2019structurenet, gao2019sdm, wu2020pq, schor2019componet, gadelha2020learning, jones2020shapeassembly, jie2020dsgnet} that incorporate some notion of assembly structure to describe how the parts of a shape form a whole. 
Rather than synthesize the parts themselves, we focus instead on assembling existing parts in the industry-standard B-Rep format.

\begin{figure*}[!t]
    \begin{center}
        \includegraphics[width=\textwidth]{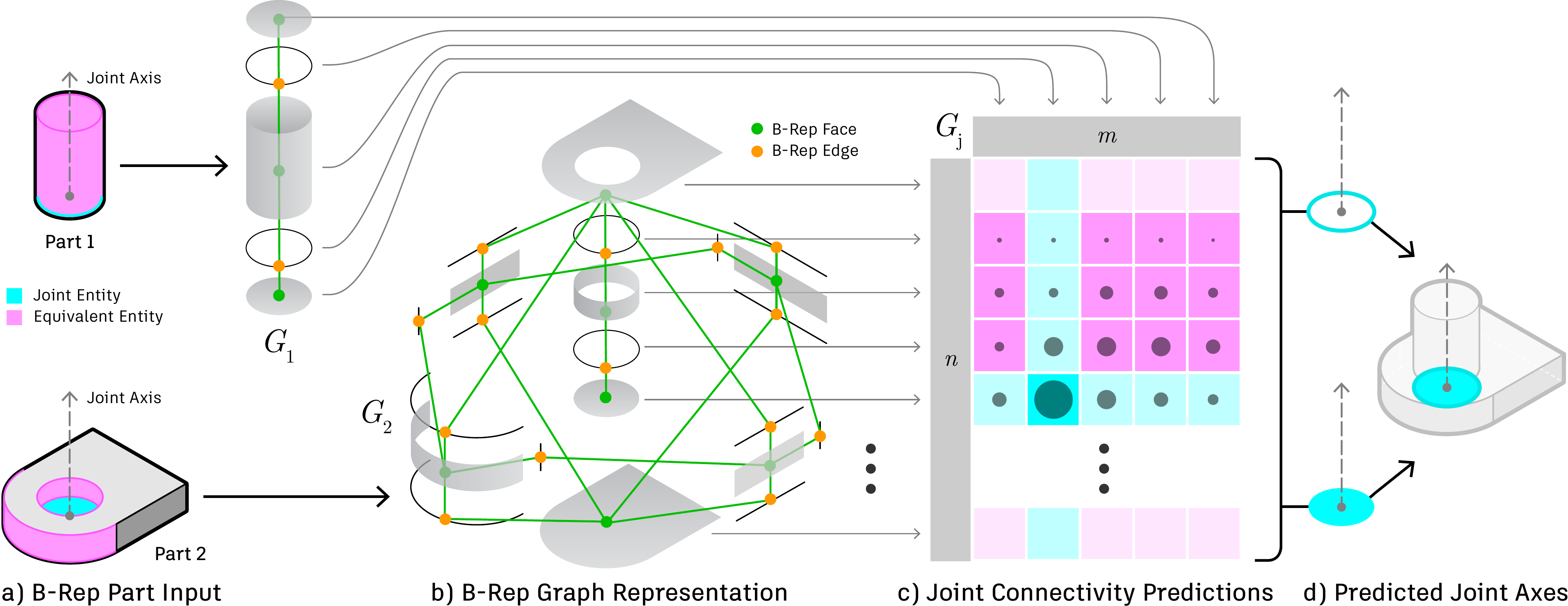}
        \caption{\textit{JoinABLe} is used to assemble a pair of parts in the B-Rep format (a). We use supervision from parametric CAD files containing user selected B-Rep faces and edges that define joints (cyan). We also identify `equivalent' faces and edges (pink) sharing the same joint axis for use during evaluation.
        Graphs for each part $G_1, G_2$ are constructed from adjacent B-Rep faces and edges (b), then joint connectivity predictions are made over a graph $G_j$ containing dense connections between all graph vertices. $G_j$ is shown as an $n\times m$ matrix (c) to visualize the prediction space. Finally, the parts are aligned along the predicted joint axes (d), ready for a subsequent search stage.}
        \label{figure:representation}
    \end{center}
\end{figure*}

\textbf{CAD Informed Robotic Assembly}~~
Prior knowledge of CAD assemblies has been leveraged for robotic assembly planning~\cite{ghandi2015review, halperin2000general} and sequencing~\cite{jimenez2013survey, de1989correct} to constrain the search process and validate assembly sequences.
Although not addressed in this work, we envision our approach can aid in improving the sampling efficiency of reinforcement learning based robotic assembly~\cite{thomas2018learning} by inferring joint information when it is absent or not fully specified.

\textbf{Learning to Assemble}~~
Learning-based assembly methods from the literature largely follow a top-down approach that predicts the absolute pose of a set of parts to form an assembly~\cite{sung2017complementme, li2020learning3D, yin2020coalesce, huang2020generative}. Predicting the absolute pose, however, can lead to noisy results where parts fail to completely align. To deal with this issue several recent works have leveraged supervision from local contact points between parts~\cite{han2020compositionally, harish2021rgl}. We believe a bottom-up approach is a critical part of solving the assembly problem. Rather than rely on contact points, our work uses the joint information found in parametric CAD files as weak supervision. This allows the output of our method to be reconstructed as fully editable parametric CAD files.

Critical to prior work is training on synthetic assemblies~\cite{mo2019partnet, xiang2020sapien} that belong to set object classes, e.g. chairs, drawers, etc., and are manually segmented, annotated with part class labels, and oriented in a consistent manner.
However, semantic segmentation is often incompatible with real-world CAD assemblies that segment parts by manufacturing process~\cite{lupinetti2019content}.
Moreover, while training on set object classes greatly improves within-class performance, generalization to unseen categories is an ongoing area of research~\cite{han2020compositionally}. Rather than rely on heavily annotated datasets with strong class priors, our work leverages the weak supervision readily available in standard parametric CAD files, and is trained \textit{without} object classes.

Concurrent to our work, AutoMate~\cite{jones2021sb} leverages similar joint information for use with a learning based recommendation system. Here the user selects an area on each part as guidance, and using those selections, AutoMate recommends to the user multiple joint solutions confined to the user-selected input area. Similar to AutoMate, our method enables editable joints to be created in CAD, but we do so in an automated way that does not require user guidance and is not limited to a predefined area. We believe providing an automated solution is critical to enabling advanced assembly applications for CAD and robotics.

\textbf{Part Mobility}~~
Understanding how assembled parts might move, i.e. \textit{part mobility}, is an important problem in both CAD and robotics where the goal is to articulate a given part, such as a hinged door, without knowing the part mobility in advance. Most relevant to our work are systems that automatically predict the relative joint configurations between pairs of parts~\cite{wang2019shape2motion, yanRPMNet19, li2020category}. Here the input is a point cloud and the output joint axis parameters that define how the parts move in relation to one another. Again, these works rely on strong class priors and heavily annotated synthetic assembly data. We compare our method with adaptions of several part mobility baselines in Section~\ref{section:experiments}.

\section{Method}
We now present our method, \textit{JoinABLe}, for automatically assembling pairs of parts with joints.

\subsection{CAD Joints}
Assembly parts are typically represented in the B-Rep format, containing a watertight collection of trimmed parametric surfaces connected together by a well-structured graph~\cite{weiler1986}. Each face contains a parametric surface, and is bounded by edges that define the trimmed extent of the surface using parametric curves such as lines, arcs, and circles. The B-Rep format is used in all mechanical CAD tools and the selection of B-Rep entities, i.e. faces and edges, is a critical but time-consuming manual task required to set up joints. Our method proposes to learn from these user selections to automate the process of joint creation.

The best practice for CAD assembly is to define relative relationships between pairs of parts, to form joints, also known as \textit{mates}. Joints define the degrees of freedom between the two parts, the parameters for a rest state pose, and the overall motion limits. CAD users select B-Rep entities on each part (highlighted in cyan in Figure~\ref{figure:representation}a) to define a per-part joint axis consisting of an origin point and a direction vector. The joint axes are determined by the type of geometry selection, for a circle the center point becomes the origin point and the normal becomes the direction vector. These two parts can then be aligned along their axes into an assembled state (Figure~\ref{figure:representation}d).

\subsection{Joint Prediction Problem Statement}
Given a pair of parts (Figure~\ref{figure:representation}a), we aim to create a parametric joint between them, such that the two parts are constrained relative to one another with the same joint axis and pose as defined by the ground truth (Figure~\ref{figure:representation}d). 
Here the joint axis is defined by two joint origin points and joint direction vectors relative to each part, and the pose is defined by a single rigid transformation in absolute coordinates. We refer to the tasks of predicting these values as \textit{joint axis prediction} and \textit{joint pose prediction}, respectively.
We consider only pairs of parts that form rigid joints, and leave full multi-part assembly and non-rigid joints to future work. We assume that object or part class labels and any form of human guidance are unavailable. We train only using the weak supervision provided by standard parametric CAD files without any manual human annotation such as canonical alignment.

\subsection{Input Representation}
\label{section:representation}
Our method takes a pair of parts in the B-Rep format (Figure~\ref{figure:representation}a), building upon a line of recent work~\cite{jayaraman2021uvnet, lambourne2021brepnet, willis2020fusion, xu2021zone} that utilizes the topology and geometry available within B-Rep CAD data. This approach enables us to make predictions over the exact entities used to define joints, rather than an intermediate representation such as a mesh or point cloud. Importantly, it allows us to frame the problem as a categorical one, by making predictions over the discrete set of B-Rep entities that contain ground truth information about the joint axis. Joints are commonly defined between \textit{both} B-Rep face and edge entities, e.g. a cylinder (face) can be constrained to another cylinder (face) or a circle (edge). To accommodate this, for each part we build a graph representation, $G(V, E)$, from the B-Rep topology where graph vertices $V$ are either B-Rep faces or edges, and graph edges $E$ are defined by adjacency (Figure~\ref{figure:representation}b). 

For graph vertex features we use information about individual B-Rep faces and edges readily available in the B-Rep data structure. For B-Rep faces, we use a one-hot vector for the surface type (plane, cylinder, etc.) and a flag indicating if the surface is reversed with respect to the face. For B-Rep edges, we use a one-hot vector for the curve type (line, circle, etc.), the length of the edge, and a Boolean flag indicating if the curve is reversed with respect to the edge. We evaluate the performance of these input features and others in Section~\ref{section:appendix_experiments} of the supplementary material. 

Finally, given the two graphs $G_1, G_2$ that we wish to assemble, with $n$ and $m$ vertices respectively, we form a third `joint connectivity graph' $G_j$ that densely connects the vertices between $G_1$ and $G_2$. $G_j$ has $n\times m$ edges and allows us to formulate a link prediction problem~\cite{liben2007link}, by identifying the connections between $G_1$ and $G_2$ that form a joint. $G_j$ can be easily visualized as an $n\times m$ matrix (Figure~\ref{figure:representation}c).

\begin{figure*}
    \begin{center}
        \includegraphics[width=\textwidth]{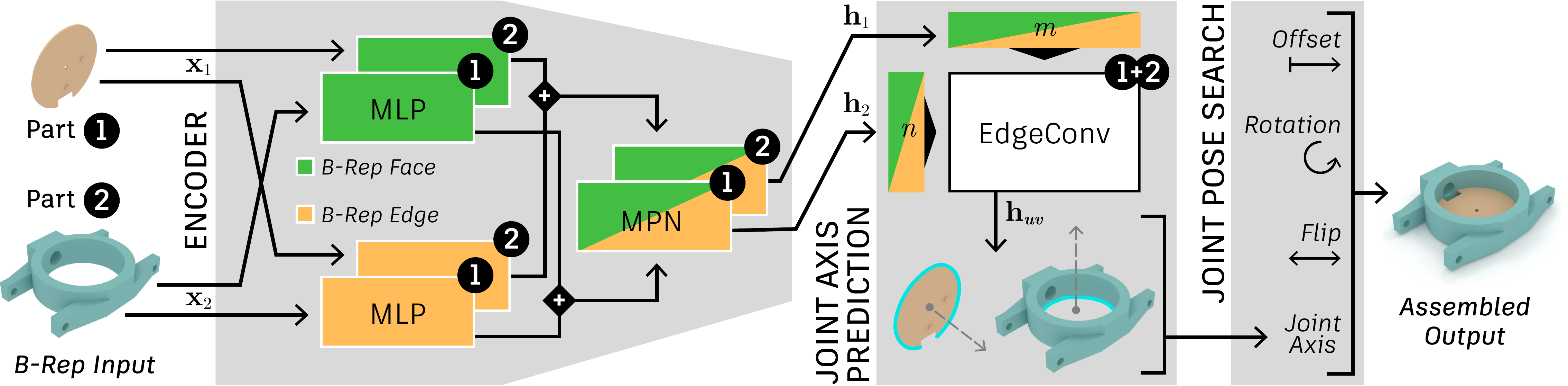}
        \caption{\textit{JoinABLe} architecture. Given two B-Rep parts in our graph representation,
        the vertex features from B-Rep faces (green) and edges (orange) pass through separate multi-layer perceptrons (MLP) before being concatenated together and passed through a message passing network (MPN). This yields local vertex embeddings representing each B-Rep entity in the two parts. Our joint axis prediction branch then performs edge convolution \textit{between} the two graphs to estimate the presence of joints over all possible pairs of connections. Finally, the joint parameters are discovered via search, with respect to the predicted joint axis, to complete the assembly.}
        \label{figure:network}
    \end{center}
\end{figure*}

\subsection{Weak Supervision from CAD Joints}
\label{section:weak-supervision}
A pair of parts in the B-Rep format have a finite number of faces and edges that can be paired to form a joint, specifically the $n\times m$ edges in $G_j$. Each ground truth joint results in a single positive label in the $n \times m$ prediction space and all remaining combinations are negative labels. For complex parts, such as mechanical gears that may contain thousands of discrete B-Rep entities, this results in an \textit{extreme} imbalance between positive and negative labels. 

The problem is further compounded by having only weak supervision available in standard parametric CAD files. This is due to several reasons: firstly, specifying joints between parts is time consuming and is often skipped by CAD designers; secondly, each CAD assembly is designed for a specific purpose, rather than to create an exhaustive set of assembly configurations. This weak supervision results in a positive and unlabeled (PU) learning problem~\cite{bekker2020learning} where the joints are known positive labels, but the remaining negative labels could be positive (i.e. an unseen but plausible joint) or negative (i.e. an implausible joint). To address the data imbalance and PU learning problem, we organize and augment our data using the following three techniques.

\textbf{Joint Consolidation}~~ To increase the number of positive labels, we consolidate joints between identical pairs of parts into \textit{joint sets}.  Figure~\ref{figure:example_designs}, right shows an example joint set where the same two parts are connected in multiple different ways. This approach allows us to present the network with a \textit{single} data sample, i.e. a joint set, that contains all known joints between a pair of parts. Importantly, joint consolidation avoids presenting the network with multiple contradictory data samples, where a negative label in one sample may be a positive label in another sample.
We provide additional implementation details about joint consolidation in Section~\ref{section:appendix_dataset} of the supplementary material.

\textbf{Joint Equivalents}~~ To further counter the extreme data imbalance, we identify and label `equivalent' entities that share the same joint axis as the ground truth. For example, if a circle is the labeled entity (highlighted in cyan in Figure~\ref{figure:representation}a), its neighbouring faces, such as the cylinder highlighted in pink, will be labeled as equivalent. 
These entities represent the same user-selected joint axis and only differ by the origin point that locates the joint axis in 3D space. 
As we consider a predicted joint axis to be correct if it is co-linear with the ground truth joint axis, we include equivalent labels during \textit{evaluation}. We perform an ablation study in Section~\ref{section:appendix_experiments} of the supplementary material to evaluate the contribution of equivalent labels.

\textbf{Unambiguous Evaluation Sets}~~ A challenge with PU Learning is establishing a `clean' test set to accurately measure network performance. Parts that have multiple plausible joints, such as a plate with multiple holes for fasteners, are problematic if only partial joint labels exist, leading to ambiguity at test time. 
We make a best effort to avoid positive unlabelled samples in the test and validation set by excluding geometrically similar, but unlabeled, `sibling' entities, e.g. the faces and edges of an unlabeled hole with the same size as a labelled hole.
We identify sibling entities by matching the entity type, area or length, and number of connected graph edges to the labeled entities.
In Section~\ref{section:appendix_experiments} of the supplementary material we study the effect of evaluating with sibling entities on a withheld test set that matches the original data distribution.

\subsection{JoinABLe Architecture}
Our overall architecture is shown in Figure~\ref{figure:network} and consists of an encoder module that outputs per-vertex embeddings for each B-Rep face and edge in our graph representation of the input parts. Using these embeddings we can predict a joint axis and then search for joint pose parameters.

\subsubsection{Encoder}
Our encoder neural network $f_\text{enc}$ is a Siamese-style network with shared weights for the two parts. It firstly creates graph vertex embeddings by passing the vertex features $\mathbf{x}_1$ and $\mathbf{x}_2$ from the two graphs, through two separate multi-layer perceptrons (MLP). One MLP is used for vertices representing B-Rep faces and another for those representing B-Rep edges; the resulting vertex embeddings are then concatenated together. We next perform message passing \textit{within} each part's graph using a two-layer Graph Attention Network v2 (GATv2)~\cite{brody2021attentive} to obtain the per-vertex embeddings $\mathbf{h}_1$ and $\mathbf{h}_2$ for both the graphs.
\begin{equation}
\mathbf{h}_1 = f_\text{enc}(\mathbf{x}_1, G_1), \quad
\mathbf{h}_2 = f_\text{enc}(\mathbf{x}_2, G_2). \\
\label{eq:vertex_embeddings}
\end{equation}
The idea here is to extract local features within each part that consider each B-Rep entity and its neighborhood.

\subsubsection{Joint Axis Prediction}
When creating a joint, a key piece of design intent is the definition of a joint axis by which two parts can be aligned and constrained to one another. The joint axis forms the basis for the degrees of freedom to be defined and enables downstream tasks such as assembly, part mobility, and animation. We formulate joint axis prediction as a link prediction problem, where the goal is to correctly identify a connection between $G_1$ and $G_2$ that aligns the two parts along a ground truth joint axis.
This is done by aggregating information \textit{between} parts using an edge convolution along the edges of $G_j$. The node features $\mathbf{x}_1$ and $\mathbf{x}_2$ from graphs $G_1$ and $G_2$, are passed through our shared encoder network $f_\text{enc}$ to get 384-dimensional embeddings $\mathbf{h}_1$ and $\mathbf{h}_2$ (Eq.~\ref{eq:vertex_embeddings}).
Then for each edge $(u, v)$ in the graph $G_j$ which densely connects $G_1$ and $G_2$, we predict a logit indicating the presence of a joint:
\begin{equation}
\mathbf{h}_{uv} = \phi(\mathbf{h}_u \oplus \mathbf{h}_v),
\label{eq:edge_prediction}
\end{equation}
where $\phi : \mathbb{R}^{768} \mapsto \mathbb{R}$ is a 3-layer MLP, $\oplus$ is the concatenation operator and $\mathbf{h}_u$ and $\mathbf{h}_v$ are gathered from $\mathbf{h}_1$ and $\mathbf{h}_2$ based on the source and target vertices for each edge in $G_j$.

We train the network with a loss function that has two terms.
The first term $\mathcal{L}_{\text{CE}}$ is the cross-entropy between the edge predictions $\mathbf{h}_{uv}$ and the ground truth edge labels $\mathbf{j}_{uv} \in \{0, 1\}$ normalized into a probability distribution $\widehat{\mathbf{j}}_{uv}$.
\begin{equation}
\begin{aligned}
\widehat{\mathbf{h}}_{uv} &= \text{softmax}_{\text{all}}(\mathbf{h}_{uv}), \\
\mathcal{L}_{\text{CE}} &= \text{CE}\left(\widehat{\mathbf{j}}_{uv}, {\widehat{\mathbf{h}}_{uv}}\right).
\end{aligned}
\label{eq:loss_ce}
\end{equation}
Here the subscript in the softmax operation indicates that it is applied over all edges in $G_j$, and $\text{CE}(\mathbf{p}, \mathbf{q}) = -\sum_i \mathbf{p}_i \log\mathbf{q}_i$. This loss encourages true joints to have higher values while simultaneously suppressing non-joints. We observe this is sub-optimal due to the sparsity of positive labels, where $\mathcal{L}_{\text{CE}}$ is summed over a large number of terms. To better focus the loss term so that the joints are better contrasted against more likely non-joints, we use a symmetric cross entropy loss $\mathcal{L}_{\text{Sym}}$ as the second term in the loss function.
\begin{equation}
\begin{aligned}
\widehat{\mathbf{h}}_{\text{row}} = \text{softmax}_{\text{row}}(\mathbf{h}_{\text{2D}})&, \quad
\widehat{\mathbf{h}}_{\text{col}} = \text{softmax}_{\text{col}}(\mathbf{h}_{\text{2D}}),\\
\mathcal{L}_{\text{Sym}} = \text{CE}(\widehat{\mathbf{j}}_{\text{2D}}, \widehat{\mathbf{h}}_{\text{row}}) &+ \text{CE}(\widehat{\mathbf{j}}_{\text{2D}}, \widehat{\mathbf{h}}_{\text{col}}).
\end{aligned}
\label{eq:loss_sym}
\end{equation}
Here the subscript of the softmax indicates that it is taken over a single axis, and the 2D subscript instead of $uv$ indicates that the predictions and ground truth labels on the edges of $G_j$ are reshaped into $n\times m$ matrices.

\subsubsection{Joint Pose Search}
\label{section:joint-pose-search}
The B-Rep entities predicted by our network allow us to query the ground truth B-Rep data to obtain a joint axis prediction for each part. Once these axes are aligned together, three secondary parameters define a rigid joint and can be used for joint pose prediction. An \textit{offset} distance along the joint axis, \textit{rotation} about the joint axis, and a \textit{flip} parameter to reverse the joint axis direction. We find these parameters using a neurally guided search that allows us to enumerate over the top-$k$ joint axis predictions and directly consider interaction between both parts. To evaluate a candidate joint configuration, we propose a cost function $\mathcal{C}_{\text{joint}} = \mathcal{C}_{\text{overlap}} + \lambda \mathcal{C}_{\text{contact}}$ that considers two general criteria for well defined joints: overlap volume and contact area between parts, formulated as:
\begin{equation}
\mathcal{C}_{\text{overlap}} = \frac{\volume_{1 \cap 2}}{\min{(\volume_1, \volume_2)}} , \quad
\mathcal{C}_{\text{contact}} = \frac{A_{1 \cap 2}}{\min{(A_1, A_2)}}.
\label{eq:overlap}
\end{equation}
Here, $\volume_1$ and $\volume_2$ are the volume of the two parts, and $\volume_{1 \cap 2}$ represents their overlap volume. Similarly, $A_1$ and $A_2$ are the surface area of the two parts, and their contact area is $A_{1 \cap 2}$. Intuitively, for the two parts to align closely to each other, minimizing the cost function should encourage a larger contact area while penalizing the overlap volume to prevent penetration. Therefore, we let $\lambda = -10$ if ${\mathcal{C}_{\text{overlap}} < 0.1}$. Otherwise, we set $\lambda = 0$ to increase the overlap penalty. Given this cost function, we search for the optimal joint pose using the Nelder-Mead algorithm~\cite{nelder1965simplex} as a standard derivative-free optimization.

\section{Dataset}
\label{section:dataset}

To evaluate the performance of our method we create the \FData{}, derived from designs created in Autodesk Fusion 360 and submitted to the publicly available Autodesk Online Gallery~\cite{autodeskOnlineGallery}. The dataset consists of two inter-related sets of data, \textit{Assembly Data}, containing 8,251 assemblies with 154,468 separate parts, and \textit{Joint Data} containing 32,148 joints defined between 23,029 different parts.
The data and supporting code are publicly available on GitHub\footnote{\href{https://github.com/AutodeskAILab/Fusion360GalleryDataset}{https://github.com/AutodeskAILab/Fusion360GalleryDataset}} with a license allowing non-commercial research.
We now describe the joint data used in our experiments and provide information on the overall dataset in Section~\ref{section:appendix_dataset} of the supplementary material.

Figure~\ref{figure:example_designs}, left shows an overview of the joint data in our dataset. We consider a data sample to be a joint set, such as shown in Figure~\ref{figure:example_designs}, right, containing a pair of parts with one or more joints defined between them. 
The user-selected B-Rep faces and edges form the ground truth labels together with the joint axis and pose information of each joint. We provide an approximate 70/10/10/10\% data split, for the training, validation, test, and original distribution test sets respectively. The validation and test sets do not include samples with potentially ambiguous sibling entities, while the original distribution test set does.

\begin{figure}
    \begin{center}
        \includegraphics[width=\columnwidth]{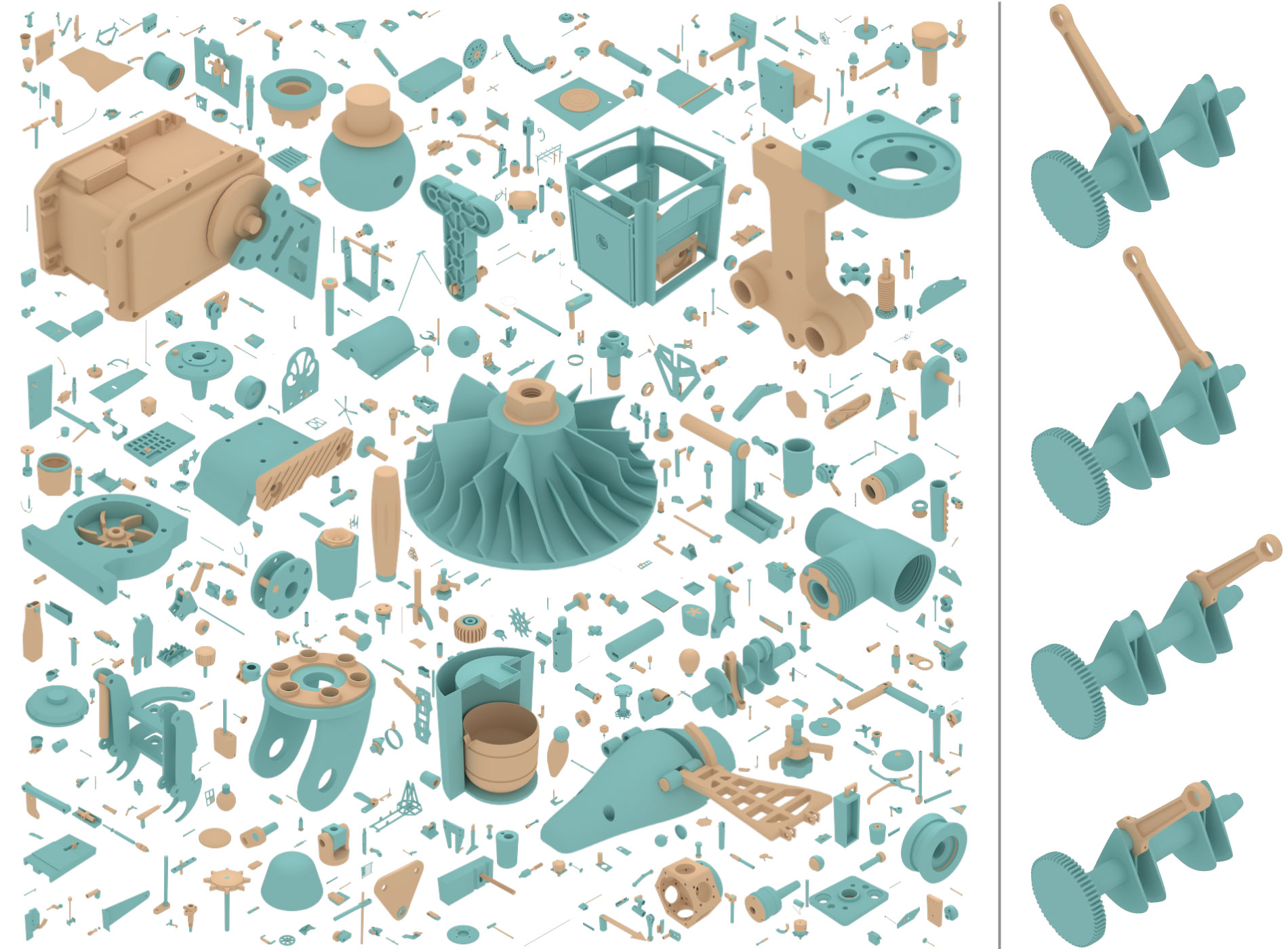}
        \caption{An overview of joint data from the \FData{} (left). Each sample consists of a unique pair of parts with \textit{one or more} joints defining how they are locally constrained and positioned together (right).}
        \label{figure:example_designs}
    \end{center}
\end{figure}

\section{Experiments}
\label{section:experiments}
In this section we perform experiments to qualitatively and quantitatively evaluate our method on two tasks: joint axis prediction and joint pose prediction. We examine how our method compares with a human CAD expert and other methods from the literature.
A key criteria for evaluating performance is to gauge how the network performs in scenarios that traditional algorithms find challenging. One such scenario involves designs that do not contain connections between cylindrical shafts and holes, such as a bolt and a hole similar to Figure \ref{figure:representation}a. Commercial products exist which infer joints of this type by searching for fasteners and holes with similar radii \cite{SolidworksSmartFasteners}.
In our dataset we see that 82\% of data samples contain holes and 47.5\% of joints constrain circular or cylindrical entities on one part to a hole on the opposing part.
In our experiments we report results that gauge the ability of our approach to correctly infer joints, both in the simple \textit{Hole} case and the more complex \textit{No Hole} case. Details of experiment procedures are provided in Section~\ref{section:appendix_experiments} of the supplementary material.

\subsection{Human CAD Expert Baseline}
\label{section:human-cad-expert-baseline}
Understanding how a human CAD expert performs in a similar setting is important to gauge the efficacy of each method. We conduct a study to establish a human baseline by recruiting a CAD expert, who works on commercial CAD design, and ask them to assemble pairs of parts from our dataset with a known ground truth joint. We use 100 data samples picked randomly from a distribution excluding the potentially ambiguous sibling entities.
We randomly rotate and translate each part and conduct the study using Fusion 360.
We compare the joint axis created by the CAD expert with the ground truth. We find that the CAD expert results match the ground truth 80\% of the time.
This shows that determining how two isolated parts should be assembled is challenging for CAD experts without the valuable context provided by the object assembly. We provide additional details in Section~\ref{section:appendix_experiments} of the supplementary material.

\begin{table}
    \centering
    \small
    \begin{tabular}{l|cccc}
        \toprule
        & \textbf{All} & \textbf{Hole} & \textbf{No Hole} & \textbf{Param.}  \\
        & \textbf{Acc.\%} $\uparrow$ & \textbf{Acc.\%} $\uparrow$ & \textbf{Acc.\%} $\uparrow$ & \textbf{\#} $\downarrow$  \\
        \midrule
        \textbf{Ours}   & \textbf{79.53}     & \textbf{80.15}     & \textbf{76.59}     & \textbf{1.3M}  \\ 
        B-Dense         & 10.59     & 10.36     & 10.59     & 3.2M  \\ 
        B-Discrete      & ~4.28     & ~4.18    & ~4.79     & 4.0M  \\
        B-Grid          & 65.21     & 65.09     & 65.81     & 3.1M  \\ 
        B-Heuristic     & 71.39     & 72.74     & 64.97     & -     \\
        B-Random        & 21.55     & 21.92     & 23.29     & -     \\
        \midrule
        Human           & 80.00         & -         & -         & -     \\
        \bottomrule
    \end{tabular}
    \caption{Joint axis prediction accuracy results are shown for all data samples in the test set (All), the subset of data samples with holes (Hole) and without holes (No Hole). The number of network parameters is also shown (Param.). Finally, results from a human CAD expert on 100 test samples are shown.}
    \label{table:joint-axis-prediction}
\end{table}

\subsection{Joint Axis Prediction}
\label{section:joint-axis-experiment}

Although there are no previous works that address the exact same setting as ours, we adapt several related methods to compare with our approach.

\textbf{Point Cloud Baselines}~~ We adapt two point cloud based methods designed to predict a joint axis for part mobility. For each baseline we use a common architecture, based on a PointNet++~\cite{qi2017pointnet++} encoder, and adapt the decoder strategy and loss functions from related work.
\textbf{B-Dense} follows Li et al.~\cite{li2020category} to densely regress a joint origin projection vector, projection distance, and joint direction for each point in the point cloud.
\textbf{B-Discrete} follows Shape2Motion~\cite{wang2019shape2motion} and uses a hybrid of discrete classification and regression to predict the joint origin point and direction vector.

\textbf{B-Rep Baselines}~~ We compare our method against several baseline methods that take B-Rep graphs as input.
\textbf{B-Grid} follows UV-Net~\cite{jayaraman2021uvnet} and uses grid features (points, normals, trimming mask, and tangents) sampled on B-Rep faces and edges together with a CNN encoder. We use the same graph topology, prediction head, and loss as our network.
\textbf{B-Heuristic} uses a rule-based approach that operates on B-Rep graphs and assigns a score to each B-Rep entity. Higher scores are assigned to entities that are similar, based on the entity type, area, and length information, and that match the training data distribution of entity type pairings. For cylinders and circles, the radius of the entity is also employed.  A higher score is given to entity pairs where the radii match to within 5\%. 
\textbf{B-Random} makes random predictions over all B-Rep entities and represents the lower bound for B-Rep performance.

Table~\ref{table:joint-axis-prediction} shows results for the joint axis prediction task on the test set. We report the accuracy of regression based approaches by considering a joint axis prediction to be a `hit' if it is collinear within a distance and angular threshold of 5\%.
For classification based approaches we report the top-1 accuracy. We also report accuracy for the subset of data samples that have holes (\textit{Hole}) and those that do not (\textit{No Hole}).  Recall that traditional algorithms are good at working with the special case of matching fasteners to holes. 
We observe that the performance gap between our approach and the next highest performing B-Heuristic approach is 8.14\%, however this widens to 11.62\% for the important \textit{No Hole} subset where traditional algorithms are known to struggle.
We find that the B-Rep based approaches outperform those based on point clouds while also using fewer parameters. Although point cloud approaches perform well with axis aligned parts from the same object class~\cite{wang2019shape2motion}, our results show that real world data is significantly more challenging. Finally we note that our approach is within 0.5\% of the performance of a human CAD expert. We provide additional details in Section~\ref{section:appendix_experiments} of the supplementary material.

\begin{table}
    \centering
    \small
    \begin{tabular}{l|cccc}
        \toprule
        & \textbf{All} & \textbf{Hole} & \textbf{No Hole} & \textbf{Param.} \\
        & \textbf{CD} $\downarrow$ & \textbf{CD} $\downarrow$ & \textbf{CD}$\downarrow$ & \textbf{\#} $\downarrow$ \\
        \midrule
        \textbf{Ours + Search} & \textbf{0.0580} & \textbf{0.0570} & \textbf{0.0628} & \textbf{1.3M} \\ 
        \textbf{Ours} & 0.0627 & 0.0624 & 0.0657 & \textbf{1.3M} \\ 
        B-Pose & 0.0700 & 0.0693 & 0.0730 & 2.3M \\ 
        \bottomrule
    \end{tabular}
    \caption{Joint pose prediction results using average chamfer distance (CD) where lower is better. We show results for all samples in the test set (All), and the subset of data samples with holes (Hole) and without holes (No Hole). The number of network parameters is also shown (Param.)}
    \label{table:joint-pose-prediction}
\end{table}

\begin{figure*}
    \begin{center}
        \includegraphics[width=\textwidth]{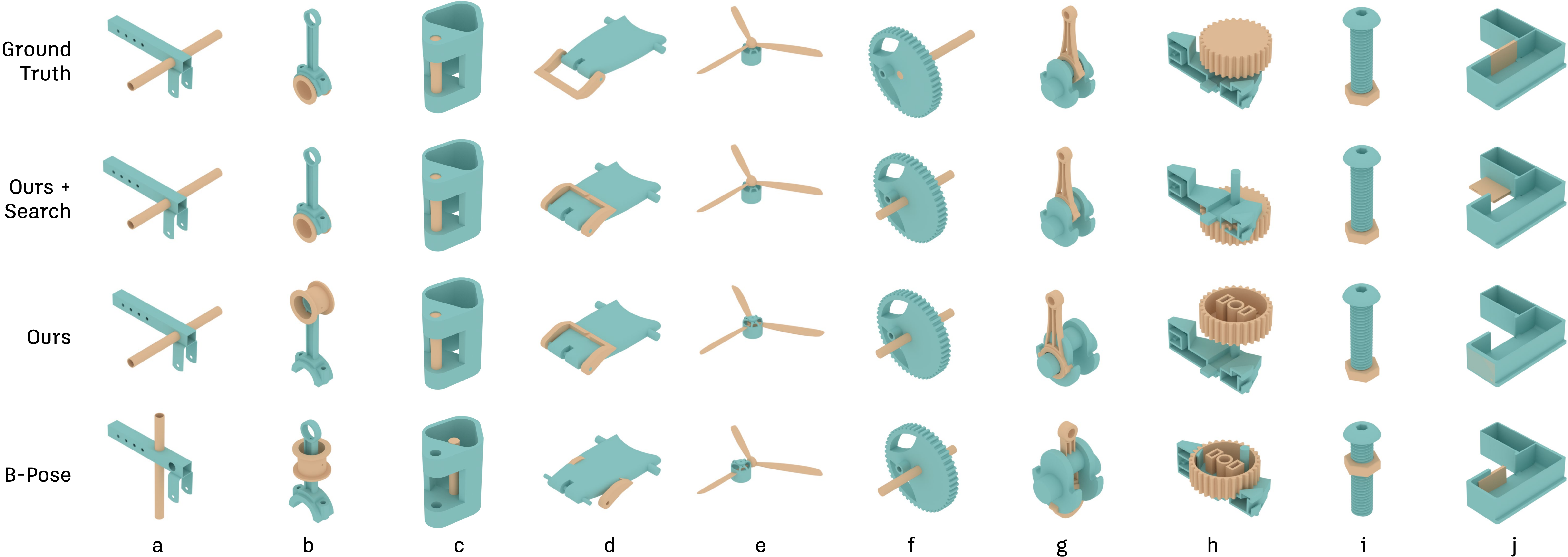}
        \caption{Qualitative comparison of joint pose prediction results comparing our method, with and without search, with the B-Pose baseline.}
        \label{figure:qualitative-results}
    \end{center}
\end{figure*}

\subsection{Joint Pose Prediction}
\label{section:joint-pose-experiment}
For the joint pose prediction task we again adapt a baseline method from the literature to our setting. \textbf{B-Pose} follows Huang et al.~\cite{huang2020generative} to regress a translation point and rotation quaternion using a combination of L2 and chamfer distance (CD) loss terms. Although a parametric joint is not created, B-Pose represents a common approach used with top-down assembly. We evaluate the performance of our method in two different configurations. \textbf{Ours} uses the joint axes derived from network predictions to align the two parts together without an offset, rotation, or flip. \textbf{Ours + Search} additionally performs joint pose search over the top 50 predictions to find suitable offset, rotation, and flip parameters.

Table~\ref{table:joint-pose-prediction} shows results for the joint pose prediction task. We record the minimum CD calculated between the $\ge 1$ ground truth joints from the joint set and the predicted assembly. We then report the average CD across all samples in the test set. 
We find that using our network predictions alone (Ours) can better match the ground truth when compared with the B-Pose baseline. Introducing search (Ours + Search) can help resolve areas of overlap (Figure~\ref{figure:qualitative-results}e) and in some cases resolve incorrect axis predictions (Figure~\ref{figure:qualitative-results}b,g). 
It is important to note that the ground truth data only contains a finite set of discrete states (e.g. door open, door closed) rather than continuous states (e.g. door \textit{opening}) that may also be valid. For example, our predictions for the belt buckle in Figure~\ref{figure:qualitative-results}d do not match the ground truth state but appear plausible. As such, CD should be considered an approximate metric for comparing the relative performance of each method. We provide further qualitative results in Section~\ref{section:appendix_experiments} of the supplementary material.

\section{Discussion}
\label{section:discussion}

\begin{figure}[b]
    \begin{center}
        \includegraphics[width=\linewidth]{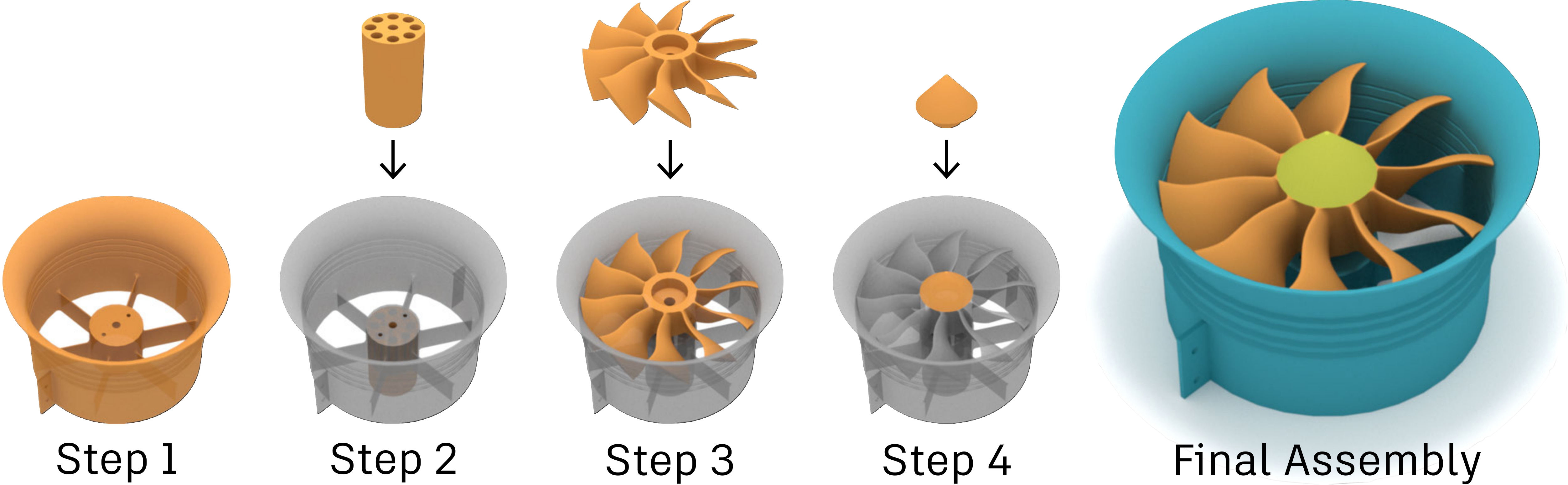}
        \caption{Multi-part assembly demonstration. Parts are aligned sequentially from a given assembly sequence using our joint axis prediction network and pose search. }
        \label{figure:multi-part-assembly}
    \end{center}
\end{figure}

\textbf{Future Applications}~~
Our joint axis prediction network and search approach can serve as fundamental building blocks for a number of applications. One such application is the automated assembly of multiple parts in a design. As a preliminary demonstration we assemble a multi-part design given only the individual parts and the sequence of part pairs derived from our assembly dataset. We amend our search strategy to minimize the overlap volume between the new part and the partially assembled design at each assembly step and maximize the contact area between them using a similar cost function. Figure~\ref{figure:multi-part-assembly} shows an example sequence of parts that are assembled correctly in a bottom-up fashion. We provide further details in Section~\ref{section:appendix_future_work} of the supplementary material.

\textbf{Limitations}~~
A bottom-up approach to assembly may be limited when scaling to large assemblies where global composition is important. 
Reliance on B-Rep CAD data is another limitation of the current work. Although data availability is improving~\cite{koch2019abc, wu2021deepcad, jones2021sb}, our method has not been tested beyond mechanical CAD data. 
Finally, our network does not leverage geometric loss terms that may help with avoiding undesirable overlap between parts and generalize to predicting other joint parameters. 
\section{Conclusion}
Our long-term motivation is to enable assembly-aware design tools, capable of suggesting and automatically placing parts. Such a system could enable greater reuse of existing physical components in new designs and potentially reduce the cost and environmental impact associated with manufacturing and associated supply chains~\cite{kerr2001eco}. Understanding how parts are assembled is also critical for robotic assembly and disassembly. CAD-informed robotic disassembly systems may enhance our ability to reuse and recycle components~\cite{marconi2019feasibility, li2019multi, liu2019human, chang2017approaches}. In this work we have begun the first steps to address these challenges by learning the bottom-up assembly of parametric CAD joints. Our results show the promise of learning-based methods to approach the performance of human CAD experts, and with the publication of our dataset we hope to further aid future research.

\bibliography{main}
\bibliographystyle{plain}

\clearpage
\setcounter{section}{0}
\renewcommand\thesection{\Alph{section}}
\renewcommand\thesubsection{\thesection.\arabic{subsection}}

\section{Supplementary Material}

\subsection{Dataset}
\label{section:appendix_dataset}
The \FData{} consists of two inter-related sets of data with assembly data and joint data. The data and supporting code are publicly available on GitHub\footnote{\href{https://github.com/AutodeskAILab/Fusion360GalleryDataset}{https://github.com/AutodeskAILab/Fusion360GalleryDataset}} with a license allowing non-commercial research. We now outline related datasets, data processing steps, documentation, and statistics about the dataset.

\subsubsection{Related Datasets}
\label{section:appendix_dataset_related}
The ABC dataset~\cite{koch2019abc} provides 1 million CAD assemblies in the B-Rep format, containing valuable analytic representations of surfaces and curves. However, each assembly contains individual part files positioned in global space without the critical joint information describing how parts are connected and constrained together. Other datasets providing designs in B-Rep format only include part geometry and lack assembly data entirely~\cite{lambourne2021brepnet,jayaraman2021uvnet,willis2020fusion,wu2021deepcad}.

Recently a number of datasets have extended existing 3D shape datasets, such as ShapeNet~\cite{chang2015shapenet} and PartNet~\cite{mo2019partnet}, with additional human annotated labels for part mobility~\cite{xiang2020sapien, wang2019shape2motion, yanRPMNet19, hu2017learning}. The resulting synthetic assemblies have joint type and range of motion information included.
Our dataset differs from these datasests in several ways: 
\begin{enumerate}
    \item We provide CAD assemblies that are more representative of real world design, including detailed design such as fasteners.
    \item Joint connectivity and component hierarchy are defined by the designers themselves rather than human annotators.
    \item Joints between parts are defined by discrete designer-selected entities, such as B-Rep faces and edges, making them well suited to learning tasks.
    \item We provide B-Rep and mesh representations together with extensive metadata.
\end{enumerate}
We believe it is critical to leverage and learn from the rich sources of information available inside of existing CAD models. Rather than rely on extensive human annotation, our dataset exploits the knowledge of domain experts on how shapes are defined and assembled using industrial CAD modeling software.

Concurrent to our work, AutoMate~\cite{jones2021sb} announced a similar dataset to ours with a larger number of overall designs. Based on the description of the dataset in~\cite{jones2021sb}, we note several advantages that may be helpful for users of our dataset: 
\begin{enumerate}
    \item We preserve the sub-assembly hierarchy for all assemblies, allowing for the creation of multiple hierarchical representations via contacts, joints, or the designer-defined assembly tree.
    \item We provide assembly metadata listing the contact surfaces, hole types, materials, and various user specified tags for each design. 
    \item We consolidate joints across the dataset, meaning that for two given parts in our joint data we list all known ground truth joints between them. This avoids presenting the network with contradictory labels during training, where multiple different versions of a positively labelled joint configuration could exist across data samples.
    \item We provide a `clean' test and validation set for joint prediction by removing potential positive unlabeled samples.
\end{enumerate}
Ultimately we believe both datasets will be helpful to cross-validate using designs created in different CAD software and increase the robustness of learning-based methods. Future updates to the AutoMate dataset may include a number of the capabilities listed above.

\begin{figure*}
    \begin{center}
        \includegraphics[width=0.95\textwidth]{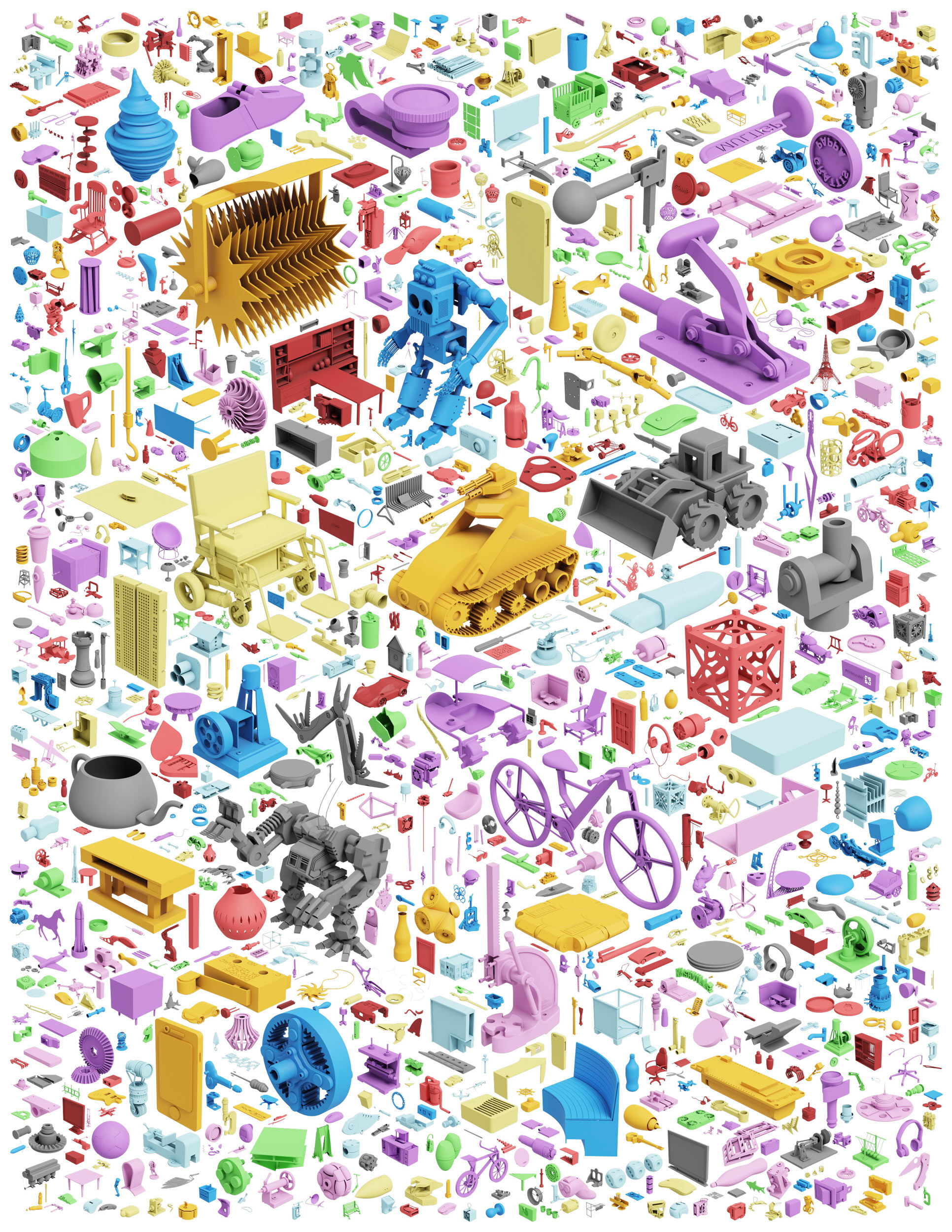}
        \caption{An overview of assemblies in the \FData{}.}
        \label{figure:assembly_mosaic}
    \end{center}
\end{figure*}

\subsubsection{Data Processing}
We create the \FData{} from approximately 20,000 designs in the native Fusion 360 .f3d CAD file format. We use the Fusion 360 Python API\footnote{\href{https://help.autodesk.com/view/fusion360/ENU/?guid=GUID-7B5A90C8-E94C-48DA-B16B-430729B734DC}{https://help.autodesk.com/view/fusion360/ENU/}} to parse the native .f3d files into JSON format text files containing the main CAD parameter information, and geometry files in both B-Rep and mesh format. We use separate pipelines to process the assembly and joint data. After the data has been extracted we rebuild each design and compare it with the original to ensure data validity. Failure cases and any duplicate designs, are not included in the dataset. For the assembly data, we consider a design a duplicate when there is an exact match in all of the following: body count, occurrence count, component count, joint count, contacts count, hole count, surface area to one decimal point, and volume to one decimal point. This process allows us to remove duplicate designs from the dataset that have been uploaded multiple times. Using this process we identify and remove approximately 840 designs from the assembly data. For the joint data, we handle duplicates using the process of joint consolidation described in Section~\ref{section:joint_dataset}.

\subsubsection{Assembly Data}
\label{section:assembly_dataset}

\begin{figure}
    \begin{center}
        \includegraphics[width=\columnwidth]{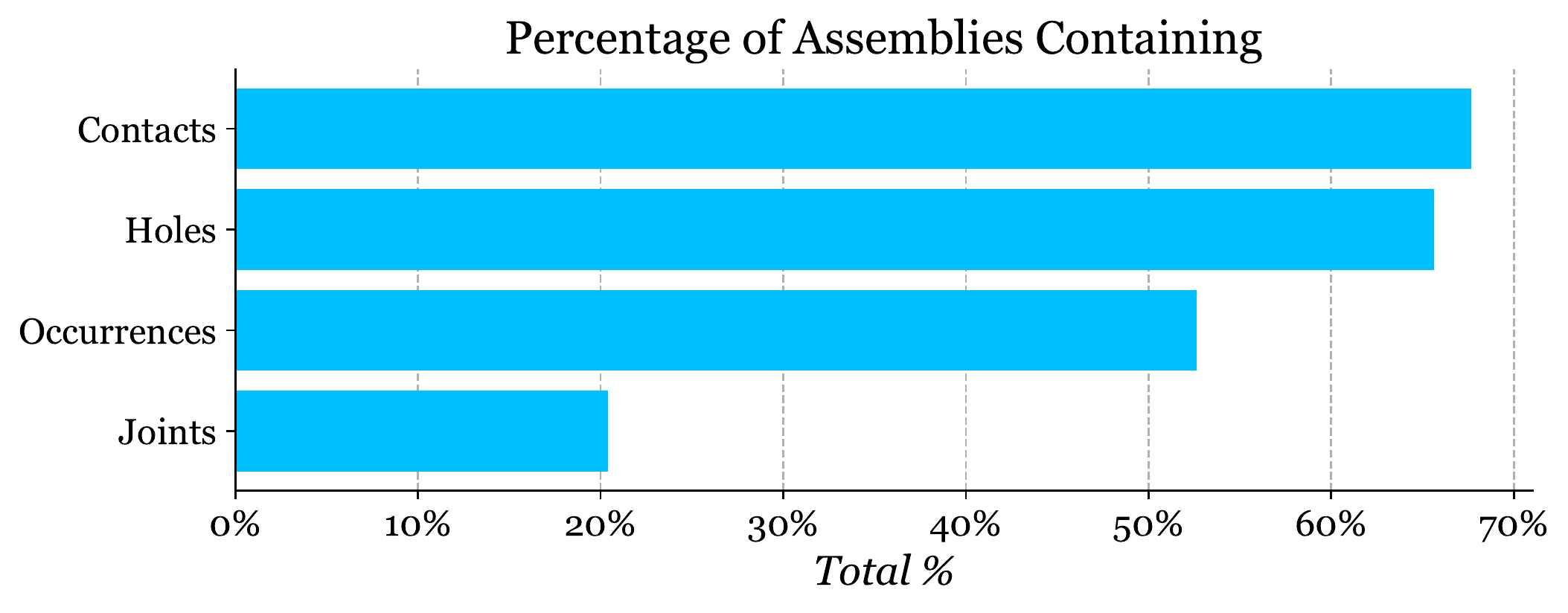}
        \caption{The percentage of assemblies containing contacts, holes, occurrences (instances of components), and joints.}
        \label{figure:assembly_percentages}
    \end{center}
\end{figure}

In mechanical CAD software, \textit{assemblies} are collections of \textit{parts}, represented as 3D shapes, that together represent an overall design, or object. In our assembly data we filter out designs that contain only a single part, leaving us with 8,251 assemblies containing a total of 154,468 separate parts. A random sampling of these assemblies is shown in Figure~\ref{figure:assembly_mosaic}. 
Individual parts can be grouped together into \textit{components} that represent reusable parts of a design, for example a single screw or a sub-assembly containing multiple components. Components can be positioned in global coordinates or constrained to one another using \textit{joints}. 
\textit{Contacts} exist when faces of parts in the assembly touch, within a tolerance, with the faces of other parts. Our assembly data is provided in a JSON text format containing the top-level data elements listed in Table~\ref{table:assembly-features}. 
The structure and representation of the data follows the Fusion 360 API.  We use universally unique identifiers (UUID) to cross reference elements with the JSON data.
In the following paragraphs we describe the top-level elements in the data with more detail. 
Figure~\ref{figure:assembly_percentages} lists the percentage of assemblies containing several of the top-level data elements described in Table~\ref{table:assembly-features}.

\begin{figure*}
    \begin{center}
        \includegraphics[width=\textwidth]{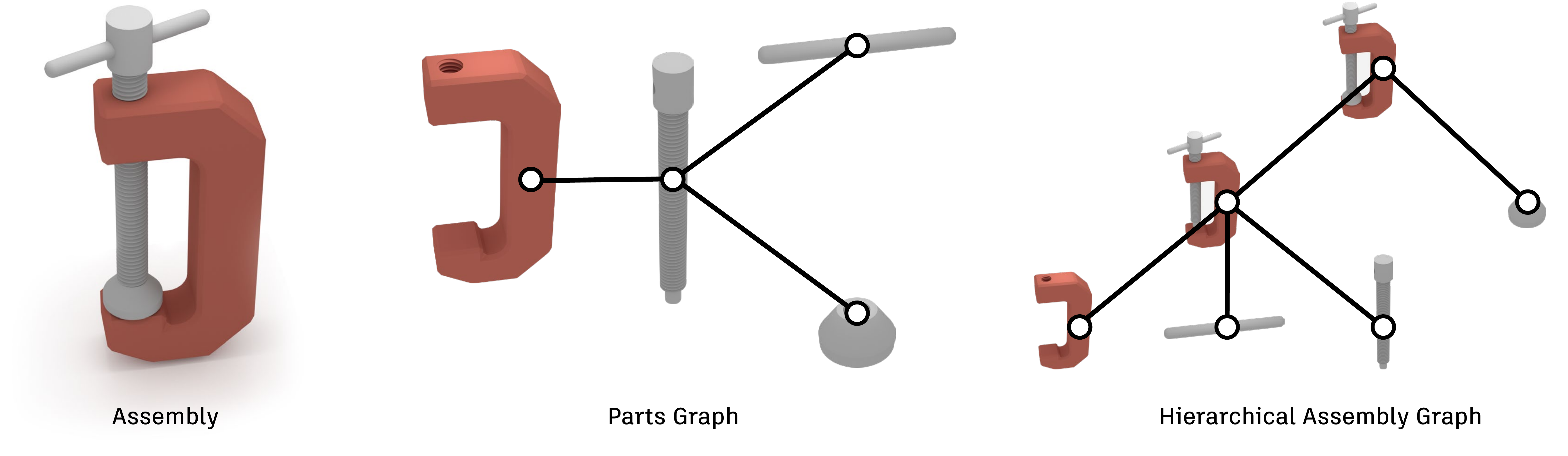}
        \caption{An example assembly (left) represented as a parts graph (middle) built from contact information, and as a hierarchical assembly graph (right) built from the designer-defined assembly tree.}
        \label{figure:example_assembly_graph}
    \end{center}
\end{figure*}

\begin{table}
    \centering
    \renewcommand{\arraystretch}{1.6} 
    \small
    \begin{tabular}{r|p{5.8cm}}
        \toprule
        \textbf{Element} & \textbf{Description}\\
        \midrule
        \textbf{Root} & The root component of the design as defined by the designer.\\
        \textbf{Components} & Components containing bodies or other components to form sub-assemblies.\\
        \textbf{Bodies} & The underlying 3D shape geometry in the B-Rep format. \\
        \textbf{Occurrences} & Instances of components, referencing the parent component with instance properties such as location, orientation, and visibility. \\
        \textbf{Tree} & The designer-defined hierarchy of occurrences in the design. Often used to organize sub-assembles into a meaningfully hierarchy. \\
        \textbf{Joints} & Constraints defining the relative pose and degrees of freedom (DOF) between a pair of occurrences. \\
        \textbf{Contacts} & Faces that are in contact between different bodies. \\
        \textbf{Properties} & Statistical information and metadata about the overall assembly. \\
        \textbf{Holes} & A list of hole features with information about the type of hole, size, direction, and location. \\
        \bottomrule
    \end{tabular}
    \caption{Descriptions of the top-level elements provided in our assembly data.}
    \label{table:assembly-features}
\end{table}

\paragraph{Root} The root of the assembly refers to the root node from which the hierarchical assembly graph can be constructed, as shown in Figure~\ref{figure:example_assembly_graph}, right. The root links to the component UUID that the designer specified to be at the top of the tree, and it also links to any bodies that might be contained by the root.

\paragraph{Components} Components are the building blocks that make up assemblies. Each component contains one or more bodies, a name, a part number and is assigned a UUID. Further information on components can be found in the Fusion 360 API documentation for the \href{https://help.autodesk.com/cloudhelp/ENU/Fusion-360-API/files/Component.htm}{\code{Component}} class.

\begin{figure}
    \begin{center}
        \includegraphics[width=\columnwidth]{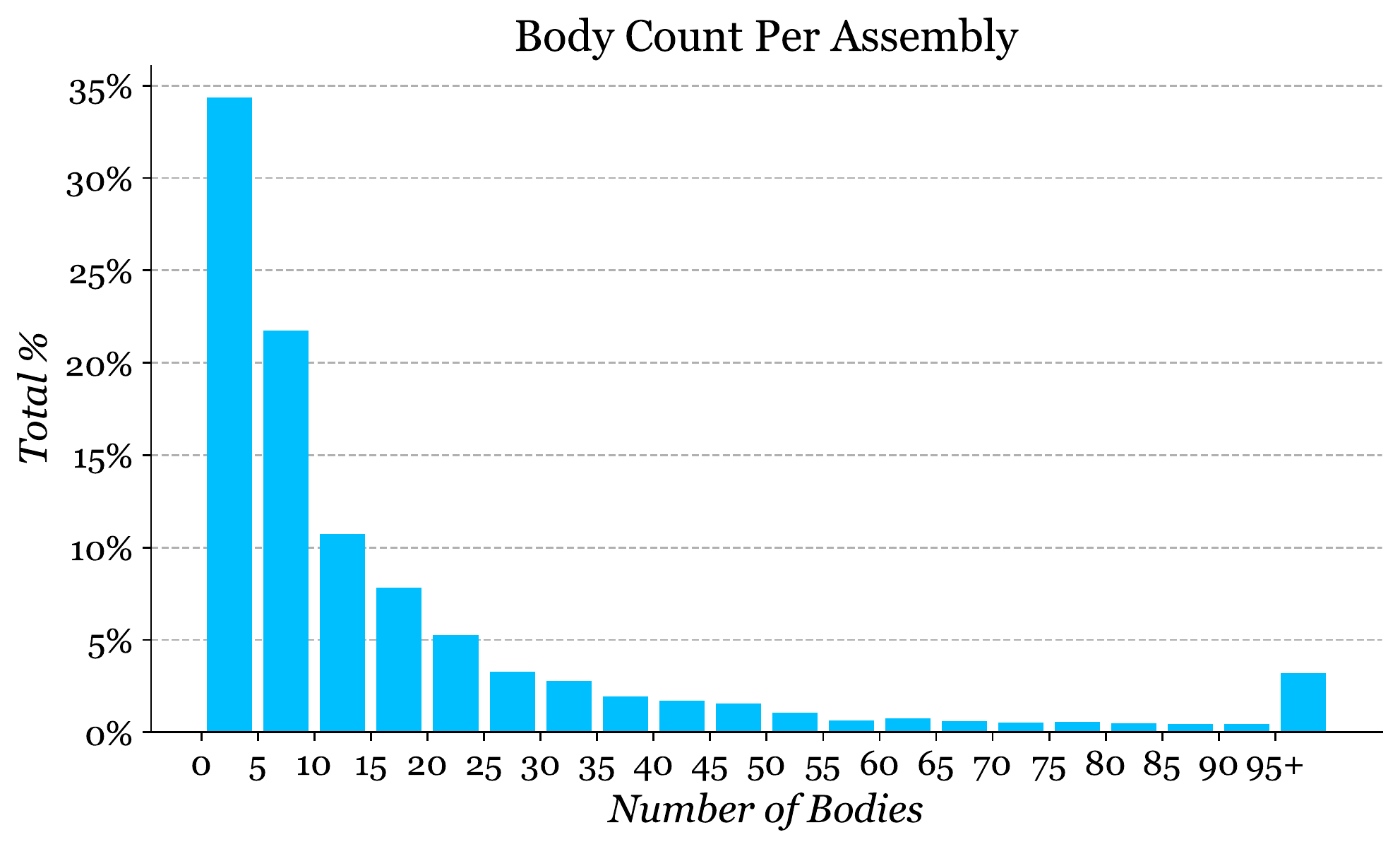}
        \caption{The number of bodies (separate 3D shapes) in an assembly, shown as a distribution across the assembly data.}
        \label{figure:assembly_body_count}
    \end{center}
\end{figure}

\paragraph{Bodies} Bodies are the geometric elements, represented as B-Reps, that make up components. The geometric data of each body is included in the dataset as described in Section~\ref{section:assembly_data_geometry_format}. Each body is assigned a UUID, and contains a name, physical properties, appearance, material, as well as the file names of the corresponding B-Rep, mesh, and image files. The physical properties of the body include the center of mass, area, volume, density, and mass. The appearance of the body refers to the material used for visual appearance, such as rendering, and contains the UUID and name of the user-assigned appearance. The material of the body, on the other hand, refers to the physical material from which the physical properties of the body are derived, such as the weight and density, and contains the UUID and name of the  material. Figure~\ref{figure:assembly_body_count} shows the number of bodies in an assembly as a distribution across the dataset. Further information on bodies can be found in the Fusion 360 API documentation for the \href{https://help.autodesk.com/cloudhelp/ENU/Fusion-360-API/files/BRepBody.htm}{\code{BRepBody}} class.

\begin{figure}
    \begin{center}
        \includegraphics[width=\columnwidth]{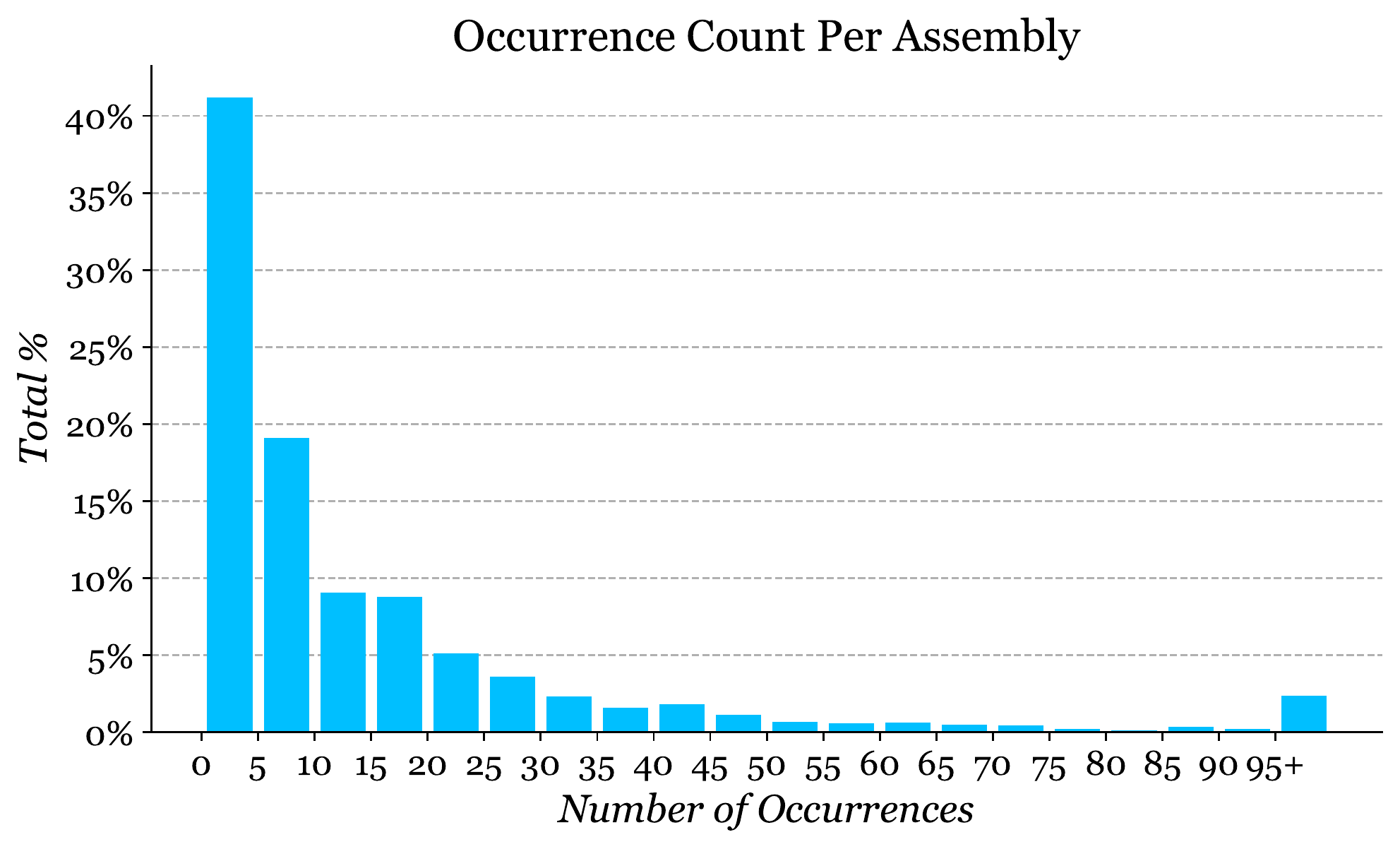}
        \caption{The number of occurrences (instances of a component) in an assembly, shown as a distribution across the assembly data excluding assemblies without occurrences.}
        \label{figure:assembly_occurrence_count}
    \end{center}
\end{figure}

\paragraph{Occurrences} Occurrences are instances of components that can have independent parameters applied, such as visibility, location, and orientation, while maintaining the same geometry as their parent component. An occurrence is to component, as Object is to Class in object oriented programming. Occurrences are given a UUID and link to a parent component. The flag \code{is\textunderscore grounded} indicates whether the user locked the position of the occurrence, preventing further movements from happening via mouse-dragging in the Fusion 360 UI. The flag \code{is\textunderscore visible} indicates whether the occurrence was displayed or not in the UI. Each occurrence also has information about the physical properties (aggregating the center of mass, area, volume, density, and mass of all included components and bodies), as well as the transformation matrix necessary to orient the occurrence within the global space. Figure~\ref{figure:assembly_occurrence_count} shows the number of occurrences in an assembly as a distribution across the dataset. Further information on occurrences can be found in the Fusion 360 API documentation for the \href{https://help.autodesk.com/cloudhelp/ENU/Fusion-360-API/files/Occurrences.htm}{\code{Occurrence}} class.

\begin{figure}
    \begin{center}
        \includegraphics[width=\columnwidth]{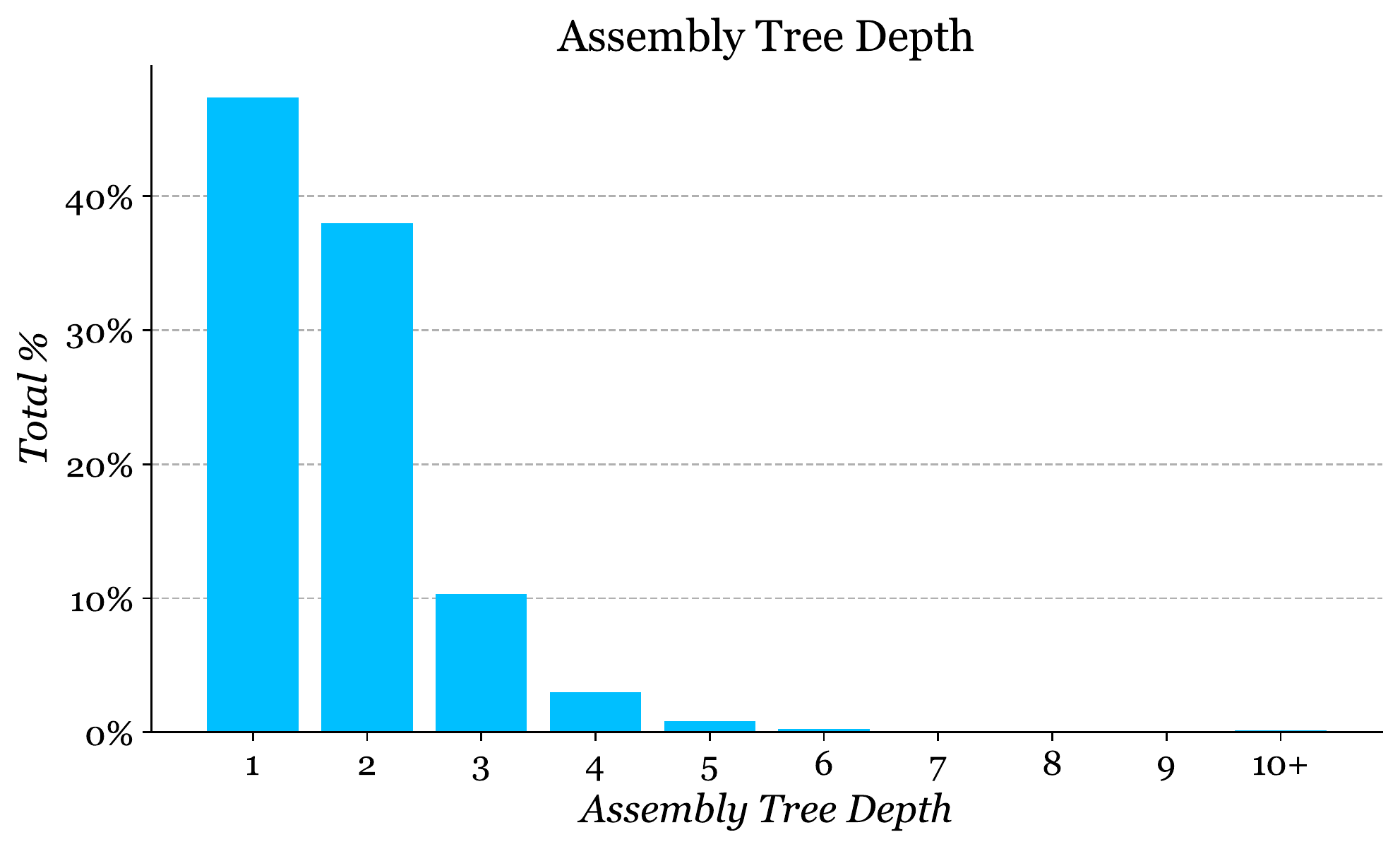}
        \caption{Assembly tree depth shown as a distribution across the assembly data.}
        \label{figure:assembly_tree_depth}
    \end{center}
\end{figure}

\paragraph{Tree} The dataset contains information about the hierarchy of occurrences defined by the designer, as shown in Figure~\ref{figure:example_assembly_graph} (right). The tree contains this hierarchy information by linking to occurrence UUIDs. Figure~\ref{figure:assembly_tree_depth} shows the distribution of assembly tree depth, as defined by how many occurrences of components are nested in hierarchical layers below the root level.

\begin{figure}
    \begin{center}
        \includegraphics[width=\columnwidth]{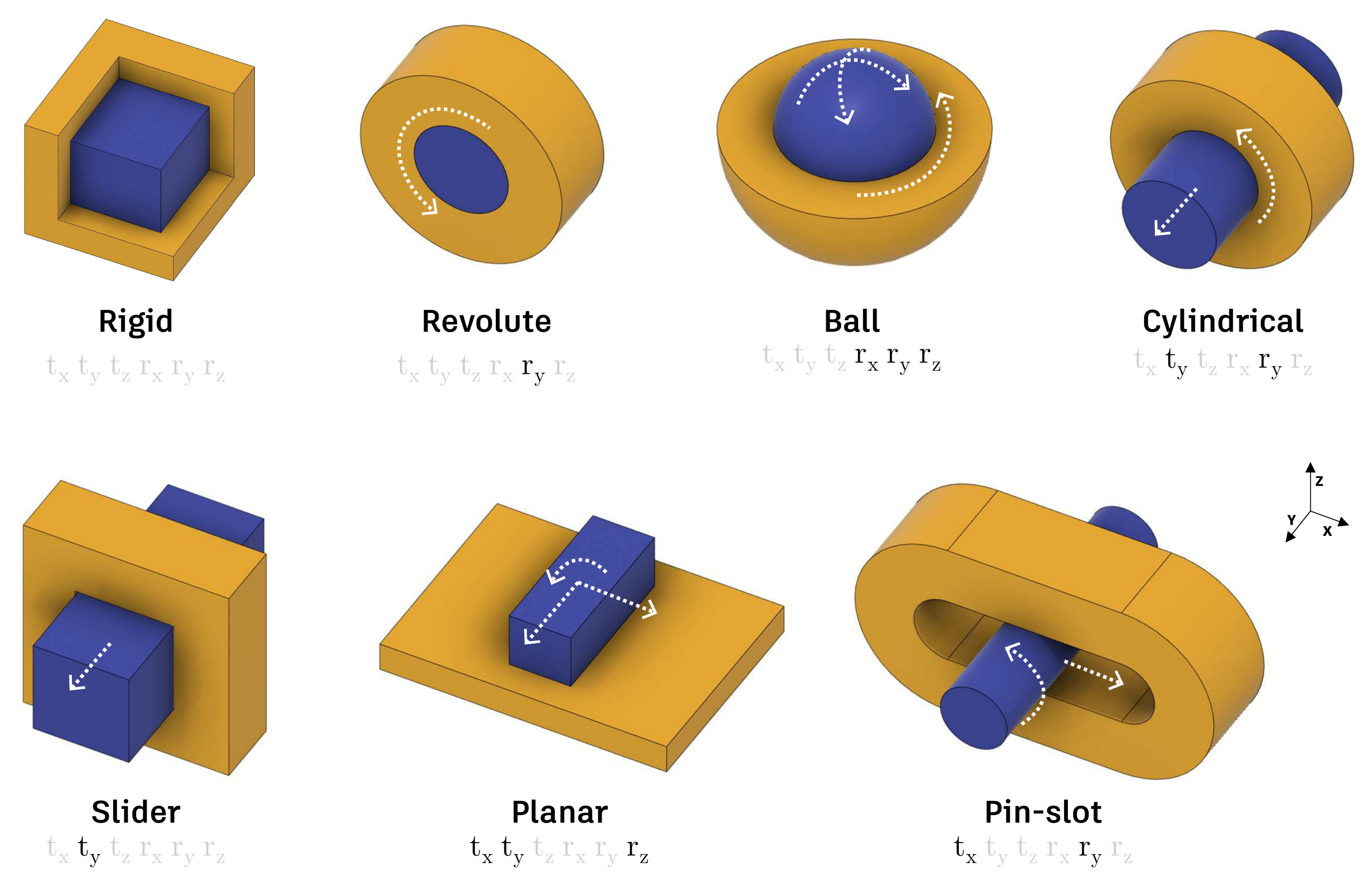}
        \caption{An overview of the joint types in the \FData{} and their degrees of freedom (DOF).}
        \label{figure:joint-types}
    \end{center}
\end{figure}

\begin{figure}
    \begin{center}
        \includegraphics[width=\columnwidth]{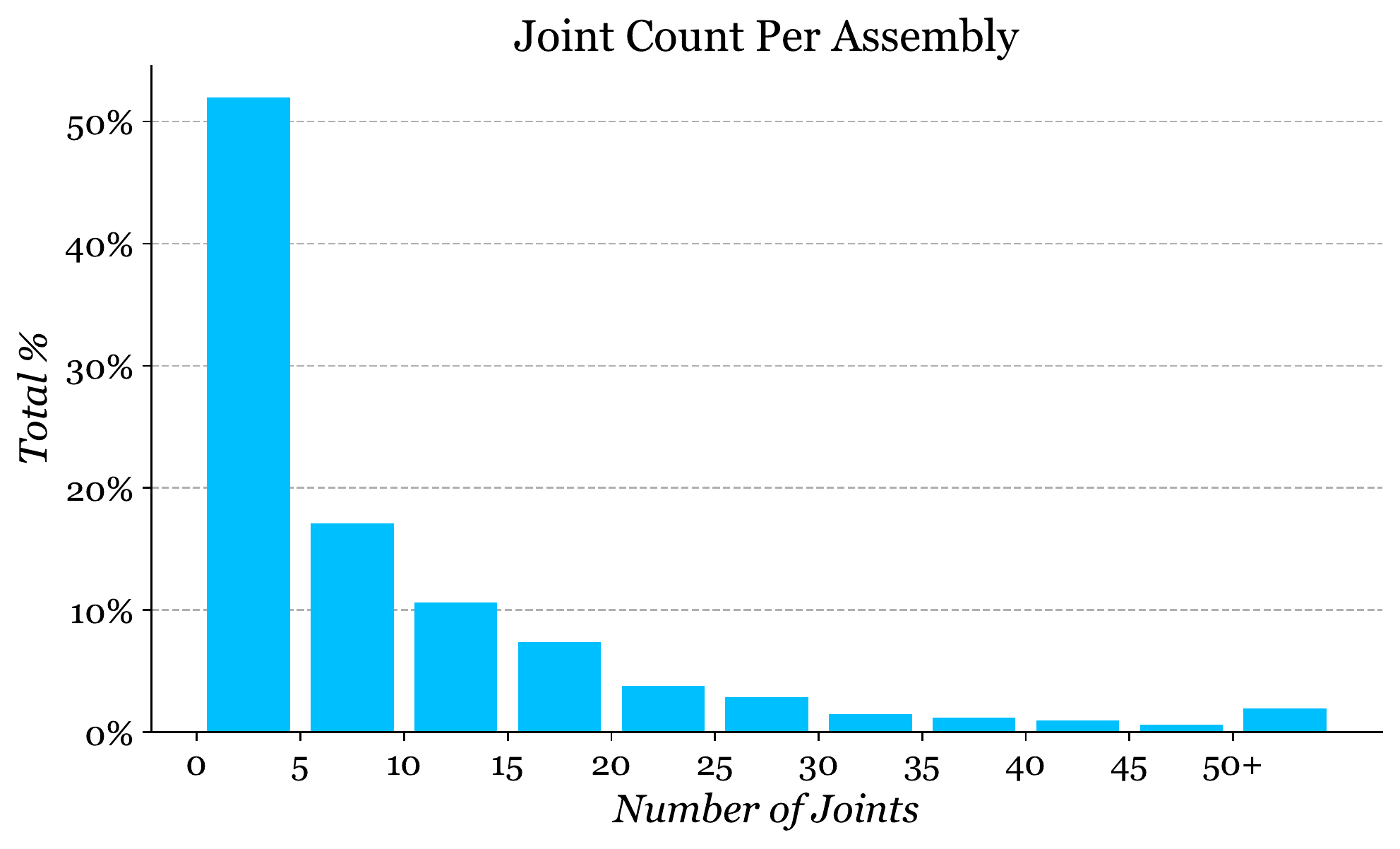}
        \caption{The number of joints in an assembly, shown as a distribution across the assembly data excluding assemblies without joints.}
        \label{figure:assembly_joint_count}
    \end{center}
\end{figure}

\paragraph{Joints} In CAD, joints specify movement between parts by constraining the degrees of freedom (DOF) of one part with respect to another. Specifically, joints are defined between occurrences. Joints are given a UUID, and contain the following information: name, type, parent component, occurrence one (the first occurrence that is part of the joint), occurrence two (the second occurrence being mated to the first), geometry or origin one (containing information about the designer-selected B-Rep entity and joint axis on body one), geometry or origin two (containing information about the designer-selected B-Rep entity and joint axis on body two), timeline index (indicating the order in which the joint was added relative to other joints or occurrences), offset (distance separating geometry one from geometry two), angle (angle between geometry one and geometry two), and the flag \code{is\textunderscore flipped} (indicating the positive or negative direction of the joint). As shown in Figure~\ref{figure:joint-types}, Fusion 360 has seven different types of joints, each with associated joint motion information defining the DOF, motion limits, and rest state. We list below the different types of joints and the associated Fusion 360 API class.
\begin{itemize}
    \footnotesize
    \item \code{RigidJointType}: \href{https://help.autodesk.com/cloudhelp/ENU/Fusion-360-API/files/RigidJointMotion.htm}{\code{RigidJointMotion}}
    \item \code{RevoluteJointType}: \href{https://help.autodesk.com/cloudhelp/ENU/Fusion-360-API/files/RevoluteJointMotion.htm}{\code{RevoluteJointMotion}}
    \item \code{SliderJointType}: \href{https://help.autodesk.com/cloudhelp/ENU/Fusion-360-API/files/SliderJointMotion.htm}{\code{SliderJointMotion}}
    \item \code{CylindricalJointType}: \href{https://help.autodesk.com/cloudhelp/ENU/Fusion-360-API/files/CylindricalJointMotion.htm}{\code{CylindricalJointMotion}}
    \item \code{PinSlotJointType}: \href{https://help.autodesk.com/cloudhelp/ENU/Fusion-360-API/files/PinSlotJointMotion.htm}{\code{PinSlotJointMotion}}
    \item \code{PlanarJointType}: \href{https://help.autodesk.com/cloudhelp/ENU/Fusion-360-API/files/PlanarJointMotion.htm}{\code{PlanarJointMotion}}
    \item \code{BallJointType}: \href{https://help.autodesk.com/cloudhelp/ENU/Fusion-360-API/files/BallJointMotion.htm}{\code{BallJointMotion}}
\end{itemize}
Figure~\ref{figure:assembly_joint_count} shows the number of joints in an assembly as a distribution across the dataset, excluding assemblies without joints. Further information on joints can be found in the Fusion 360 API documentation for the \href{https://help.autodesk.com/cloudhelp/ENU/Fusion-360-API/files/Joint.htm}{\code{Joint}} and \href{https://help.autodesk.com/cloudhelp/ENU/Fusion-360-API/files/AsBuiltJoint.htm}{\code{AsBuiltJoint}} classes.

\begin{figure}
    \begin{center}
        \includegraphics[width=\columnwidth]{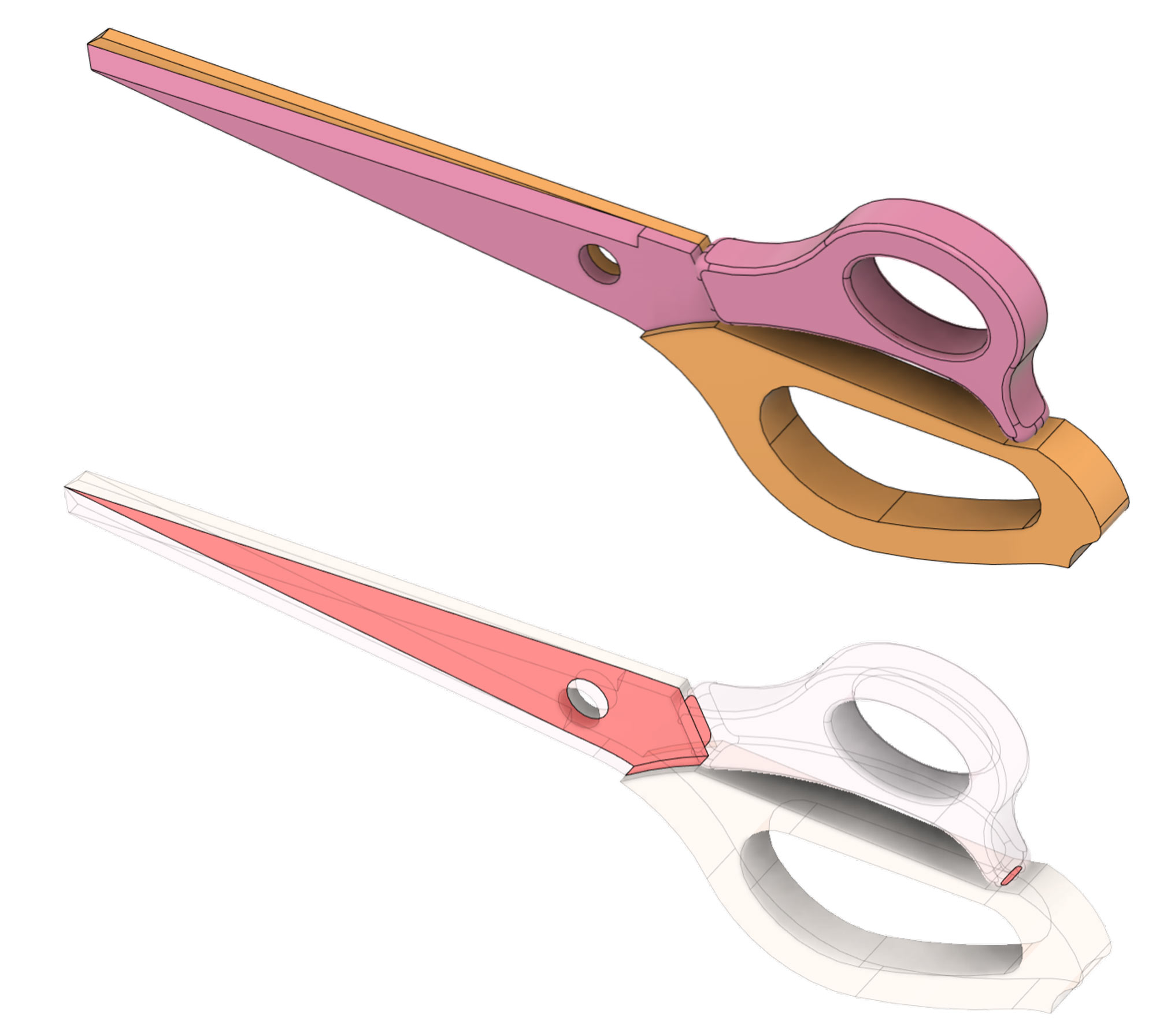}
        \caption{Two bodies (top) and their respective contacts highlighted in red (bottom).}
        \label{figure:contact-example}
    \end{center}
\end{figure}

\begin{figure}
    \begin{center}
        \includegraphics[width=\columnwidth]{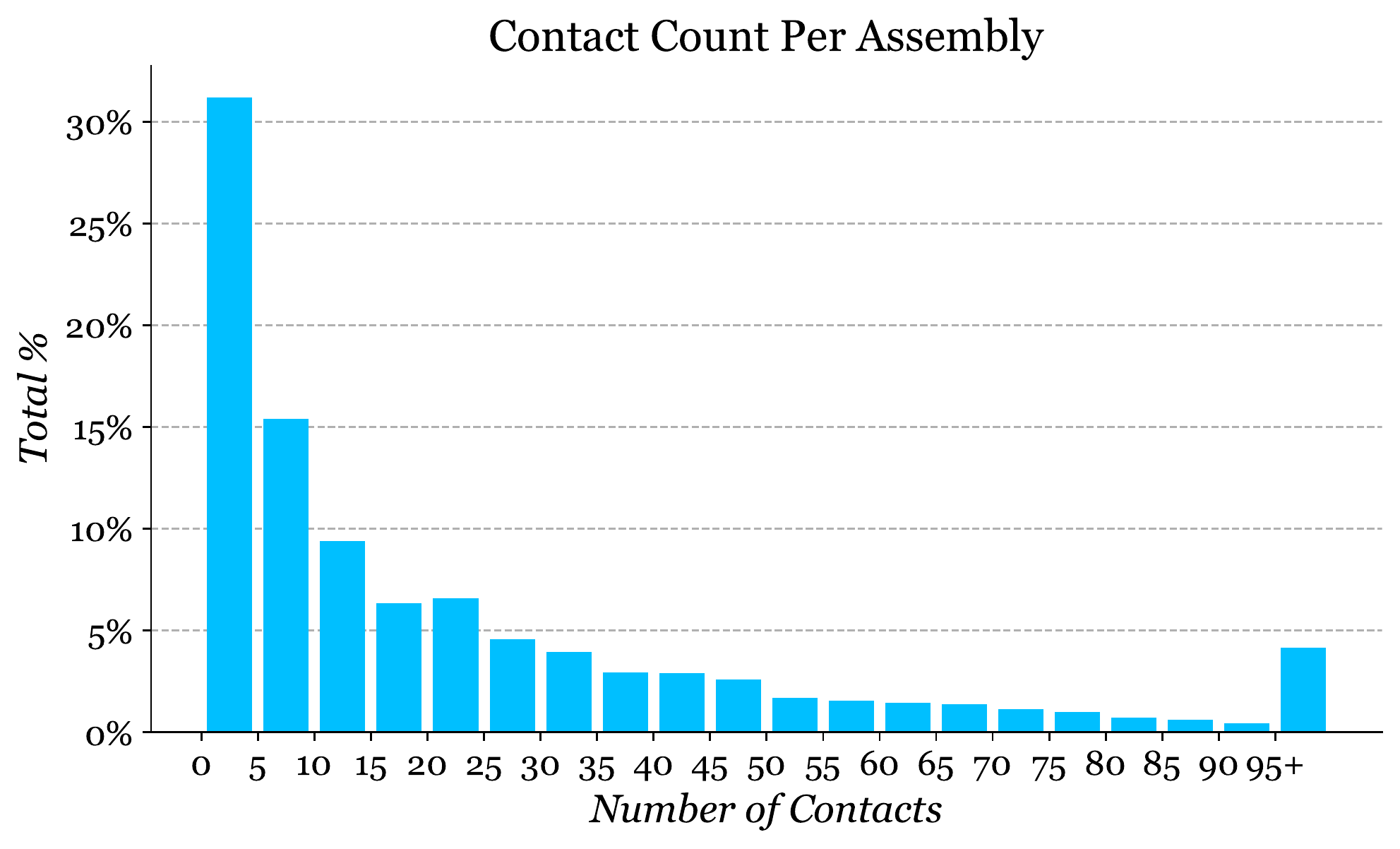}
        \caption{The number of contacts (B-Rep faces that are in contact with another body) in an assembly, shown as a distribution across the assembly data excluding assemblies without any contacts.}
        \label{figure:assembly_contact_count}
    \end{center}
\end{figure}

\paragraph{Contacts} Contacts are present when two bodies share coincident faces or are within a tolerance of 0.1 mm. An example of a contact is shown in Figure~\ref{figure:contact-example}. Each contact present in the assembly is defined in the JSON with a pair of entities, indicating which faces are in contact. Each entity includes information about the body it belongs to, the occurrence it belongs to, the type of surface in contact (cylindrical, planar, etc.), the bounding box surrounding the entity, and an index that can be used to uniquely identify the face. Figure~\ref{figure:assembly_contact_count} shows the number of contacts in an assembly as a distribution across the dataset.

\paragraph{Properties} We provide assembly-level metadata about the design under the properties element in the JSON. Table~\ref{table:assembly-properties} provides a summary of this metadata, including geometric information about the whole assembly, physical properties of the geometry, online statistics derived from the public web page hosted on the Autodesk Online Gallery \cite{autodeskOnlineGallery} at the time the data was downloaded, and some user-selected categorical tags that provide more information about the context of the assembly. 

\begin{table}
    \centering
    \small
    \begin{tabular}{r|p{5.6cm}}
        \toprule
        \textbf{Category} & \textbf{Property}\\
        \midrule
        \textbf{Geometric} & Vertex count\\
         & Edge count\\
         & Face count\\
         & Loop count\\
         & Shell count\\
         & Body count\\
         & Surface type count (plane, torus, cylinder, NURBS, cone, sphere, etc.)\\
         & Vertex valence instance count\\
         \midrule
        \textbf{Physical} & Bounding box\\
         & Area\\
         & Volume\\
         & Density\\
         & Mass\\
         & Center of mass\\
         & Principal axes\\
         & Moments of inertia\\
         \midrule
        \textbf{Online} & Likes count\\
         & Comments count\\
         & Views count \\
         \midrule
        \textbf{Tags} & Products used tag\\
         & Category tag (automotive, art, electronics, engineering, game, machine design, interior design, medical, product design, robotics, sport, tools, toys, etc.)\\
         & Industry tag (architecture, engineering \& construction; civil infrastructure; media \& entertainment; product design \& manufacturing; other industries). \\
        \bottomrule
    \end{tabular}
    \caption{Each assembly includes the metadata listed in this table.}
    \label{table:assembly-properties}
\end{table}

\begin{figure}
    \begin{center}
        \includegraphics[width=\columnwidth]{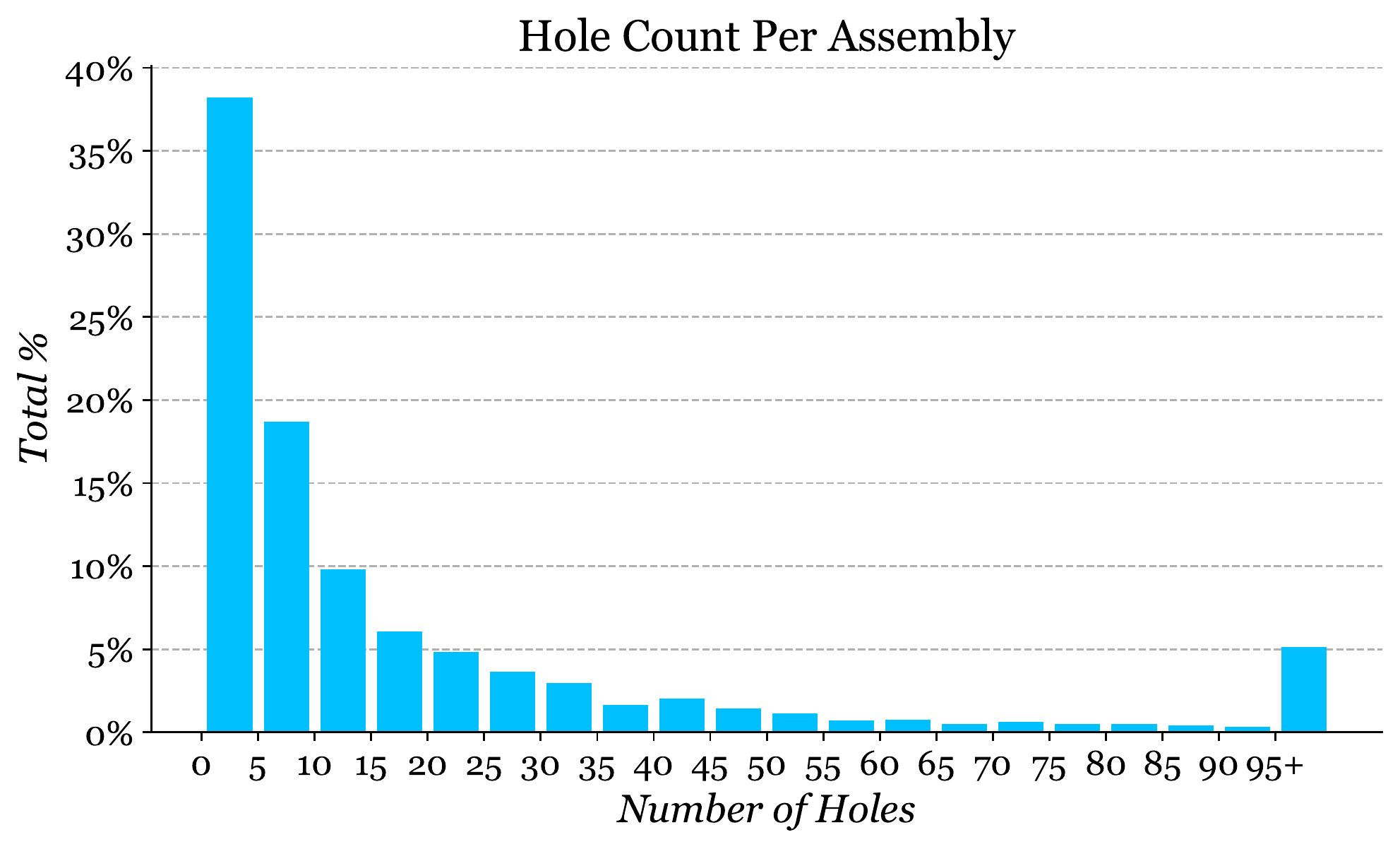}
        \caption{The number of holes in an assembly, shown as a distribution across the assembly data excluding assemblies without any holes.}
        \label{figure:assembly_hole_count}
    \end{center}
\end{figure}

\paragraph{Holes} In CAD models, holes are common design features that often serve a specific purpose. Parts are commonly held together with bolts and screws, which either pass through or end in holes in the parts. As holes are an important design feature, we use an industrial CAD feature recognition tool to identify holes in each assembly for inclusion in the JSON data. Each hole lists information about the body it is in, diameter, length, direction, and faces and edges that belong to the hole. Holes are also labeled with a hole type denoting the shape at the hole entrance, e.g. counterbore, countersunk, and at the end of the hole, e.g. through-hole, blind hole. Figure~\ref{figure:assembly_hole_count} shows the number of holes in an assembly as a distribution across the dataset.

\paragraph{Assembly Data Geometry Format} 
\label{section:assembly_data_geometry_format}
We provide geometry in several data formats, described below.

\textit{Boundary Representation.} B-Rep data consists of faces, edges, loops, coedges and vertices \cite{weiler1986}. A face is a connected region of the model’s surface. An edge defines the curve where two faces meet and a vertex defines the point where edges meet. Faces have an underlying parametric surface which is divided into visible and hidden regions by a series of boundary loops. A set of connected faces forms a body. 
B-Rep data is provided as .smt files representing the ground truth geometry and .step as an alternate neutral B-Rep file format. The .smt file format is the native format used by Autodesk Shape Manager, the CAD kernel within Fusion 360, and has the advantage of minimizing conversion errors.

\textit{Mesh.} Mesh data is provided in .obj format representing a triangulated version of the B-Rep. Triangles belonging to each B-Rep faces are denoted in the .obj file as groups, for example, \code{g face 1}, indicates the next series of triangles in the file belong to the B-Rep face with index 1. B-Rep edges are converted to poly lines and added to the .obj file. The B-Rep edge and half-edge index is also denoted, for example, \code{g halfedge 7 edge 3}.  Using these group indices it is possible to map directly from B-Rep faces and edges to mesh triangles and poly lines. Note that meshes provided in the dataset are not guaranteed to be manifold. 

Other representations, such as point clouds or voxels, can be generated from the mesh or B-Rep data using existing data conversion routines and are not included in the dataset. For convenience we include a thumbnail .png image file together with each body and one for the overall assembly.
Geometry files are named according to the UUID of the body in the assembly, with the overall assembly files given the name `assembly' with the appropriate file extension.

\paragraph{Assembly Data Split} The assembly data is divided into a train-test split found in the file `train\textunderscore test.json', and a train-test-cross split found in the file `train\textunderscore test\textunderscore cross.json'. 

The main train-test split is created by randomly sampling 80\% of the data into train, and 20\% into test. 

The train-test-cross split, on the other hand, is a cross dataset split with the joint data. For this split, data is chosen such that there is no overlap between the train assembly data and the test joint data, and no overlap between the test assembly data and the train joint data. This train-test-cross split allows for a model to be trained using train joint data and tested with test assembly data. It's important to note that the train-test-cross split does not represent a normal distribution of the assembly data, and be may be biased. Nonetheless, the train-test-cross split is included to support applications leveraging both the assembly and joint data.

\begin{figure}
    \begin{center}
        \includegraphics[width=\columnwidth]{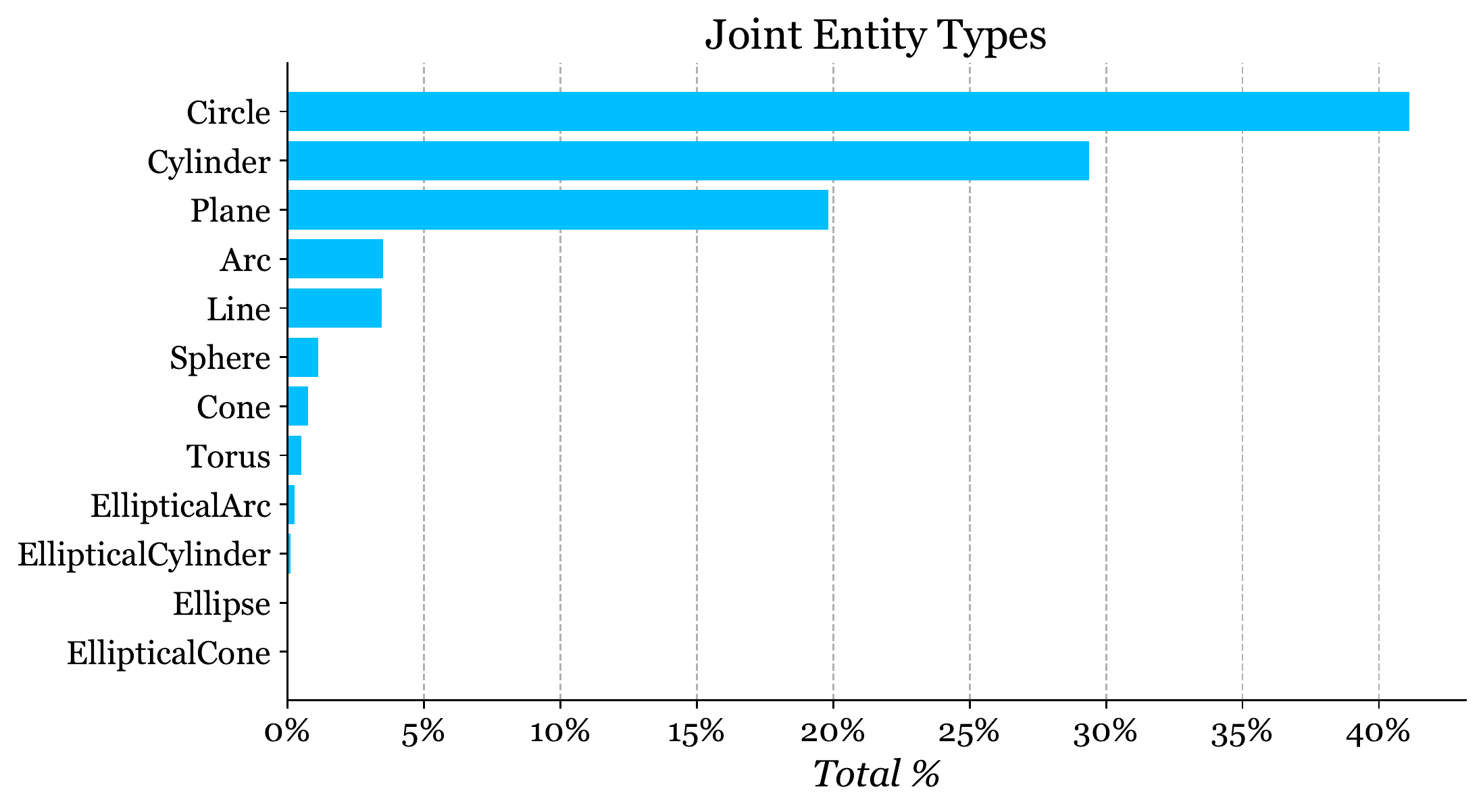}
        \caption{The distribution in the joint data of B-Rep entity types, from both surfaces and curves, selected by designers when creating a joint.}
        \label{figure:dataset_joint_entity_types}
    \end{center}
\end{figure}

\paragraph{Assembly Graph Representations} Using the contact, joint, or designer-defined assembly tree, various graph representations can be formed. For example, contact surface information and joint data can be used to construct a parts graph~\cite{fu2015computational} where graph vertices represent parts and graph edges denote user-defined relative motion and constrained DOFs between part pairs (Figure~\ref{figure:example_assembly_graph}, center). 
Similarly, the part hierarchy information found in the design tree can be used to form a hierarchical assembly graph~\cite{slyadnev2020role} where the vertices of the graph are components and the graph edges denote parent-child relationships (Figure~\ref{figure:example_assembly_graph}, right).

\paragraph{Assembly Data Use Cases} We envision numerous use cases our dataset could enable. For example, project-level metadata such as the user-defined category and industry tags, could serve as a proxy for design requirements and support research around design synthesis from design requirements \cite{yoo2021integrating}. Similarly, project level meta-data about the popularity of the assemblies in the Autodesk Online Gallery (view counts, like counts, and comment counts) could support research around customer requirement analysis \cite{maguire2002user}. Per-body material metadata could support material prediction tasks \cite{arnold2012materials}. This data could also support model-reuse workflows as well as global or part-level similarity retrieval tasks \cite{lupinetti2018multi}.

\paragraph{Assembly Data Limitations}
Due to the complexity of some large assemblies, we encounter data processing failures that force us to exclude some assemblies and elements of assemblies. In cases where the exported assembly cannot be rebuilt to match the original assembly, we are forced to discard the assembly but use the joints individually as described in Section~\ref{section:joint_dataset}. When processing of contact or joint information fails, but the exported assembly matches the original, we retain the assembly and mark the contact or joint information as \code{null} to indicate a processing error. We find that processing failures occur for contacts with 13\% of assemblies and for joints with 2\%. A general limitation of the assembly data is that only 20\% of designs have joints defined. We attribute this to the extra work required to manually define joints, which \textit{JoinABLe} seeks to address by offering improved automation of joint setup.

\begin{figure}
    \begin{center}
        \includegraphics[width=\columnwidth]{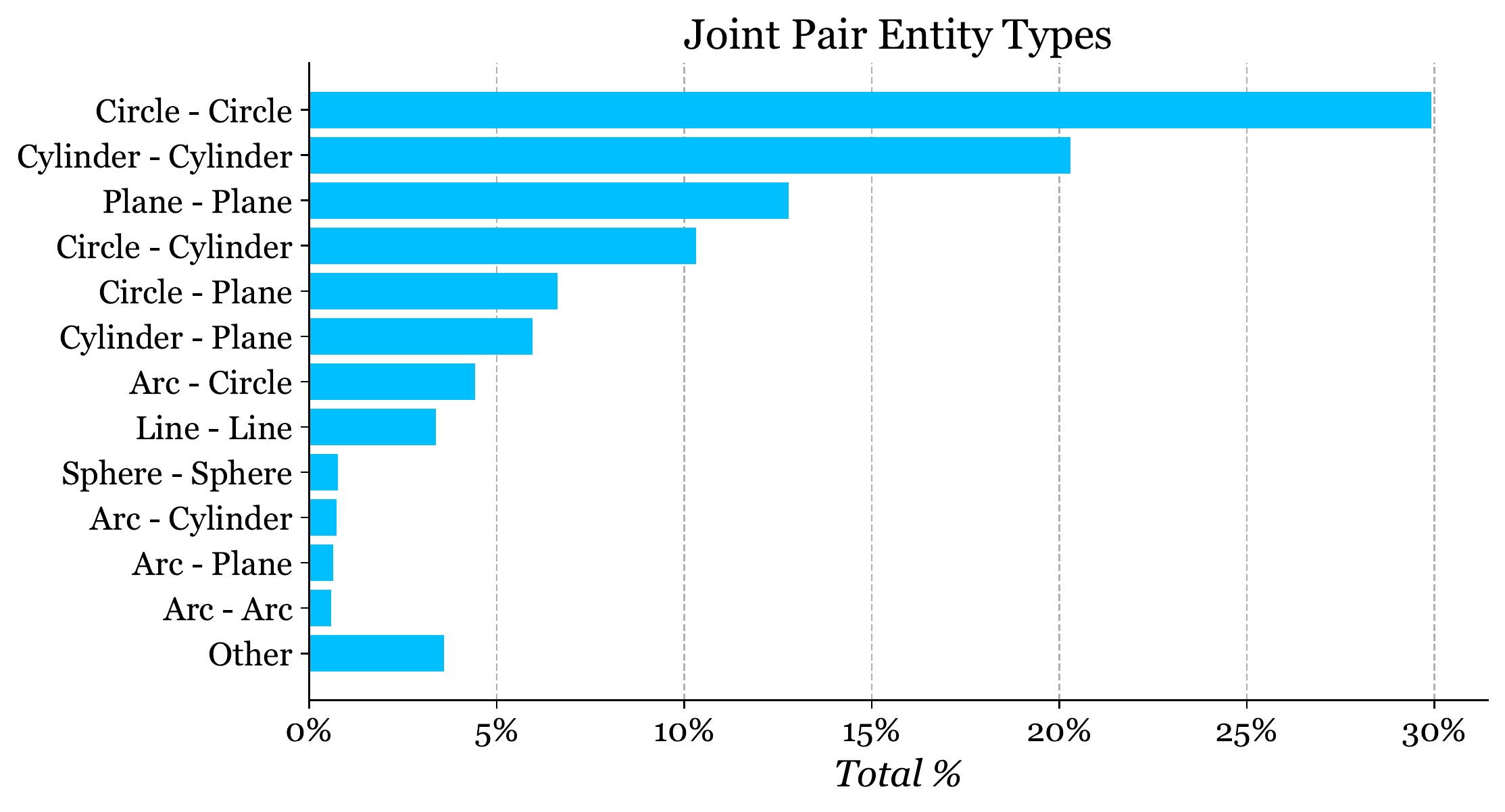}
        \caption{The distribution in the joint data of B-Rep entity types, shown as a pair, selected by designers when creating a joint between two bodies.}
        \label{figure:dataset_joint_pair_entity_types}
    \end{center}
\end{figure}


\subsubsection{Joint Data}
\label{section:joint_dataset}
Our joint data consists of pairs of parts with multiple joints defined between them, as described in Section~\ref{section:dataset} and illustrated in Figure~\ref{figure:example_designs}. We provide the joint data separate from the assembly data as an accessible standalone dataset for the joint prediction task and to increase data quantity by including valid joints that were excluded from the assembly data due to unrelated data processing issues. Our joint data consists of 19,156 joint sets, containing 32,148 joints between 23,029 different parts. We provide an approximate 70/10/10/10\% data split, for the train, validation, test, and original distribution test sets respectively.

\paragraph{Joint Data Labels}

The designer-selected B-Rep faces and edges form the ground truth labels and are stored as indices that map to B-Rep entities, or alternatively triangle and poly line groups in the mesh representation we provide. Figure~\ref{figure:dataset_joint_entity_types} shows the overall distribution of entity types (from both surfaces and curves) that are selected by designers to create joints. Circle and cylinder types are most prevalent due to their use with fasteners. Figure~\ref{figure:dataset_joint_pair_entity_types} further shows the relationship between pairs of joint entities by their entity type.

\begin{figure}
    \begin{center}
        \includegraphics[width=\columnwidth]{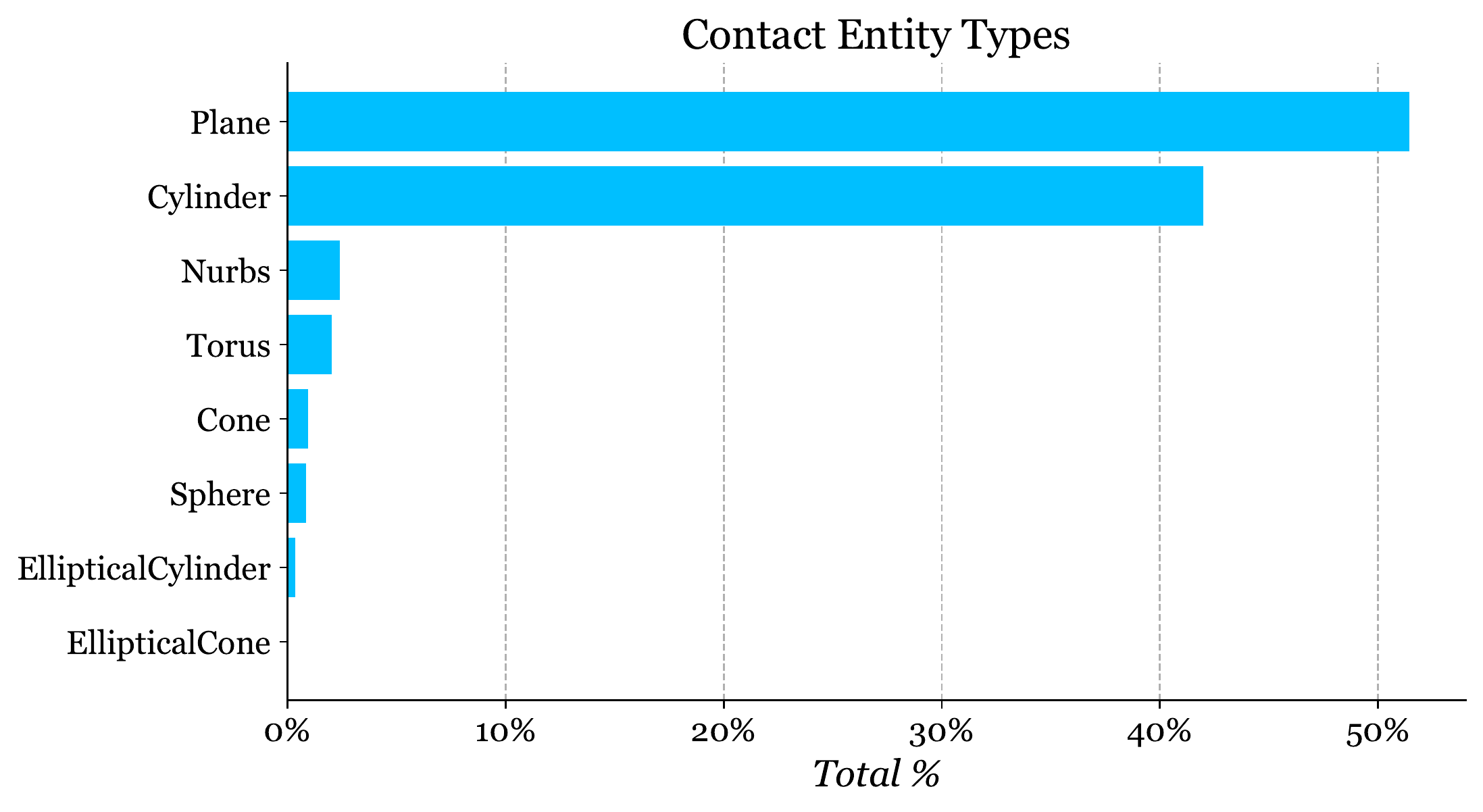}
        \caption{The distribution of B-Rep surface types for all contacts in the joint data.}
        \label{figure:dataset_joint_contact_entity_types}
    \end{center}
\end{figure}

\begin{figure}
    \begin{center}
        \includegraphics[width=\columnwidth]{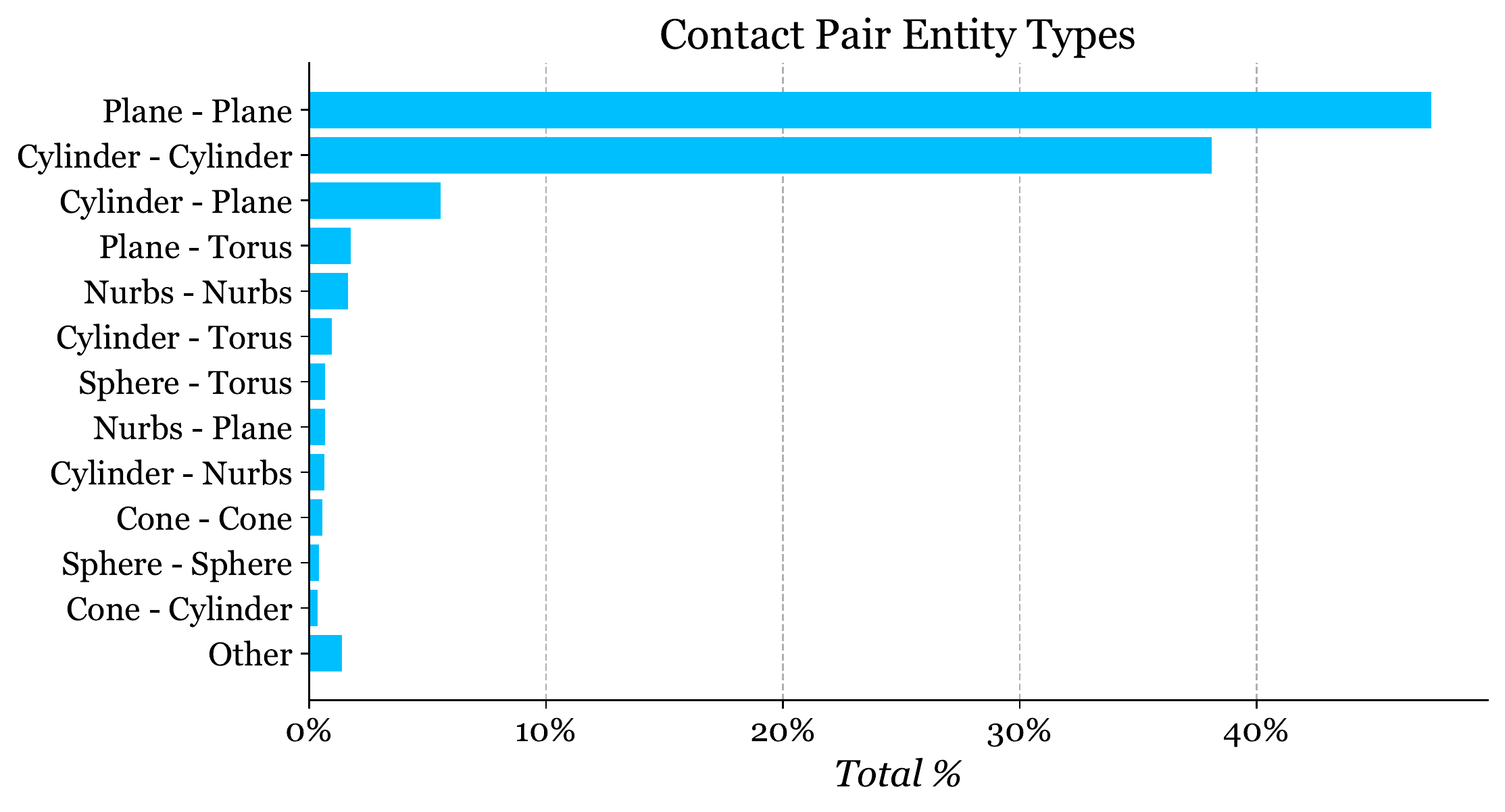}
        \caption{The distribution of B-Rep surface type pairs, in contact with one another in the joint data.}
        \label{figure:dataset_joint_contact_pair_entity_types}
    \end{center}
\end{figure}

We also provide contact labels that indicate which B-Rep faces are coincident or within a tolerance of 0.1mm, when a joint is in an assembled state. Figure~\ref{figure:dataset_joint_contact_entity_types} shows the overall distribution of surface entity types that are in found to be in contact. Figure~\ref{figure:dataset_joint_contact_pair_entity_types} further shows the relationship between pairs of surfaces that are found to be in contact. Finally we provide hole labels as found in the assembly data.

\begin{figure}
    \begin{center}
        \includegraphics[width=\columnwidth]{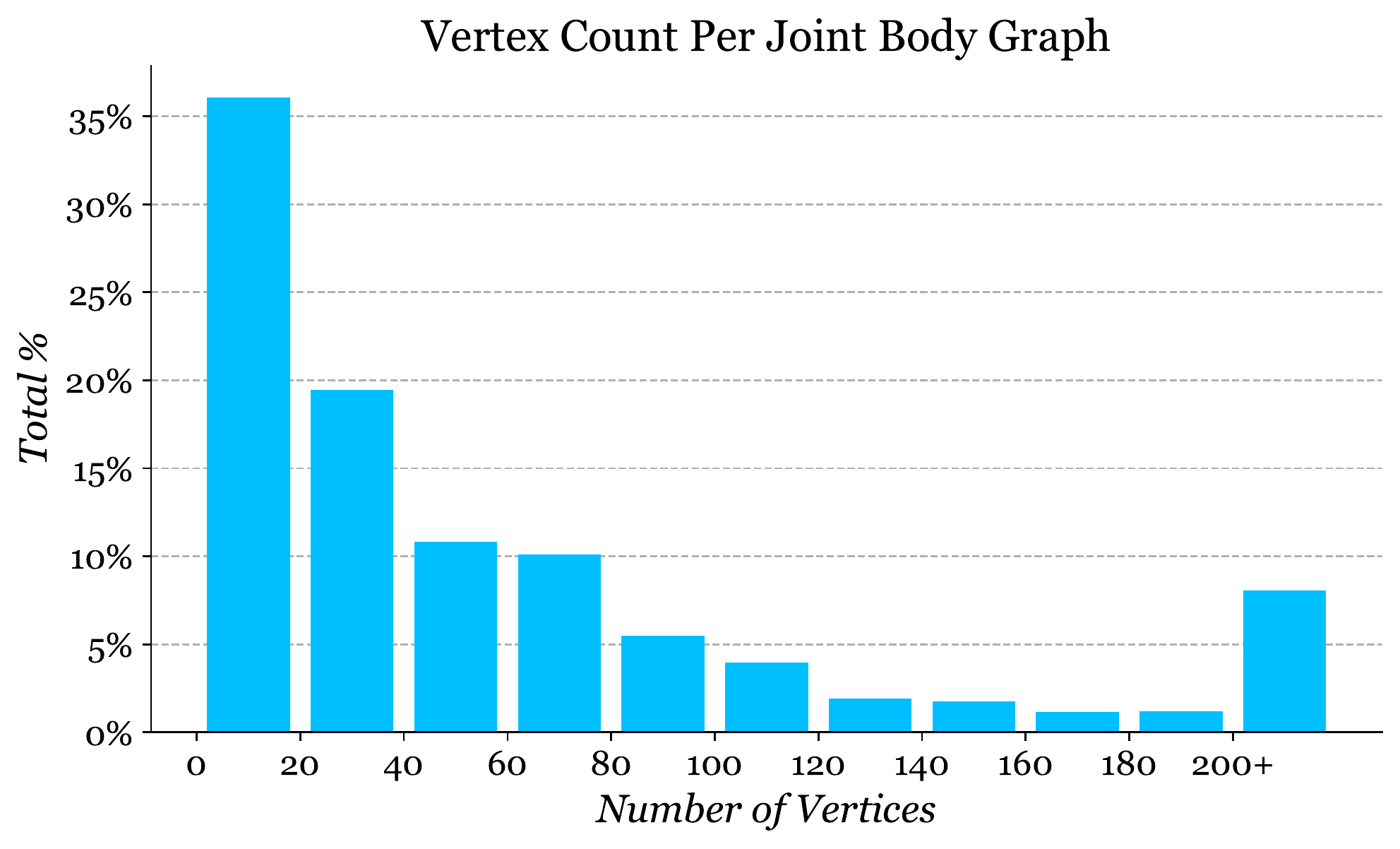}
        \caption{The number of vertices in each graph created from the faces and edges in a B-Rep body forming a joint. Shown as a distribution to indicate the complexity of designs in the joint data.}
        \label{figure:joint_graph_vertex_count}
    \end{center}
\end{figure}

\paragraph{Joint Data Geometry Format}
We provide geometry in the same B-Rep and mesh data formats as the assembly data. In addition we provide a graph representation of the B-Rep topology and features used in our experiments. Here each graph vertex represents a B-Rep face or edge, with the graph edges defined by adjacency. We include the input features described in Section~\ref{section:feature-ablation} as well as the UV-grid features (points, normals, trimming mask, tangents) used by the B-Grid baseline method. We extract the input features using the Fusion 360 API. We store the graph data as a JSON file in the NetworkX\footnote{\href{https://networkx.org}{https://networkx.org}} node-link data format for easy integration with common graph neural network frameworks. Figure~\ref{figure:joint_graph_vertex_count} shows the distribution of graph vertices in our graph representation of each part. This provides an approximate indication of the complexity of designs in the joint data.

\begin{figure}
    \begin{center}
        \includegraphics[width=\columnwidth]{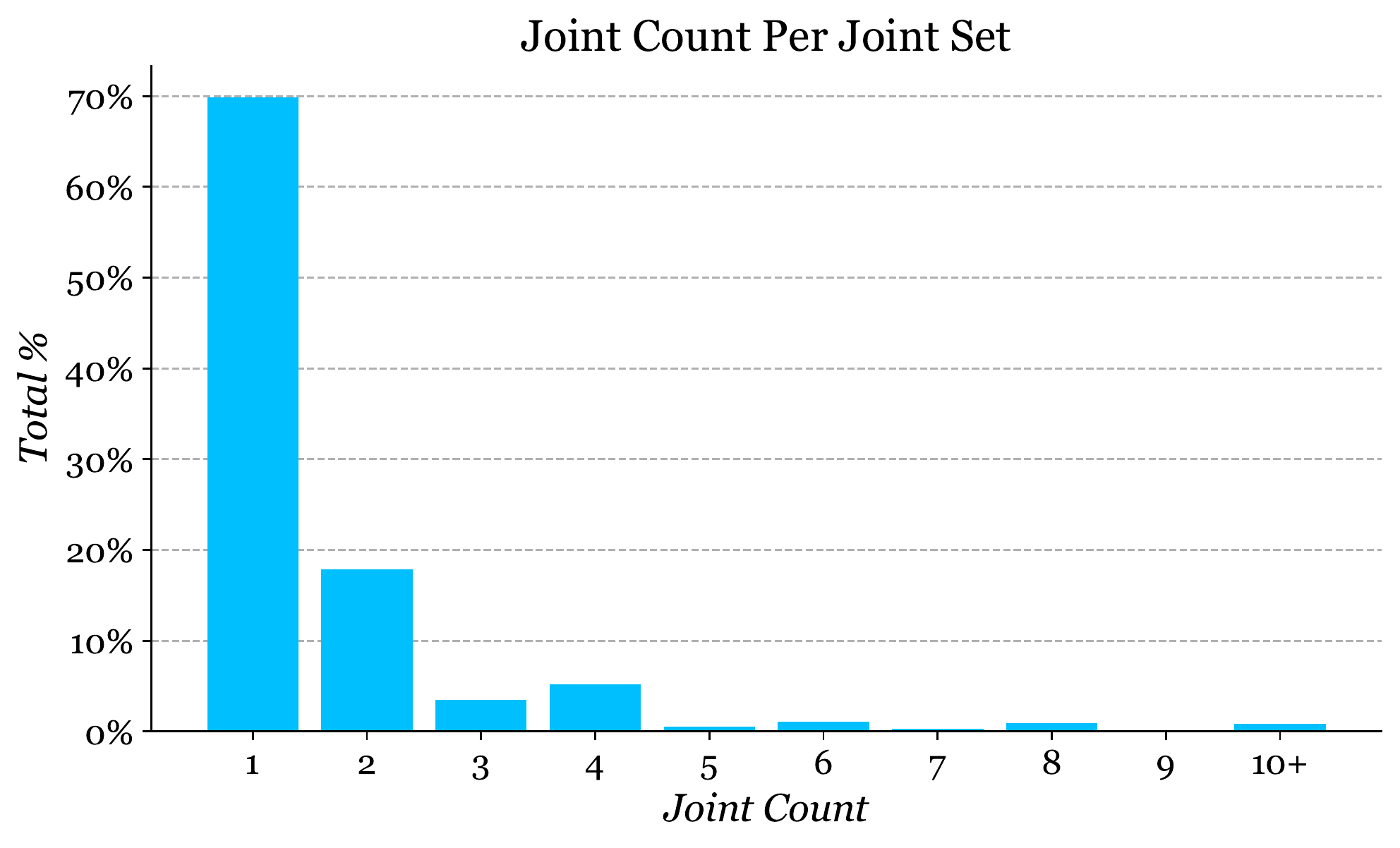}
        \caption{The distribution of joints in each joint set across the joint data.}
        \label{figure:joint_count_per_set}
    \end{center}
\end{figure}

\paragraph{Joint Consolidation}
We now provide additional details about the process of joint consolidation described in Section~\ref{section:weak-supervision}.
We perform joint consolidation across all joints in the dataset. We begin by creating a unique hash for each pair of parts based on the B-Rep topology and geometric properties. For each B-Rep face and edge in a part we add the volume, moments of inertia, surface type, curve type, area, and length to the hash. Floating point values are truncated to 3 decimal place for moments of inertia and 1 decimal place for all other values. For surface and curve type we use the string values directly. This approach ensures matches between parts have identical topology, a key requirement to ensure the ground truth entities are mapped correctly. We then concatenate the hash for each pair of parts and group all joints with the same combined hash together to form joint sets. Figure~\ref{figure:joint_count_per_set} shows the number of joints in each joint set after joint consolidation has been performed. Roughly 70\% of joint sets have a single joint, while the remaining 30\% have more than one joint.

\paragraph{Joint Data Limitations}
To better scope the joint data and joint prediction task, we exclude some advanced joint configurations. We include only joints where a user has selected a single B-Rep entity on each part. In Fusion 360 it is possible to select multiple entities, for example the user may select a plane face, then a corner of the plane to further refine the position of the joint axis. Although we exclude these samples from the joint data to narrow the scope of the joint prediction task, we include them in the assembly data where they represent 9\% of all joints.

\subsubsection{Supporting Code}
Together with the dataset we provide supporting code for working with the data in Python. To work with B-Rep data, we provide several Fusion 360 add-ins that can rebuild the assembly and joint data as a parametric CAD model from the JSON data provided in the dataset. The rebuilt CAD models have a component hierarchy and parametric joints fully defined. Fusion 360 is available free for students and educators. To work with the mesh data we provide example code to assembly and visualize the .obj files using common open source Python libraries. Finally we provide an assembly graph class to construct NetworkX graphs, such as those show in Figure~\ref{figure:example_assembly_graph}, from the assembly data.

\subsection{Experiments}
\label{section:appendix_experiments}
In this section we provide additional details of the experiments in Section~\ref{section:experiments} and report additional experiment results to examine which input features the network uses to make predictions, the effect of introducing label augmentation, and performance on a test set with potential positive unlabelled samples.

\begin{figure}
    \begin{center}
        \includegraphics[width=\columnwidth]{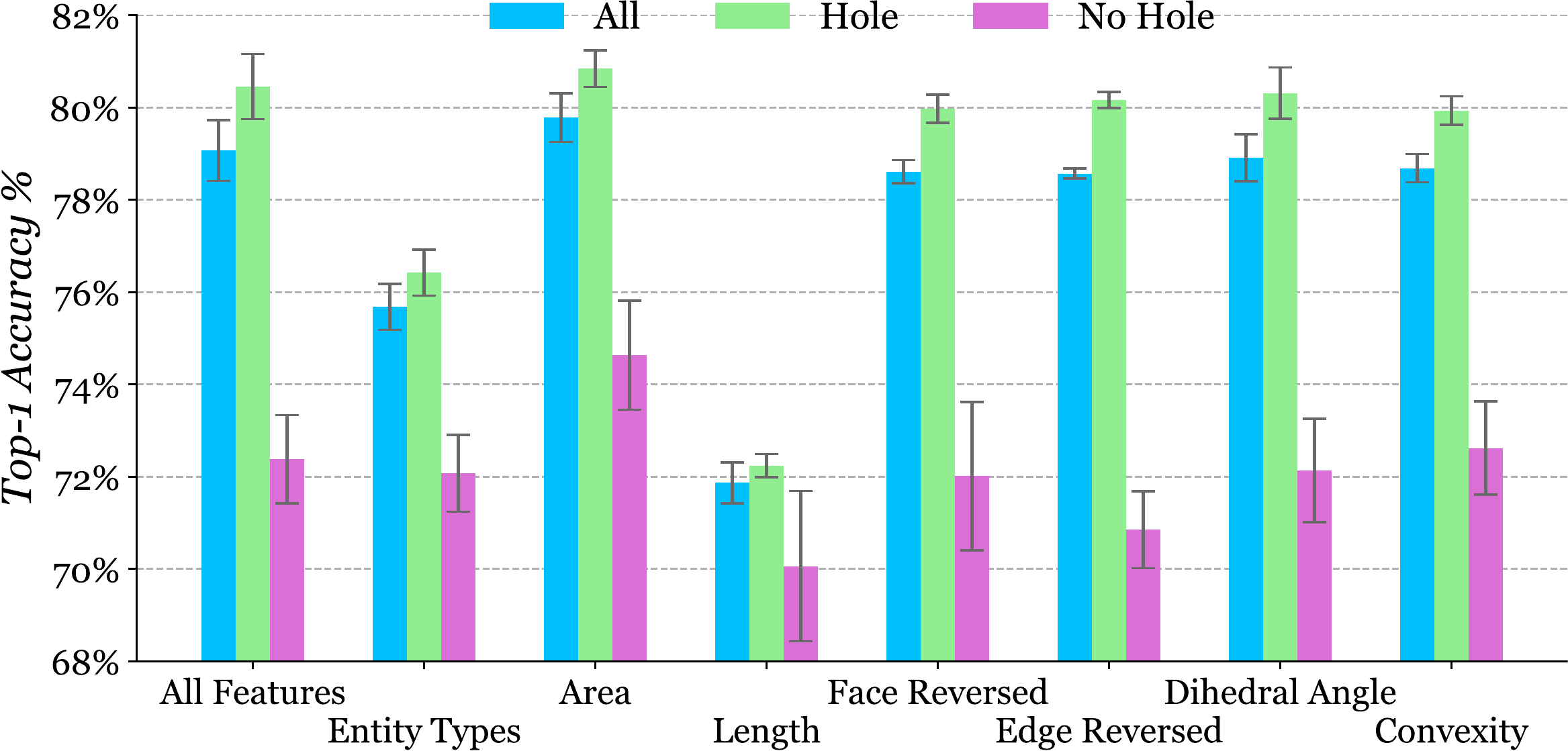}
        \caption{Effect on joint axis prediction top-1 accuracy by removing specific input features from the network. Results are shown for all data samples in the validation set (All), as well as the subset with holes (Hole) and without holes (No Hole).}
        \label{figure:feature-ablation}
    \end{center}
\end{figure}

\subsubsection{Training}
All experiments are run for 100 epochs with the final training weights used at test time. The reported values are the average over 5 runs with different random seeds and error bars indicate the standard deviation. All networks are implemented in PyTorch.

Our method uses the PyTorch Geometric\footnote{\href{https://github.com/pyg-team/pytorch_geometric}{https://github.com/pyg-team/pytorch\_geometric}} implementation of GAT v2~\cite{brody2021attentive} to perform message passing. Due to memory limitations, during training of our method we skip training samples where the graph representations of both parts have more than 950 graph vertices combined. All experiments with our method are trained with a per-GPU batch size of 2 across 4 NVIDIA V100 GPUs, dropout disabled, batch norm disabled, learning rate of 0.0001 with the Adam optimizer, and a learning rate scheduler that reduces on plateau.

\begin{figure*}
    \begin{center}
        \includegraphics[width=\linewidth]{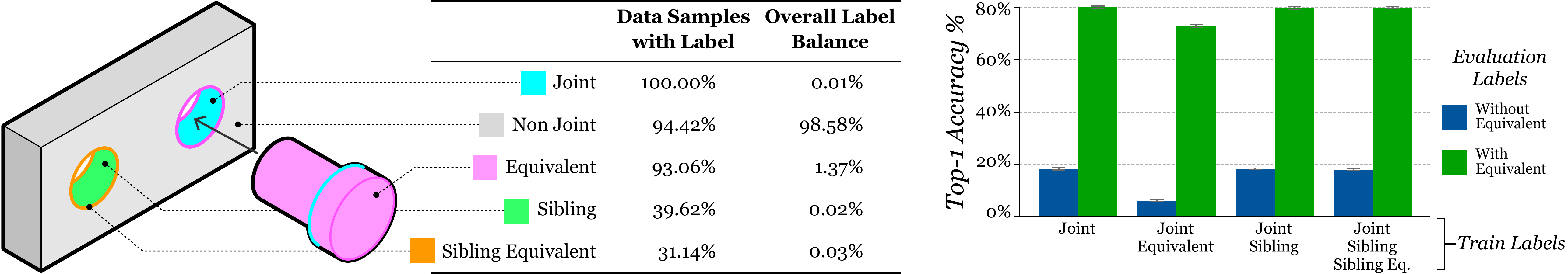}
        \caption{
        Label augmentation ablation study.
        \textit{Left}: Types of label augmentation used, the percentage of training data samples with each label type, and the overall label balance.
        \textit{Right}: Effect on joint axis prediction top-1 accuracy by training with different labels and evaluating with/without \textit{Equivalent} labels.
        }
        \label{figure:label_augmentation}
    \end{center}
\end{figure*}

\subsubsection{Evaluation}
We evaluate on all data samples in the test and uniform distribution test set (described in Section~\ref{section:uniform-test-eval}). To ensure large graphs can be evaluated without GPU memory limitations, we perform evaluation on the CPU. 

As described in Section~\ref{section:experiments} we report results for data samples both with and without holes. We consider a data sample to have holes when a hole exists in either of the two parts from a joint set. We take this approach to ensure that for the more challenging `no hole' case, the network is forced to make a prediction for parts without any form of hole, either connected with a joint or otherwise. In our test set we find that 83\% of data samples have holes and 17\% do not.

For quantitative results we evaluate the five trained models for each condition and report the mean result. We follow this procedure for the joint pose prediction task and perform search with the five different trained models. For qualitative results we pick the model with the highest quantitative result to use when generating figures. When multiple ground truth results exist, we show the closest ground truth result to those predicted.

\subsubsection{Input Feature Ablation Study}
\label{section:feature-ablation}
As described in Section~\ref{section:representation}, we use information about individual B-Rep faces and edges readily available in the B-Rep data structure for input features. In this experiment we perform an ablation study to identify which input features have the greatest impact on joint axis prediction performance. For B-Rep faces we evaluate a one-hot vector for the surface type (plane, cylinder, etc.), the area of the face, and a flag indicating if the surface is reversed with respect to the face. For B-Rep edges, we evaluate a one-hot vector for the curve type (line, circle, etc.), the length of the edge, a flag indicating if the curve is reversed with respect to the edge, the dihedral angle at the edge, and a one-hot vector for the edge convexity (convex, concave, etc.).

We train each model by sequentially removing input features and report the top-$1$ accuracy results on the validation set in Figure~\ref{figure:feature-ablation} for \textit{All} samples and the \textit{Hole} and \textit{No Hole} subsets. We also include results using \textit{All Features} for comparison purposes. We firstly observe that performance on the \textit{Hole} samples is consistently higher than on the \textit{No Hole} samples. We find that the \textit{length} feature is the most critical for performance followed by \textit{entity type}. This result aligns with the common use of cylinders at the interface between parts, with the network able to access both the B-Rep surface type and the length of neighboring B-Rep edges in the graph. As the length of the circular edge around the end of a bolt or lip of a hole is proportional to the hole radius, edge length can be effectively used as a way to identify bolts and holes of similar sizes. Conversely we find that the \textit{area} feature does not act as a proxy in the same way and negatively affects performance. In our experiments in Section~\ref{section:experiments} of the main paper we remove the lowest performing features: \textit{area}, \textit{dihedral angle}, and \textit{convexity}.

\subsubsection{Label Augmentation Study}
\label{section:label-ablation}

Our training data has both an extreme label imbalance ($99.9\%$ negative) and positive unlabelled samples on the order of $3\times$ the number of labelled samples. To counter these factors we explore the role of label augmentation with a study that adds three different types of augmented labels to the ground truth labels. Figure~\ref{figure:label_augmentation}, left, illustrates each type of label augmentation and provides statistics on how common these labels are in our training data. \textit{Equivalent} labels, as described in Section~\ref{section:weak-supervision}, share the same joint axis as the ground truth \textit{Joint} labels. \textit{Sibling} labels are used on entities that are geometrically similar to the designer-selected \textit{Joint} entities. These cover cases, such as the one illustrated on the left of Figure~\ref{figure:label_augmentation}, where unlabeled holes exist in a part that have the same diameter as labeled ones. \textit{Sibling Equivalent} labels are the equivalent entities that share the same joint axis. Finally, \textit{Non-Joint} labels are the negative labels. 

Figure~\ref{figure:label_augmentation}, right, shows the effect of adding different training label augmentations with the joint axis prediction task. We show results for evaluation with and without \textit{Equivalent} labels on the validation set. A large increase in accuracy is observed when \textit{Equivalent} labels are used for evaluation, due to the overall increase in positive labels. As we consider a joint axis prediction to be correct if it is co-linear with the ground truth joint axis in either direction, adding \textit{Equivalent} labels during evaluation allows us to better judge network performance. We find that none of the label augmentation strategies increase performance, and in fact the addition of \textit{Equivalent} labels at training time has a negative effect. We attribute this to the large increase in label quantity ($137\times$) and diversity making the task of fitting a model more complex compared to the \textit{Joint} labels alone.
Despite the extreme data imbalance our method is able to perform when trained on only the joint entities defined in the original ground truth data.

\subsubsection{Human CAD Expert Baseline}
\begin{figure}
    \begin{center}
        \includegraphics[width=\linewidth]{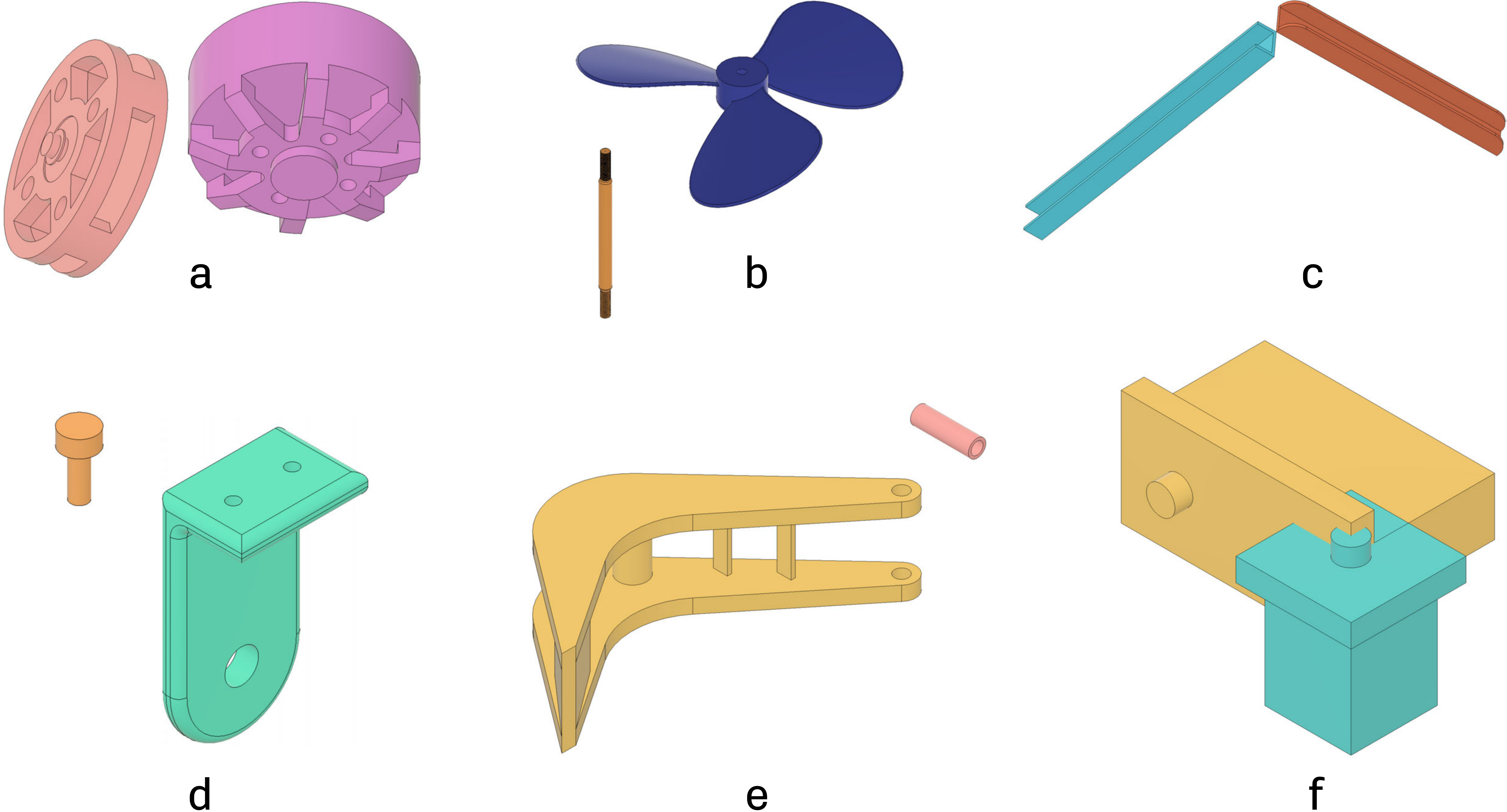}
        \caption{Example designs presented to a CAD expert during our human baseline study.}
        \label{figure:human-study-examples}
    \end{center}
\end{figure}

We now provide further details of the human CAD expert study described in Section~\ref{section:human-cad-expert-baseline}. We perform the study with $100$ samples randomly selected from distributions with and without potential positive unlabeled samples containing \textit{Sibling} entities. The same CAD expert is used for each set and asked to infer a joint configuration from two randomly rotated and translated parts based on what they believe is correct. No further guidance or details about the ground truth solution or the original assembly is provided. Example designs presented to the CAD expert are shown in Figure~\ref{figure:human-study-examples}. 

A limitation of the current study is an imbalance between \textit{Hole} and \textit{No Hole} data samples. From the 100 samples used in each study, random sampling produces $12$ and $13$ \textit{No Hole} samples in each set, with the remainder \textit{Hole} samples. To more accurately measure human performance in designs without holes, a larger sample size would be beneficial. Due to this limited sample size we do not report the break down for \textit{Hole} and \textit{No Hole} accuracy in Table~\ref{table:joint-axis-prediction} and \ref{table:joint-axis-prediction-mix-test}, but detail it here for completeness. For the test set, $77/88$ ($82.95\%$) \textit{Hole} and $7/12$ ($58.33\%$) \textit{No Hole} data samples match the ground truth. For the uniform distribution test set, described in Section~\ref{section:uniform-test-eval}, $66/87$ ($75.86\%$) \textit{Hole} and $1/13$ ($7.69\%$) \textit{No Hole} data samples match the ground truth. Although a limited sample size, the low performance on \textit{No Hole} samples matches our general observation that joints are more difficult to infer when a definitive fastening mechanism is not apparent. This can be observed in Figure~\ref{figure:human-study-examples} where the \textit{Hole} examples (a, b, d, e) generally have a more apparent solution than the \textit{No Hole} examples (c, f).

\subsubsection{Joint Axis Prediction}
We now provide further details of the joint axis prediction experiment described in Section~\ref{section:joint-axis-experiment}.

\paragraph{Joint Origin and Direction}
We derive the joint axis based on the type of B-Rep entity predicted by the network. Table~\ref{table:joint-origin-direction} lists the different B-Rep entity types and the source of the joint axis origin point and direction vector. For NURBS surfaces and curves, no standard way of deriving the joint origin and direction is used. As it is extremely rare to encounter joints defined using NURBS, we simply discard them when training with B-Rep based methods.

\begin{table}
    \centering
    \small
    \begin{tabular}{l|cc}
        \toprule
        \textbf{B-Rep Entity Type} & \textbf{Origin}  & \textbf{Direction} \\
        \midrule
        Plane Surface & Centroid & Normal\\
        Cylinder Surface & Origin & Axis\\
        Cone Surface & Origin & Axis\\
        Sphere Surface & Origin & Z\\
        Torus Surface & Origin & Axis\\
        Elliptical Cylinder Surface & Origin & Axis\\
        Elliptical Cone Surface & Origin & Axis\\
        NURBS Surface & - & - \Bstrut\\
        \hdashline
        Line Curve & Start Point & Line Direction \Tstrut\\
        Arc Curve & Center & Normal\\
        Circle Curve & Center & Normal\\
        Ellipse Curve & Center & Normal\\
        Elliptical Arc Curve & Center & Normal\\
        NURBS Curve & - & -\\
        \bottomrule
    \end{tabular}
    \caption{The joint origin and direction information used to define a joint axis, are derived from B-Rep entities as described above.}
    \label{table:joint-origin-direction}
\end{table}

\begin{figure}
    \begin{center}
        \includegraphics[width=\linewidth]{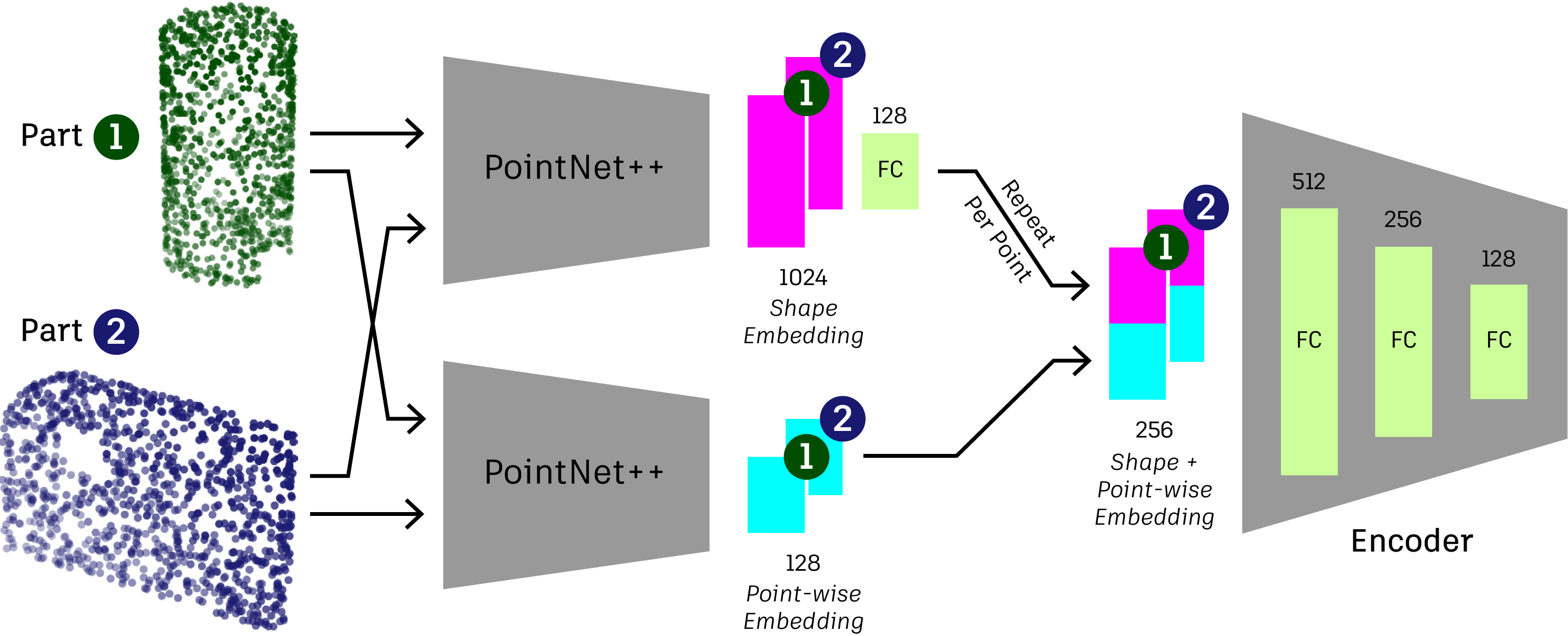}
        \caption{Encoder network architecture used for the B-Dense and B-Discrete baseline methods.}
        \label{figure:pcd-axis-encoder}
    \end{center}
\end{figure}

\paragraph{Baseline Implementation}
We now describe further implementation details for the baseline methods described in Section~\ref{section:joint-axis-experiment}. For the point cloud baselines we use area-based sampling on a mesh of each part using the Trimesh library\footnote{\href{https://trimsh.org}{https://trimsh.org}}. We create 1024 points, for a total of 2048 points combined. We normalize each part to the unit range of $-1$ to $1$ and center it at the origin.
We use a common encoder architecture for the B-Dense and B-Discrete baselines, illustrated in Figure~\ref{figure:pcd-axis-encoder}. 
Key here is adapting the architecture to our setting where \textit{two parts} are used as input rather than one. We want to ensure that the embeddings contain features from both parts, so meaningful predictions can be made.
The network takes as input two separate point clouds and creates both a shape embedding and a point-wise embedding from two separate PointNet++~\cite{qi2017pointnet++} encoders, using the PyTorch implementation\footnote{\href{https://github.com/yanx27/Pointnet_Pointnet2_pytorch}{https://github.com/yanx27/Pointnet\_Pointnet2\_pytorch}}. The shape embedding, representing the global shape of each part, is passed through a fully-connected layer then repeated and concatenated to the point-wise embedding of the opposing part. At this stage we have embeddings for each part containing combined shape and point-wise features. These embeddings are then passed through a shared 3-layer MLP encoder using ReLU activation and dropout.

Specific details of the decoder and loss functions are described below. Common to both point cloud baselines, and unique to our setting, is the need to calculate the loss over $\ge 1$ ground truth joint axes. As any ground truth joint axis is considered correct, we calculate the loss from the single network prediction against each ground truth joint axis and take the minimum. During evaluation we follow the same approach and consider a joint axis prediction a `hit' if it is collinear with \textit{any} ground truth joint axis in either direction. To evaluate collinearity we use a distance threshold of $0.1$, equating to $5\%$ of the unit range of $-1$ to $1$, and an angular threshold of $10^\circ$, equating to approximately $5\%$ of the $180^\circ$ range, as we consider either axis direction to be correct.

\paragraph{B-Dense Implementation}
The B-Dense baseline follows Li et al.~\cite{li2020category}, which densely regresses   association and joint parameters for each point in the point cloud. The  association parameters is a per-point variable estimated by the network which predicts if a given point is near a given joint. The ground truth association is calculated for a point  if the point is below a certain  Eucidean distance threshold to the joint direction axis. We follow Li et al.~\cite{li2020category} and use the cross entropy loss to compare the predicted  association  with the ground truth  association for every point. The joint parameter is a 7D vector predicted by the network for each point, where the first 3 dimensions represent the joint direction unit vector and the last 4 dimensions the pivot point. We compare the predicted joint direction unit vector with the ground truth joint direction unit vector using mean squared error loss for only the points associated with that joint. The pivot point is calculated by predicting a projection unit vector and the distance from a given point to the joint axis. Similar to predicting the joint direction, for each associated point we regress the  projection unit vector and the scalar distance with the ground truth parameters respectively. Note during test time, similar to Li et al.~\cite{li2020category}  a voting scheme is applied across the associated points to regress the pivot point and joint direction axis. 

We train the B-Dense baseline with a per-GPU batch size of 7 across 8 NVIDIA V100 GPUs, dropout of 0.1, batch norm enabled, learning rate of 0.0003 with the Adam optimizer, and a learning rate scheduler that reduces on plateau. The threshold parameter for the ground truth association is set to 0.2.

\paragraph{B-Discrete Implementation}
The B-Discrete baseline follows Shape2Motion~\cite{wang2019shape2motion} and uses a hybrid of discrete classification and regression to predict the joint origin point and direction vector. For the joint origin point prediction, the network is trained to select points in the input point cloud closest to the ground-truth. For that, a binary indicator vector is used, and a cross-entropy is employed for optimization. In addition, a displacement vector is used to estimate the displacement between anchor points and ground-truth origin points. The displacement is optimized with an L2 loss. For the direction vector estimation, we discretize each dimension of the direction vector into 14 classes. The B-Discrete model estimates a class probability and a residual for each dimension to correct the error through discretization. The orientation loss comprises the cross-entropy loss for the classification and the L2 for the residual error. The overall loss is a summation of joint origin and orientation losses. When adapting this baseline, we refer to the authors' original implementation\footnote{\href{https://github.com/wangxiaogang866/Shape2Motion}{https://github.com/wangxiaogang866/Shape2Motion}}.

We train the B-Discrete baseline with a per-GPU batch size of 16 across 4 NVIDIA Quadro RTX 6000 GPUs, dropout of 0.1, batch norm enabled, learning rate of 0.0003 with the Adam optimizer, and a learning rate scheduler that reduces on plateau.

\paragraph{B-Grid Implementation}
The B-Grid baseline follows the grid sampling approach of UV-Net~\cite{jayaraman2021uvnet} and uses a CNN-based encoder\footnote{\href{https://github.com/AutodeskAILab/UV-Net}{https://github.com/AutodeskAILab/UV-Net}} to generate vertex embeddings before message passing is performed.
The remainder of the network is identical to our method, including the joint axis prediction branch and loss function. During training we randomly rotate the input parts by $45^\circ$ increments about the $x$, $y$, and $z$ axes to improve generalization. Due to memory limitations, we skip training samples where the graph representations of both parts have more than 950 graph vertices combined.
We train the B-Grid baseline with a per-GPU batch size of 2 across 4 NVIDIA V100 GPUs, dropout disabled, batch norm enabled, learning rate of 0.0001 with the Adam optimizer, and a learning rate scheduler that reduces on plateau.

\begin{table}
    \centering
    \small
    \begin{tabular}{l|cccc}
        \toprule
        & \textbf{All} & \textbf{Hole} & \textbf{No Hole} & \textbf{Param.}  \\
        & \textbf{Acc.\%} $\uparrow$ & \textbf{Acc.\%} $\uparrow$ & \textbf{Acc.\%} $\uparrow$ & \textbf{\#} $\downarrow$  \\
        \midrule
        \textbf{Ours}   & \textbf{62.22}     & \textbf{63.47}     & \textbf{56.91}     & 1.3M  \\ 
        B-Dense         & ~6.00     & ~5.24     & 10.00     & 3.2M  \\ 
        B-Discrete      & ~2.71     & ~2.86     & ~2.09     & 4.0M  \\
        B-Grid          & 50.34     & 51.19     & 46.72     & 3.1M  \\ 
        B-Heuristic     & 57.99     & 60.54     & 47.11     & -     \\
        B-Random        & 15.85     & 16.08     & 15.59     & -     \\
        \midrule
        Human           & 69.00     & -         & -         & -     \\
        \bottomrule
    \end{tabular}
    \caption{Joint axis prediction accuracy results are shown for all data samples in the uniform distribution test set (All), the subset of data samples with holes (Hole) and without holes (No Hole). The number of network parameters is also shown (Param.). Finally, results from a human CAD expert on 100 test samples are shown.}
    \label{table:joint-axis-prediction-mix-test}
\end{table}

\paragraph{Other Point Cloud Approaches}
In addition to the point cloud baselines reported in Table~\ref{table:joint-axis-prediction}, we attempted several other approaches that failed to produce results. We find that direct regression of the joint origin point and direction vector, similar to RPM-Net~\cite{yanRPMNet19}, struggles to align with the ground truth axes until the loss approaches zero. We find that classification of the joint origin and direction, by predicting each over a quantized space, only slightly improves compared to a regression-based approach. We also attempted data augmentation using random rotation of each point cloud but found it reduces overall accuracy. We attribute this to the point cloud based networks achieving higher accuracy by over-fitting on a subset of the data.

\paragraph{Evaluation on a Uniform Distribution}
\label{section:uniform-test-eval}

In addition to the `clean' test set, we retain an additional test set that matches the original data distribution and contains potential positive unlabeled samples. Figure~\ref{figure:human-study-examples}, left shows an example of a positive unlabeled data sample where two holes exist for the bolt to be inserted, but only one is labeled. Although the uniform distribution test set cannot be used to reliably judge accuracy on the joint axis prediction task, we include the results in Table~\ref{table:joint-axis-prediction-mix-test} for completeness and to demonstrate the effect of positive unlabeled samples during evaluation. We note that the overall accuracy drops for all methods, but the relative position of each method stays the same. The results underline the importance of removing potential positive unlabeled data samples for accurate evaluation.

\begin{figure}
    \begin{center}
        \includegraphics[width=\linewidth]{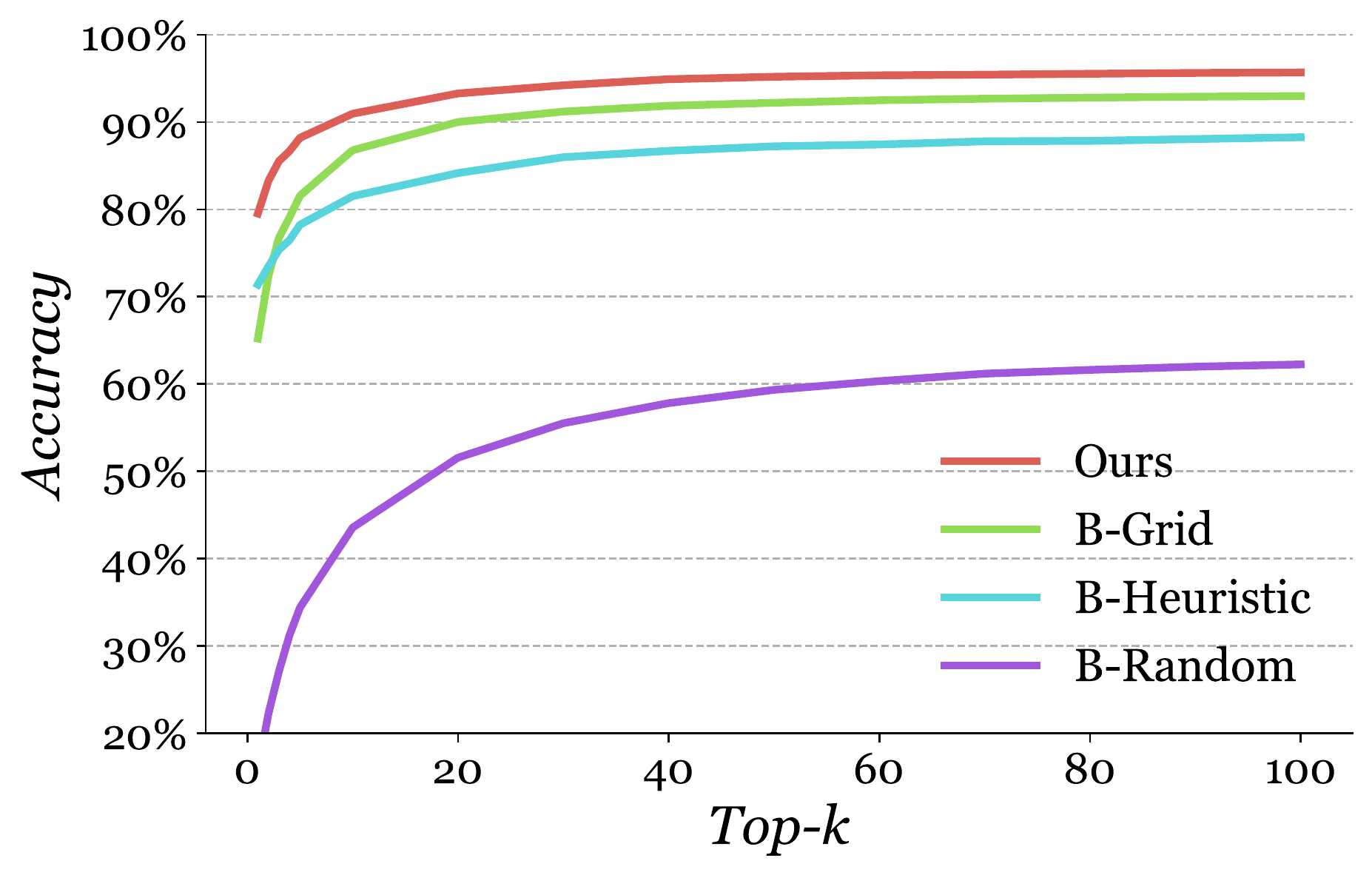}
        \caption{Joint axis prediction top-$k$ accuracy.}
        \label{figure:joint-axis-prediction-top-k}
    \end{center}
\end{figure}

\paragraph{Top-k Accuracy}
In Figure~\ref{figure:joint-axis-prediction-top-k} we show the top-$k$ accuracy results to supplement the top-$1$ results for joint axis prediction reported in Table\ref{table:joint-axis-prediction}. We plot the top-$k$ accuracy for all relevant methods. We observe that both learning-based and heuristic methods saturate at $k\approx50$. We note that the B-Grid baseline outperforms B-Heuristic when $k>2$. We attribute this to the UV-grid representation of the geometry as well as the local aggregation of features.
B-Heuristic and Ours both have access to more precise input features, such as lengths, areas, and entity types, compared to B-Grid which uses discretely sampled UV-grids to represent the surface and curve shapes. Increasing the sampling rate of curves and surfaces by using finer UV-grids, beyond the  $10 \times 10$ sampling used in the original paper~\cite{jayaraman2021uvnet}, may yield better performance at the cost of memory and compute.
B-Heuristic makes predictions on a per-entity level rather than aggregating features from a neighborhood, like B-Grid and our method does, and is unable to take the local shape into account.
By choosing the right features for this task and employing a message-passing network, our method outperforms both B-Grid and B-Heuristic in any top-$k$ setting.

\begin{figure*}
    \begin{center}
        \includegraphics[width=0.98\linewidth]{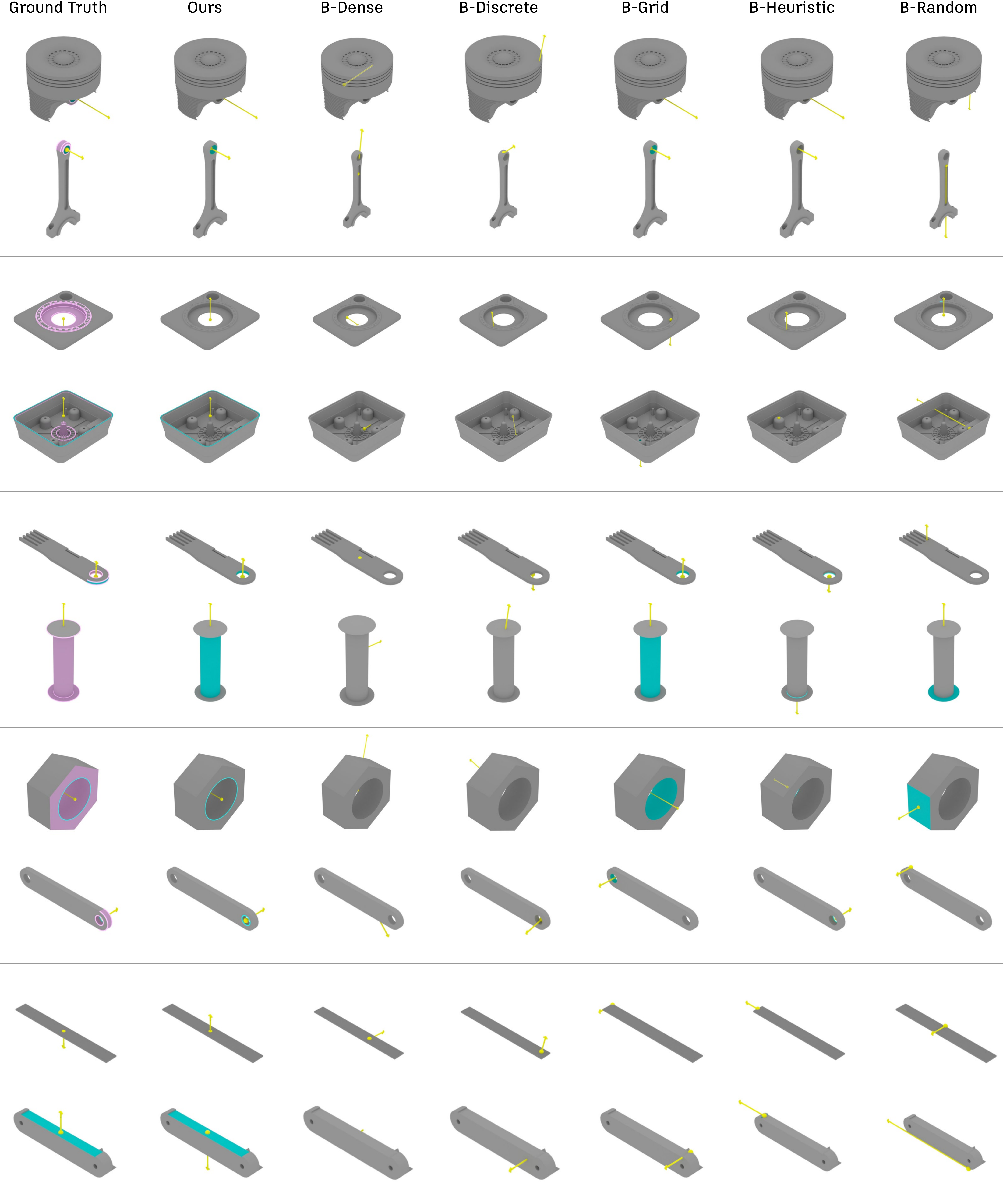}
        \caption{Qualitative results on the joint axis prediction task. Each of the two parts used to define a joint are shown individually in separate rows, with the different predictions from each method shown in columns. For all methods we visualize the joint axis with a yellow arrow. For the ground truth we show the designer-selected B-Rep entities in cyan, and the equivalent entities in pink. For B-Rep based predictions, we show the predicted B-Rep entities in cyan.}
        \label{figure:qualitative-axis-results-sup-mat-01}
    \end{center}
\end{figure*}

\begin{figure*}
    \begin{center}
        \includegraphics[width=0.98\linewidth]{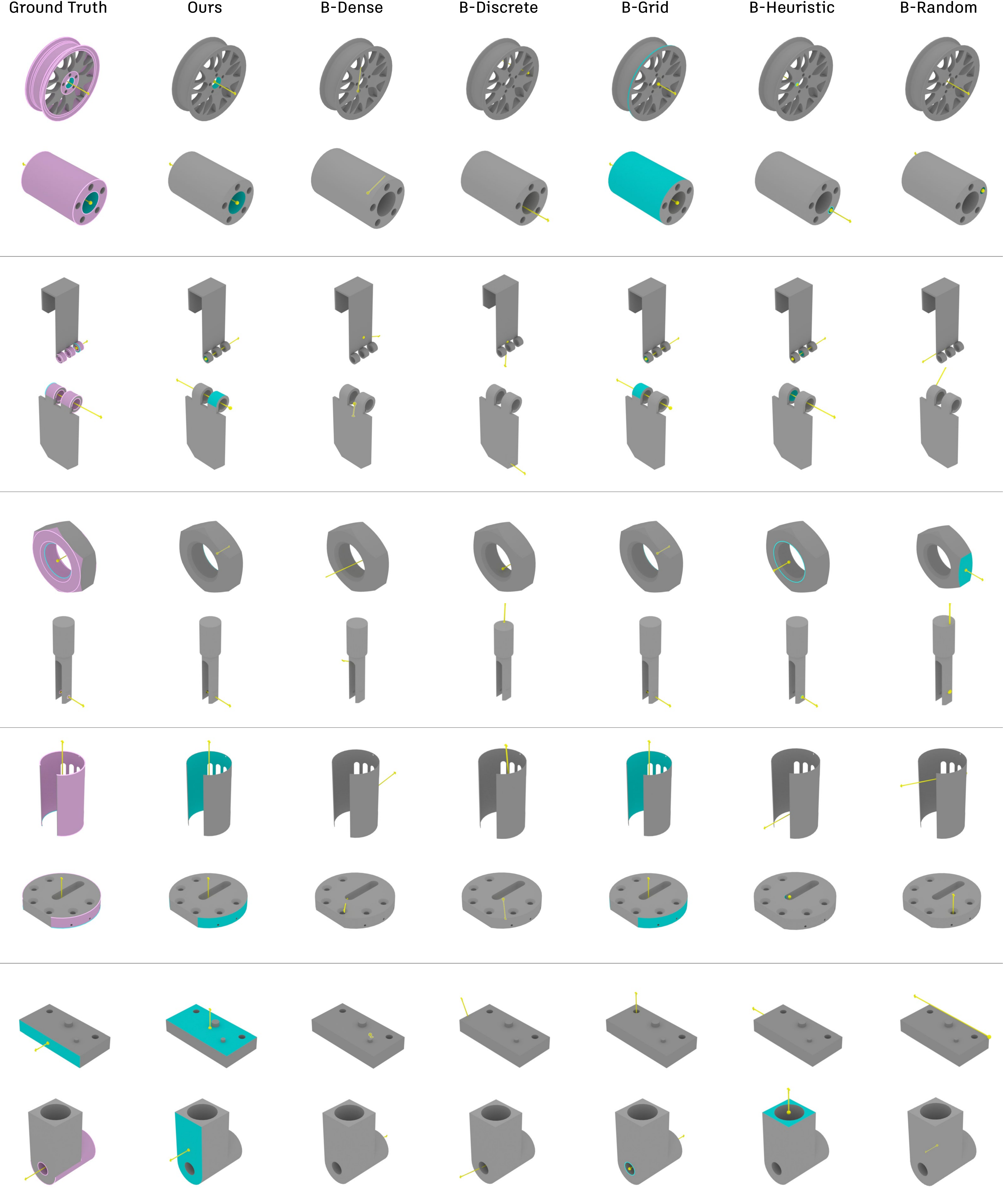}
        \caption{Qualitative results on the joint axis prediction task. Each of the two parts used to define a joint are shown individually in separate rows, with the different predictions from each method shown in columns. For all methods we visualize the joint axis with a yellow arrow. For the ground truth we show the designer-selected B-Rep entities in cyan, and the equivalent entities in pink. For B-Rep based predictions, we show the predicted B-Rep entities in cyan.}
        \label{figure:qualitative-axis-results-sup-mat-02}
    \end{center}
\end{figure*}

\paragraph{Qualitative Results}
In Figure~\ref{figure:qualitative-axis-results-sup-mat-01} and \ref{figure:qualitative-axis-results-sup-mat-02} we show qualitative results for the joint axis prediction task described in Section~\ref{section:joint-axis-experiment}. We show each part used to form a joint individually in separate rows. Horizontal dividing lines are used to separate pairs of parts from different joint sets. The different predictions from each method are shown in columns. For all methods we visualize the joint axis with a yellow arrow. For the ground truth we show the designer-selected B-Rep entities in cyan, and the equivalent entities in pink. For B-Rep based predictions, we show the predicted B-Rep entities in cyan. We observe that by making predictions over the B-Rep entities directly the derived joint axis results naturally align to geometry. On the other hand, both point cloud methods, B-Dense and B-Discrete, suffer from noisy alignment with the geometry due to the use of regression.

\subsubsection{Joint Pose Prediction}
We now provide further details of the joint pose prediction experiment described in Section~\ref{section:joint-pose-experiment}.

\paragraph{Joint Pose Search Implementation}
Our joint pose search procedure takes as input the top-$k$ joint axis predictions from our network, together with a pair of parts, and outputs a selected joint axis prediction, offset, rotation, and flip parameter that can be used to assemble the parts.
Our search procedure iterates over the top-$k$ joint axis predictions and applies the Nelder-Mead algorithm~\cite{nelder1965simplex} to optimize the offset and rotation parameters based on the cost function described in Section~\ref{section:joint-pose-search}. We optimize the continuous offset and rotation parameters with and without the discrete flip parameter enabled, for a total of $k \times 2$ optimization runs for each data sample. We set the initial simplex parameters to zero offset and rotation. For parts that have rotational symmetry about the predicted joint axis, we optimize only for the offset parameter and set the rotation parameter to zero. Finally, we take the optimal set of parameters along with the corresponding joint axis prediction that minimizes the cost function. Using these predictions we recover the final joint pose as a rigid body transform that aligns part one to part two. For each data sample, running search takes on average $2.58$ seconds, excluding inference time, on a server with an Intel Xeon Platinum 3.41 GHz CPU.

To compute the overlap area $A_{1 \cap 2}$ in the cost function (Equation \ref{eq:overlap}), we sample a dense set of points on the surface of part 1 and calculate the proportion $p_A$ of points that are also on the surface of part 2 within a small tolerance. To perform this calculation we use a pre-computed signed distance field generated from part 2. Then, $A_{1 \cap 2}$ is computed as $A_{1 \cap 2} = p_A A_1$. Similarly, we compute the overlap volume $\volume_{1 \cap 2}$ by sampling points inside the volume of part 1 and calculating the proportion $p_\volume$ of points inside of part 2 (i.e. points that have a negative distance to part 2). As a result, $\volume_{1 \cap 2}$ is computed as $\volume_{1 \cap 2} = p_\volume \volume_1$.

\begin{figure}
    \begin{center}
        \includegraphics[width=\linewidth]{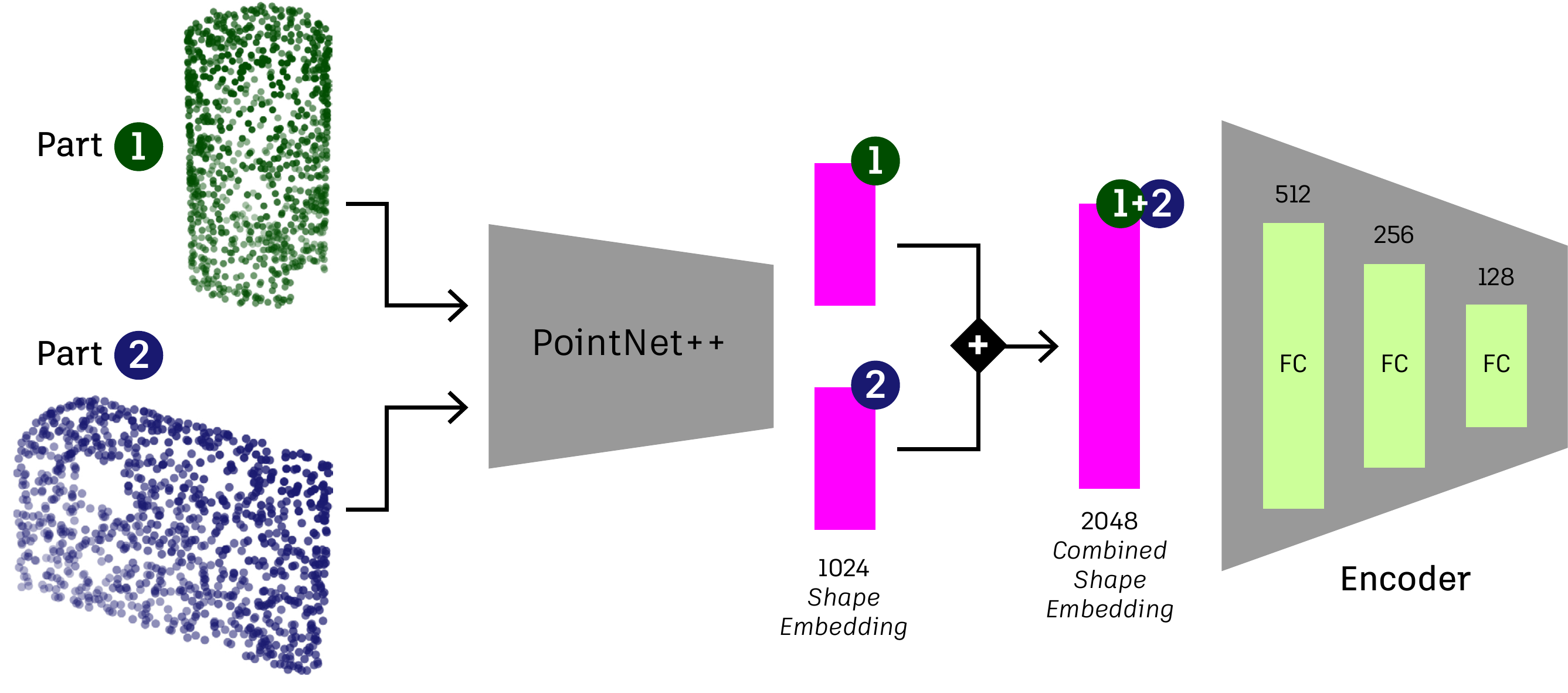}
        \caption{Encoder architecture for the B-Pose baseline method.}
        \label{figure:pcd-pose-encoder}
    \end{center}
\end{figure}

\paragraph{B-Pose Implementation}
The B-Pose baseline follows Huang et al.~\cite{huang2020generative} to regress a translation point and rotation quaternion using a combination of L2 and CD loss terms. We adapt the related code from the authors PyTorch implementation\footnote{\href{https://github.com/hyperplane-lab/Generative-3D-Part-Assembly}{https://github.com/hyperplane-lab/Generative-3D-Part-Assembly}} to our setting. We sample points with area-based sampling on a mesh of each part using the Trimesh library. We create 1024 points, for a total of 2048 points combined. We normalize each part to the unit range of $-1$ to $1$ and center it at the origin. We use a similar encoder architecture to that used with the joint axis prediction baselines, illustrated in Figure~\ref{figure:pcd-pose-encoder}. 
Key here is adapting the architecture to our setting where we predict a single rigid body transform, in the form of a translation point and rotation quaternion, to assembly a pair of parts. We want to ensure that the embeddings contain features from both parts, so meaningful predictions can be made. The network takes as input two separate point clouds and creates a shape embedding for each with a PointNet++~\cite{qi2017pointnet++} encoder, using the PyTorch implementation\footnote{\href{https://github.com/yanx27/Pointnet_Pointnet2_pytorch}{https://github.com/yanx27/Pointnet\_Pointnet2\_pytorch}}. The shape embedding for both parts are concatenated together to form a combined embedding. This embedding is then passed through a 3-layer MLP encoder using ReLU activation and dropout.

Regression of the translation point and rotation quaternion follows the original paper. We adapt the loss functions to calculate the loss over $\ge 1$ ground truth pose configurations. As any ground truth pose is considered correct, we calculate the loss from the single network prediction against each ground truth joint pose and take the minimum. We train the B-Pose baseline with a per-GPU batch size of 8 across 4 NVIDIA V100 GPUs, dropout of 0.1, batch norm disabled, learning rate of 0.0003 with the Adam optimizer, and a learning rate scheduler that reduces on plateau.

\paragraph{Evaluation}
During evaluation we compare over multiple ground truth poses and take the joint pose prediction with the lowest CD. To ensure accurate evaluation results we re-sample all parts with 4096 points, apply the predicted transform to move the parts into an assembled state, and finally calculate CD.

\begin{figure*}
    \begin{center}
        \includegraphics[width=\textwidth]{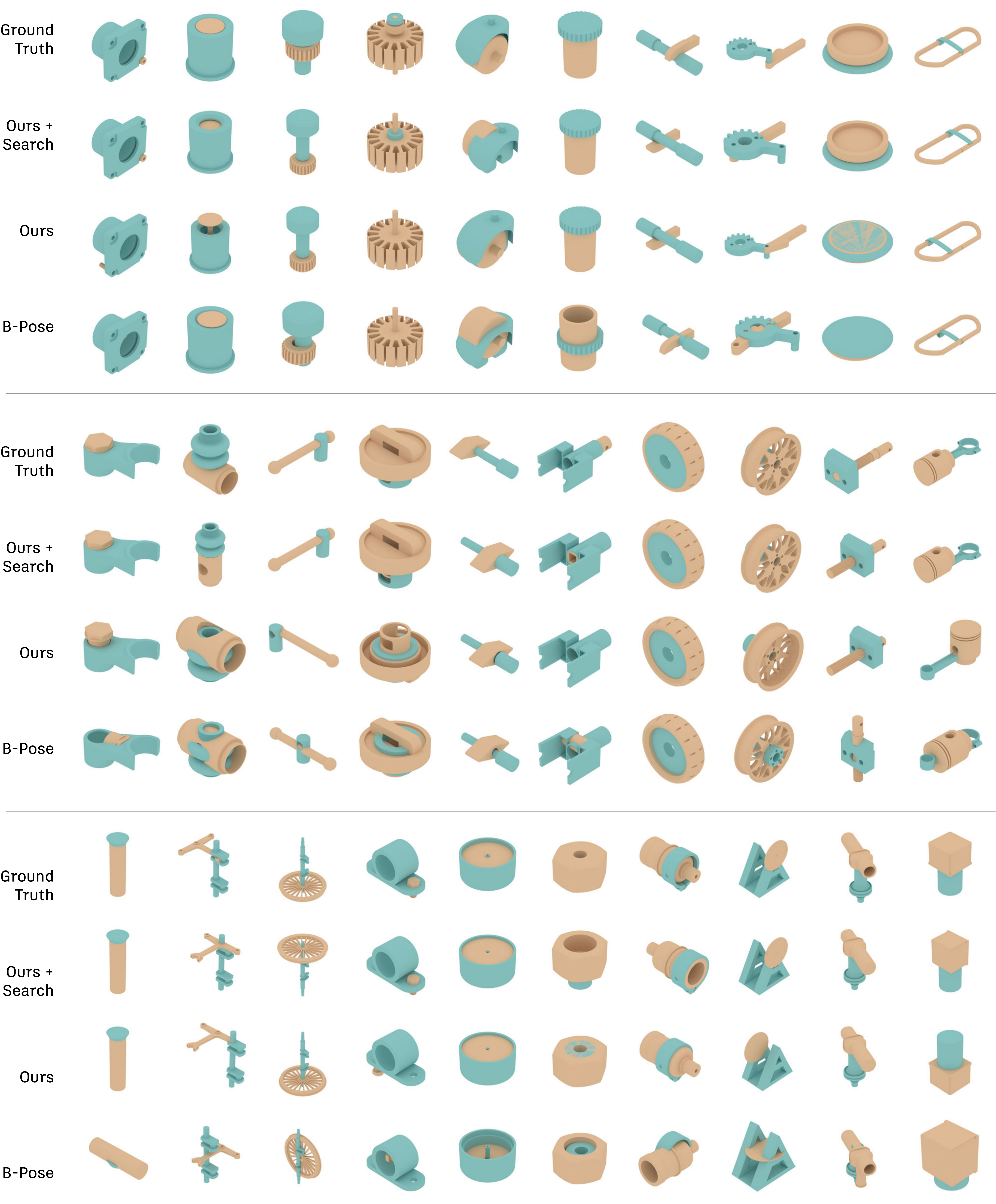}
        \caption{Additional qualitative joint pose prediction results comparing our method, with and without search, with the B-Pose baseline.}
        \label{figure:qualitative-pose-results-sup-mat}
    \end{center}
\end{figure*}

\paragraph{Qualitative Results}
In Figure~\ref{figure:qualitative-pose-results-sup-mat} we show additional qualitative results for the joint pose prediction task. The ground truth and each method is shown on its own row, with different assembled joints shown in columns.

\subsection{Future Work}
\label{section:appendix_future_work}
In this section we discuss future applications and extensions of our method.

\subsubsection{Future Applications}
\begin{figure}
    \begin{center}
        \includegraphics[width=\linewidth]{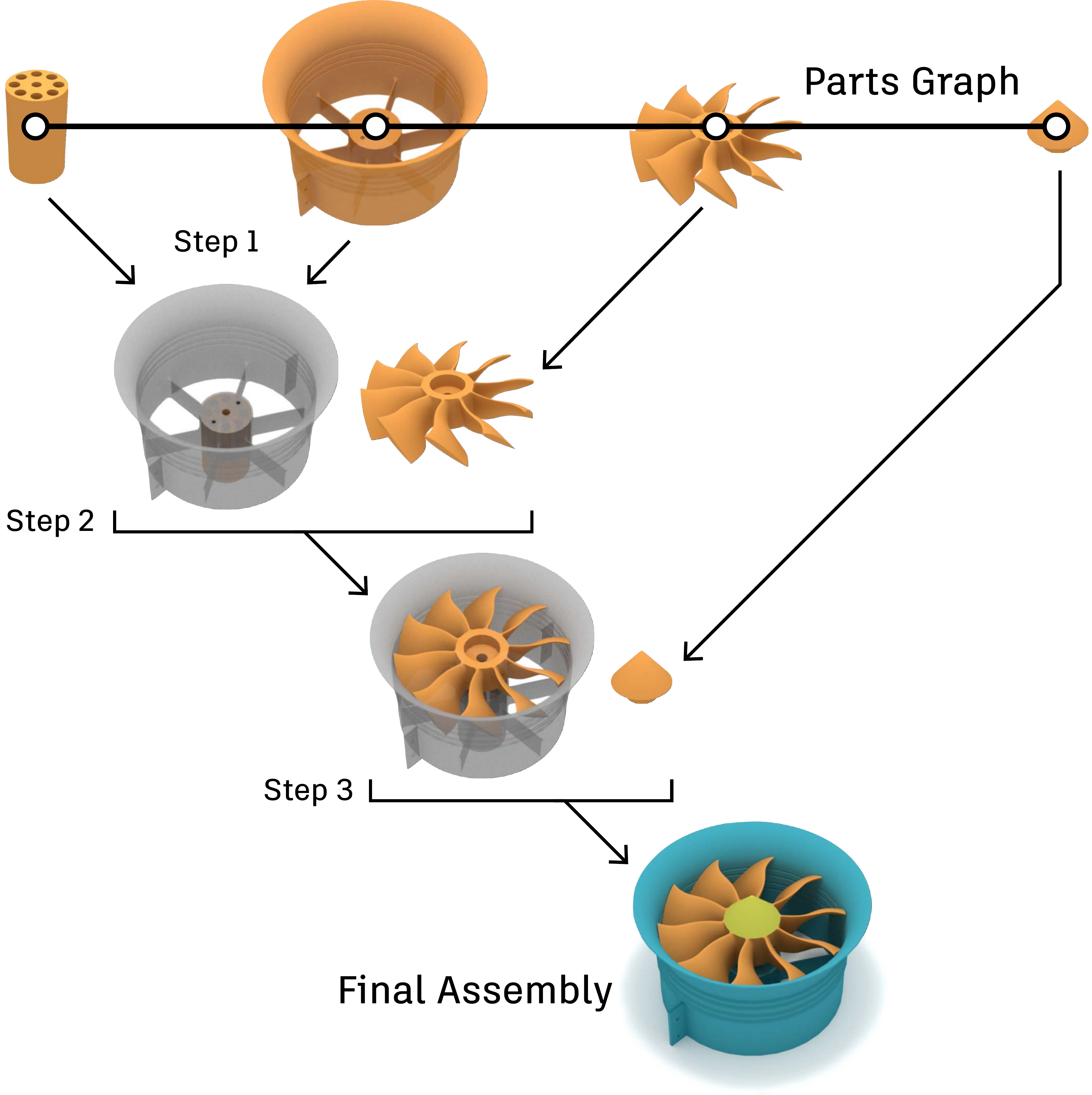}
        \caption{Example multi-part assembly sequence derived from a parts graph.}
        \label{figure:multi-part-search}
    \end{center}
\end{figure}

We now provide further details of the multi-part assembly demonstration described in Section~\ref{section:discussion}. As input we use assembly data from our dataset and select samples from a test set that our network has not been trained on. Using the joint and contact information provided with the assembly data we form a parts graph that connects the parts together (Figure~\ref{figure:multi-part-search}, top). Based on this graph we manually derive an assembly sequence of parts to be placed.

We begin the assembly process by presenting the first pair of parts (Figure~\ref{figure:multi-part-search}, Step 1) to our pre-trained network and perform joint pose search using the network predictions. Our search procedure follows that described in Section~\ref{section:joint-pose-search}. We next follow an iterative procedure where we assemble the first pair of parts to form the current state of the assembly and repeat joint pose search between the current assembly and the next part in the sequence that has yet to be placed (Figure~\ref{figure:multi-part-search}, Step 2-3). We adapt the overlap volume and contact area in the cost function to be defined between the current state of the assembly and the part to be assembled. We repeat this procedure until all parts have been placed.

\begin{figure}
    \begin{center}
        \includegraphics[width=\columnwidth]{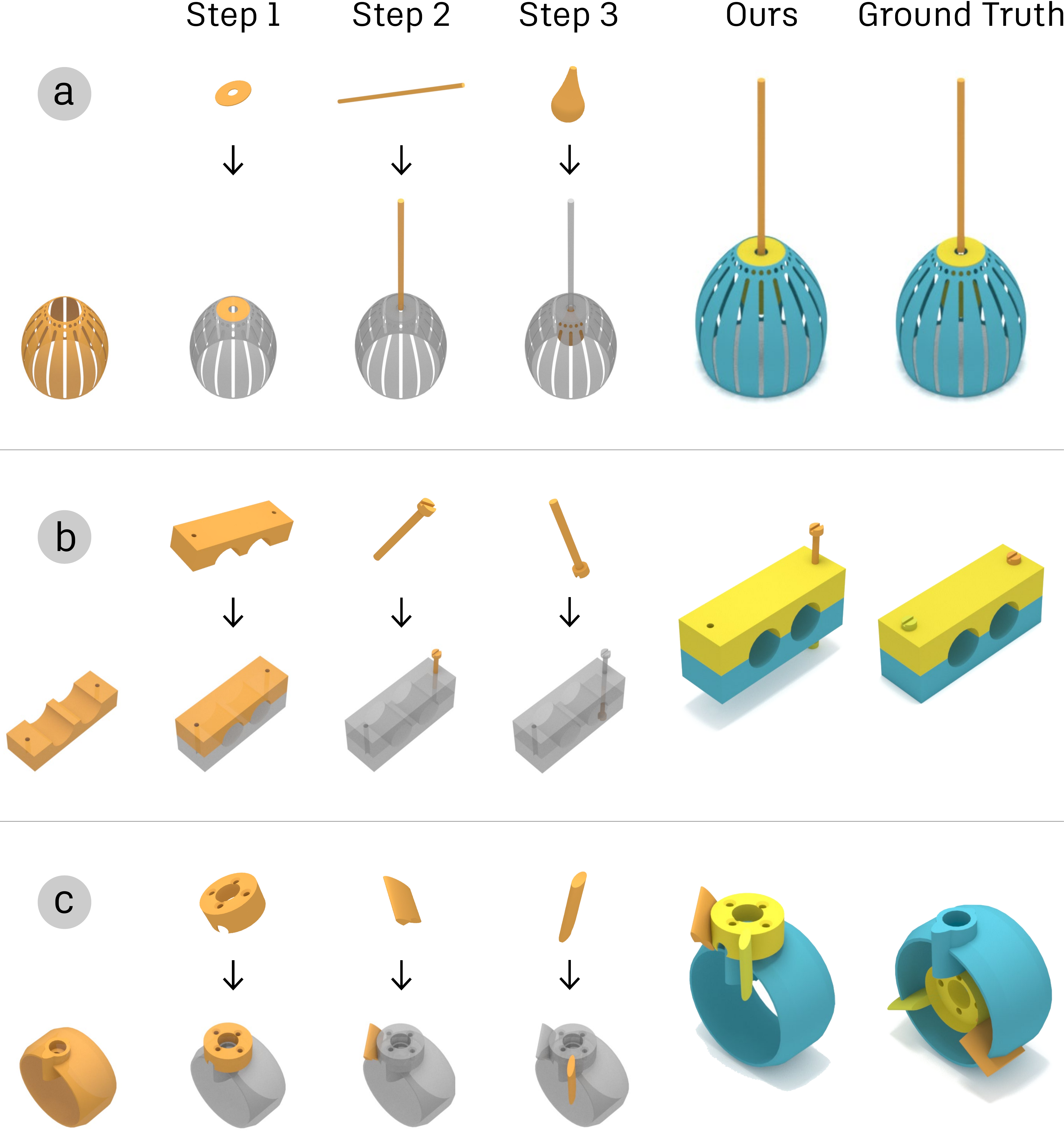}
        \caption{Additional examples of multi-part assembly.}
        \label{figure:multi-part-additional-examples}
    \end{center}
\end{figure}

Our basic extension of joint pose search to the multi-part setting has several limitations that we show in Figure~\ref{figure:multi-part-additional-examples}. Assemblies with a clear single axis, such as Figure~\ref{figure:multi-part-additional-examples}a, are less challenging for our approach compared to those with multiple axes. For example, Figure~\ref{figure:multi-part-additional-examples}b shows a partial failure case where the main parts are correctly assembled but placement of the smaller parts is incorrect. Considering symmetry as an assembly criteria may help resolve such cases. We find that errors early on compound at each step of the assembly, and reduce the overall chance of success. Figure~\ref{figure:multi-part-additional-examples}c shows a failure case where an incorrect prediction at the first step impacts subsequent assembly steps and diverges significantly from the ground truth design. Finally, in all examples we assume that a well defined assembly sequence is provided as input, when in practice that may not be possible and an automated process would be desirable. 

\begin{figure}
    \begin{center}
        \includegraphics[width=\columnwidth]{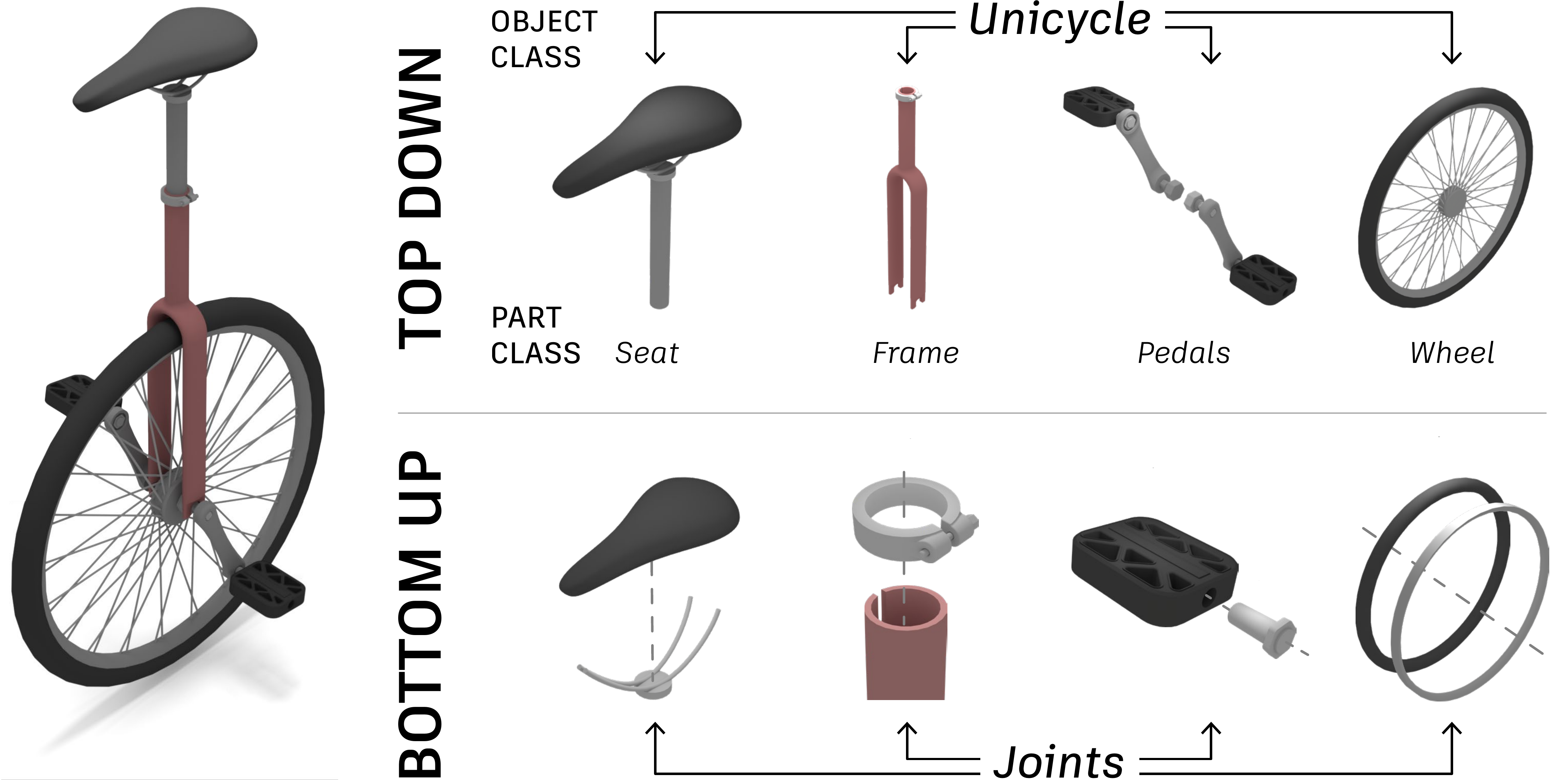}
        \caption{A future hybrid approach to assembly may look to combine \textit{top down} knowledge and composition of objects and parts, with \textit{bottom up} assembly using locally defined joints. }
        \label{figure:future-hybrid-approach}
    \end{center}
\end{figure}

Ultimately we believe a hybrid approach, such as illustrated in Figure~\ref{figure:future-hybrid-approach}, will be effective to combine \textit{top down} knowledge and composition of objects and parts, with the precision offered by \textit{bottom up} assembly of locally defined joints. We leave this investigation to future work.

\subsubsection{Future Extensions}
A limitation of our current network is that it does not leverage geometric or physics-based loss terms that may help with avoiding undesirable overlap between parts. A hybrid approach that combines points sampled from B-Rep entities~\cite{jayaraman2021uvnet} with a geometric loss term or uses physics simulations in a reinforcement learning environment, may improve upon our current results.

Another line of research is to experiment with conditioning the network to produce targeted output. For example, providing the current state of the assembly as conditioning may help resolve ambiguities when predicting where in an assembly a new part should be connected using a joint.

\subsection{Broader Impact}

This work is motivated by the opportunity to encourage sustainable design through the reuse of existing physical components. Such components are often available in recycling streams, dead inventory, or existing supply chains. By enabling assembly aware design tools that can identify and automatically place suitable components, we hope to reduce the cost and negative environmental impact of establishing new tooling for manufacturing and the associated supply chains~\cite{hatcher2011design}. Intellectual property is a key consideration with data driven approaches to design, as there exists the risk of unauthorized use or appropriation of existing designs. This risk is balanced by providing publicly available datasets, such as ours, and by component suppliers who freely provide CAD models to encourage the use and sale of their components.

\subsection{Acknowledgments}

We would like to acknowledge the assistance of Eder Duran and Kamal Rahimi Malekshan for developing data extraction tools, Yewen Pu, Chin-Yi Cheng, Ye Wang, Leonardo Hernandez Cano, Ran Zhang, Jie Xu,  and Tao Du for helpful early discussions, and Tonya Custis, Sachin Chitta, and Hui Li for reviewing early drafts of the paper.

\end{document}